\documentclass[]{academic_template}
\IfFileExists{latexml.sty}{\usepackage{latexml}}{\ifdefined\iflatexml\else\newif\iflatexml\fi
}
\providecommand{\printprojectlinks}{}
\iflatexml
  \usepackage{authblk}
  \usepackage{hyperref}
  \makeatletter
  \newcount\vxl@nextlabel
  \def\vxl@getlabel#1{\vxl@labelhead#1\vxl@end}
  \def\vxl@labelhead#1#2\vxl@end{\ifcsname vxl@label@\detokenize{#1#2}\endcsname\else
      \global\advance\vxl@nextlabel by 1
      \expandafter\xdef\csname vxl@label@\detokenize{#1#2}\endcsname{vxl\the\vxl@nextlabel}\fi
    \edef\vxl@resolved{\csname vxl@label@\detokenize{#1#2}\endcsname}}
  \def\vxl@stop{\vxl@stop}
  \def\vxl@parselabel#1,{\def\vxl@item{#1}\ifx\vxl@item\vxl@stop
      \let\vxl@next\relax
    \else
      \if\relax\detokenize{#1}\relax\else
        \vxl@getlabel{#1}\ifx\vxl@converted\@empty
          \edef\vxl@converted{\vxl@resolved}\else
          \edef\vxl@converted{\vxl@converted,\vxl@resolved}\fi
      \fi
      \let\vxl@next\vxl@parselabel
    \fi
    \vxl@next
  }
  \def\vxl@convertlabels#1{\def\vxl@converted{}\vxl@parselabel#1,\vxl@stop,
  }
  \let\vxl@nativeauthor\author
  \newcommand{\vxl@htmlauthor}[2][]{\vxl@convertlabels{#1}\expandafter\vxl@nativeauthor\expandafter[\vxl@converted]{#2}}
  \let\author\vxl@htmlauthor
  \renewcommand{\affiliation}[2][]{\vxl@convertlabels{#1}\ifx\vxl@converted\@empty
      \affil{#2}\else
      \expandafter\affil\expandafter[\vxl@converted]{#2}\fi
  }

  \def\vxl@htmlinfo{}
  \providecommand{\contribution}{}
  \renewcommand{\contribution}[2][]{\if\relax\detokenize{#1}\relax
      \g@addto@macro\vxl@htmlinfo{\par\noindent\emph{#2}\\\par}\else
      \vxl@convertlabels{#1}\expandafter\affil\expandafter[\vxl@converted]{\emph{#2}}\fi
  }
  \providecommand{\checkdata}{}
  \renewcommand{\checkdata}[2][]{\g@addto@macro\vxl@htmlinfo{\par\noindent\textbf{#1:} #2\\\par}}

  \date{}
  \renewcommand{\printprojectlinks}{\vxl@htmlinfo
    \global\let\vxl@htmlinfo\@empty
  }

  \providecommand{\setthemecolor}[1]{}
  \providecommand{\setsectionfont}[1]{}
  \providecommand{\setsubsectionfont}[1]{}
  \providecommand{\setsectionlinecolor}[1]{}
  \providecommand{\setsectionlinethickness}[1]{}
  \providecommand{\settitlefont}[1]{}
  \providecommand{\setabstractfont}[1]{}
  \providecommand{\setabstractframecolor}[1]{}
  \providecommand{\setabstractbgcolor}[1]{}
  \providecommand{\setlogoheight}[1]{}
  \providecommand{\setlogospacing}[1]{}
  \providecommand{\setlogotolineshift}[1]{}
  \providecommand{\settoprulethickness}[1]{}
  \providecommand{\setbottomrulethickness}[1]{}
  \providecommand{\settitlerulethickness}[1]{}

  \providecommand{\sectionformatnoline}{}
  \providecommand{\sectionformatwithline}{}

  \providecommand{\abstractboxon}{}

  \providecommand{\setsjtublue}{}
  \providecommand{\themeblue}[1]{\textbf{#1}}

  \makeatother
\fi

\usepackage[toc,page,header]{appendix}

\usepackage{amssymb}
\usepackage{amsmath}
\usepackage{enumitem}

\usepackage{makecell}
\usepackage{algorithm}      \usepackage{algpseudocode}  
\usepackage{graphicx}

\usepackage{listings}
\usepackage{url}
\usepackage{booktabs}
\usepackage{wrapfig}

\usepackage{colortbl}
\usepackage{xcolor}

\usepackage{caption}
\usepackage{subcaption}
\usepackage{marvosym}
\usepackage{xurl}
\usepackage{titletoc}

\definecolor{takeawaybackground}{HTML}{EAF4FE}
\definecolor{takeawayborder}{HTML}{5A9FE5}
\newtcolorbox{takeawaybox}{
  colback=takeawaybackground,
  colframe=takeawayborder,
  boxrule=0.8pt,
  arc=2mm,
  boxsep=0pt,
  left=7pt, right=7pt,
  top=6pt, bottom=6pt,
  before skip=8pt,
  after skip=8pt,
  fontupper=\itshape
}

\setlogoheight{14mm}   \setlogospacing{5mm}
\setsjtublue 
\settitlerulethickness{3pt}  
\abstractboxon \setabstractframecolor{gray} \setabstractbgcolor{gray!10} \setlogotolineshift{5mm} 
\settoprulethickness{2.5pt}
\setbottomrulethickness{1.5pt}

\title{\raisebox{-0.2\height}{\includegraphics[height=1.6em]{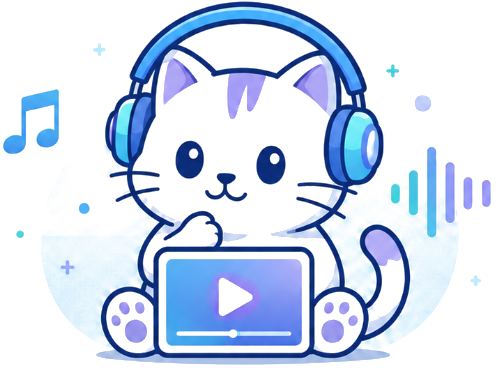}}\hspace{0.15em}EvaOmni: An Evidence-Grounded Benchmark for Audio-Visual Reasoning
}

\author[1,2\dag]{Xiaoxing Hu}
\author[1*]{Wenxi Yao}
\author[3*]{Chaoyuan Shi}
\author[2,4*]{Ziqi Ye}
\author[4]{Xiaorong Guo}
\author[4]{Peixuan Li}
\author[5]{Zhuoran Sun}
\author[1]{Ziyang Gong}
\author[1]{Peiyuan Zhang}
\author[1]{Mingxin Liu}
\author[1]{Shuran Ma}
\author[6]{Hongbo Liu}
\author[7]{Xiaomin Yu}
\author[8]{Yu Tian}
\author[1]{Wentao Wang}
\author[9]{Zhenjie Yang}
\author[10]{Jian Hu}
\author[10]{Yunhang Shen}
\author[10]{Haoyu Cao}
\author[1,\mbox{\Letter}]{Xue Yang}

\affiliation[1]{Shanghai Jiao Tong University}
\affiliation[2]{Shanghai Innnovation Institute}
\affiliation[3]{Inner Mongolia University of Technology}
\affiliation[4]{Fudan University}
\affiliation[5]{Beijing Institute of Technology}
\affiliation[6]{Tongji University}
\affiliation[7]{The Hong Kong University of Science and Technology (Guangzhou)}
\affiliation[8]{Tsinghua University}
\affiliation[9]{The University of Hong Kong}
\affiliation[10]{Tencent Youtu Lab}

\contribution[\dag]{Project Lead}
\contribution[*]{Core Contributors}
\contribution[\mbox{\Letter}]{Corresponding Author}

\abstract{Omni LLMs can naturally locate key audio and visual evidence and integrate information across modalities for audiovisual reasoning. Existing audio-visual reasoning benchmarks that mainly rely on multiple-choice questions usually score models solely based on answer correctness, making it difficult to assess whether a model truly forms a reasoning chain from evidence to the correct answer. In addition, textual or unimodal shortcuts may allow models to answer correctly without analyzing the necessary audio-visual evidence, further weakening the reliability of evaluation benchmarks. To address these issues, we introduce \textbf{EvaOmni}, an evidence-grounded benchmark containing 392 videos and 1,027 purely human-annotated question-answer pairs, covering 15 capabilities and three levels: retrieval and localization, basic understanding, and complex reasoning. We design a series of rigorous annotation procedures in which experts annotate audio and visual evidence and construct video-based distractors; humans review whether questions meet strict inspection criteria, and finally models filter textual and unimodal shortcuts. Beyond high-quality question-answer pairs, we require models to simultaneously return the answer and supporting evidence intervals in order to evaluate the complete reasoning chain. We propose the Coverage Grounded Score (C.G.) to measure how much annotated evidence a model identifies, and the Bottleneck Grounded Score (B.G.) to penalize poor temporal localization in either modality; both are incorporated into answer correctness. Across 7 open-weight and 17 proprietary Omni LLMs, Gemini3.7-Flash achieved the highest overall accuracy (79.4\%), C.G. (70.8\%), and B.G. (35.1\%). At Level 3, it reached 79.1\% accuracy, but its B.G. was only 12.3\%. These results show that even the most advanced models still have room for development in grounded audio-visual reasoning.

}

\date{\today}

\checkdata[Project Page]{\url{https://xiaoxing2001.github.io/EvaOmni.github.io}}

\checkdata[Code]{\url{https://github.com/VisionXLab/EvaOmni}}
\checkdata[ModelScope]{\url{https://modelscope.cn/datasets/VisionXLab/EvaOmni-Dataset}}
\checkdata[Hugging Face]{\url{https://huggingface.co/datasets/VisionXLab/EvaOmni-Dataset}}

\renewcommand{\topfraction}{0.9}      \renewcommand{\bottomfraction}{0.8}   \renewcommand{\textfraction}{0.07}    \renewcommand{\floatpagefraction}{0.85} 
\begin{document}
\maketitle
\printprojectlinks

\section{Introduction}

Omni-modal large language models (Omni LLMs) \citep{Qwen2.5-Omni, Qwen3-Omni, Qwen3.5-Omni,nemotron3omni} can jointly process text, audio, images, and video, integrating cross-modal cues to interpret multimodal inputs and reason over the evidence they provide. Audio-visual reasoning in video scenarios is particularly difficult because it requires models to connect and analyze auditory and visual signals simultaneously. This involves identifying relevant evidence in each modality and integrating cues from different moments, abilities that support applications such as sports event understanding and the analysis of films and television programs. With the rapid development of Omni LLMs, accurately and reliably evaluating the capabilities of existing Omni LLMs has become particularly important, and is of great significance for identifying model shortcomings and driving their development.

\begin{figure}[t!]
    \centering
    \includegraphics[width=1.0\textwidth]{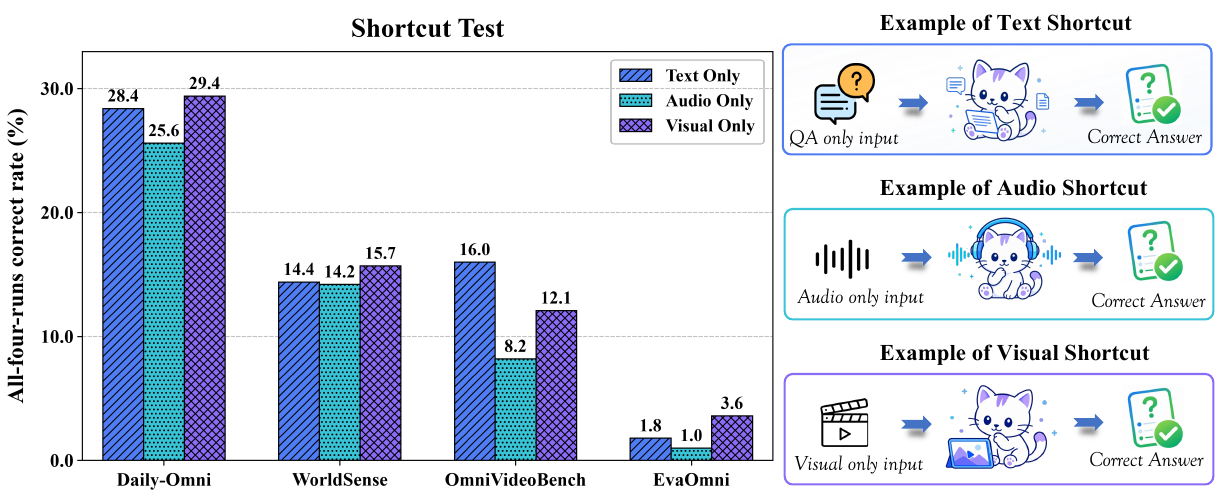}
    \caption{Existing audio-visual reasoning benchmarks suffer from shortcut issues, which weaken the reliability of benchmark evaluation. EvaOmni greatly reduces this phenomenon through a series of rigorous annotation procedures and a grounded evaluation protocol.}
    \label{fig:shortcut}
\end{figure}

Existing benchmarks for joint audio-visual reasoning \citep{DailyOmni, WorldSense, Omnivideobench,jointavbench,lvomnibench,socialomni} cover everyday activities, real-world scenarios, and diverse tasks, providing valuable ways to assess Omni LLMs. However, our analysis reveals a key limitation of dominant multiple-choice question (MCQ) evaluations: \textbf{a correct answer does not necessarily demonstrate joint use of audio and visual evidence}. As illustrated in Fig.~\ref{fig:shortcut}, some questions can be answered using a single modality or even textual cues alone. This ambiguity has two sources. First, answer-only scoring cannot distinguish evidence-supported answers from lucky guesses. Second, question and answer-option design can introduce unimodal or logical (text) shortcuts. For example, the correct option may be inferred from common knowledge, audio-only input, or vision-only input. Quantitative visualizations from existing benchmarks are provided in the Appendix~\ref{app:shortcut_evaluation}. Together, these limitations motivate an evaluation protocol that requires models to identify the audio and visual evidence supporting each answer, enabling a complete evaluation of the entire reasoning chain.

Motivated by these observations, we introduce \textbf{EvaOmni}, an \textbf{EV}idence-grounded benchmark for \textbf{A}udio-visual reasoning. It comprises \textbf{392 videos and 1,027 question-answer pairs} spanning \textbf{15 fine-grained capabilities} at three levels of reasoning and evidence complexity: \textbf{Level 1-Retrieval \& Grounding} requires locating relevant audio and visual evidence; \textbf{Level 2-Basic Understanding} requires interpreting and combining complementary cues; and \textbf{Level 3-Complex Reasoning} requires understanding with complex evidence distribution, or performing multi-step cross-modal reasoning. Experts need to annotate the audio and visual intervals supporting each answer, and models must return both an answer and its supporting intervals. This design both greatly reduces annotation errors and prevents models from using shortcuts when answering questions. To ensure question quality, we designed a series of rigorous annotation procedures: systematic video selection, expert annotation, cross-checking, and model-assisted filtering. These procedures help reduce text and unimodal shortcuts. Such procedures ensure the high quality and reliability of our questions. We also propose two new evaluation metrics: the \textbf{Coverage Grounded Score (C.G.)}, which measures the coverage of annotated evidence identified by the model, and the \textbf{Bottleneck Grounded Score (B.G.)}, which penalizes inaccurate temporal localization in either modality. Both incorporate answer correctness to comprehensively evaluate the model's reasoning ability. Additional grounded metrics are reported in the Appendix~\ref{app:additional_grounded_results}.
\begin{figure}[t!]
    \centering
    \includegraphics[width=1.0\textwidth]{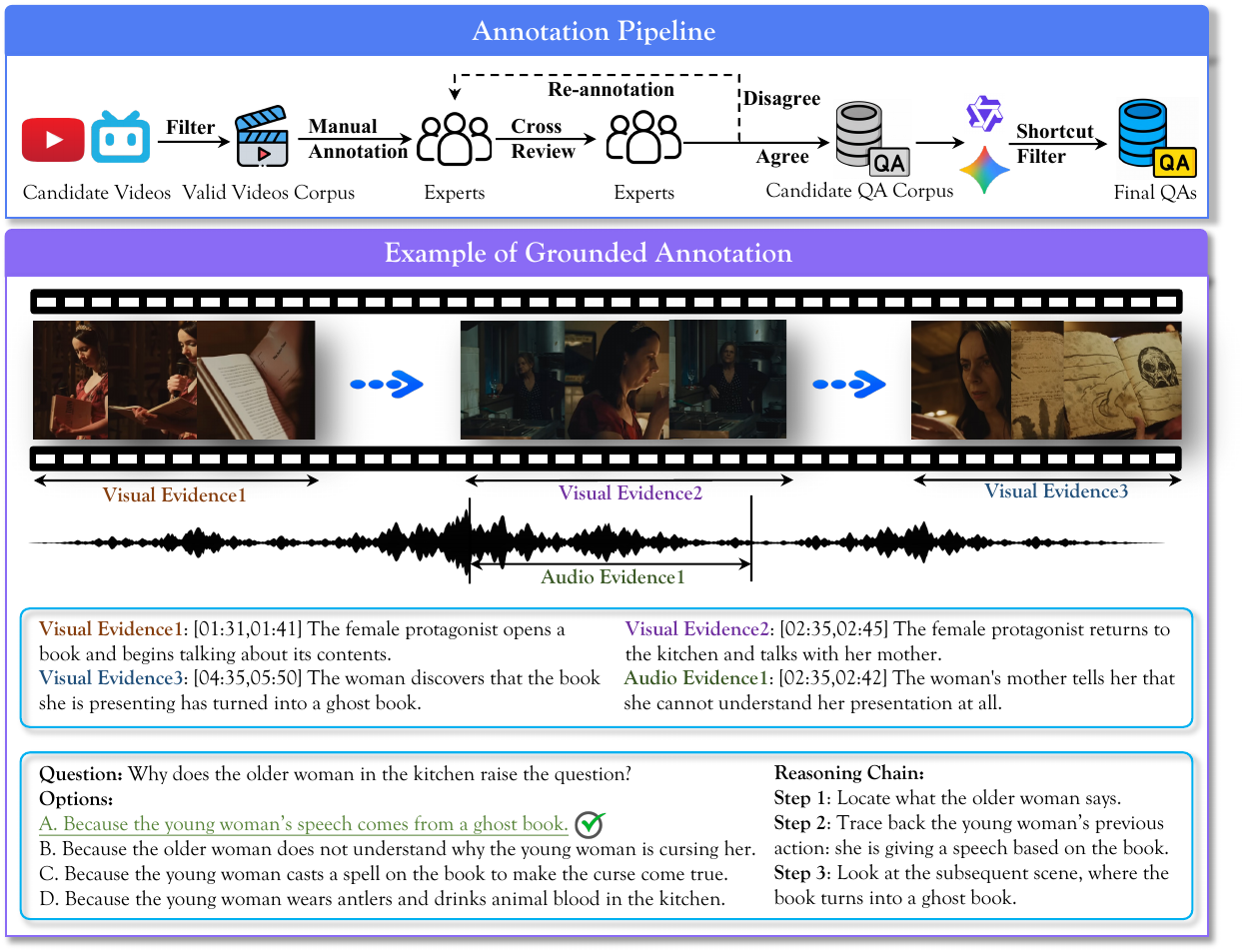}
\caption{Top: We designed a series of rigorous annotation procedures to ensure the quality and reliability of EvaOmni questions. Bottom: An example of Grounded Annotation, where, building on MCQ, we require models to output the audio and visual evidence intervals supporting the answer, in order to evaluate the complete reasoning chain and prevent blind guessing and shortcuts.}
    \label{fig:evaomni_framework}
    \vspace{-3mm}
\end{figure}
We conduct a systematic evaluation of EvaOmni, involving seven open-weight models and 17 proprietary Omni LLMs. The best-performing proprietary model, Gemini3.7-Flash, achieves 79.4\% accuracy, 70.8\% C.G., and 35.1\% B.G. overall. The strongest open-weight model, MiMo-V2.5-310B-A15B, achieves 56.8\% accuracy, 48.4\% C.G., and 20.5\% B.G. Gemini3.7-Flash leads by 22.6, 22.4, and 14.6 percentage points, respectively. Relative to Qwen3-Omni-Instruct-30B-A3B (25.2\% accuracy, 4.0\% C.G., and 0.4\% B.G.), MiMo-V2.5-310B-A15B scores 2.3, 12.1, and 51.3 times as high on these metrics, respectively. Despite this progress, the strongest proprietary models remain well below the human reference in localization quality: Gemini3.7-Flash achieves 35.1\% B.G., compared with 94.3\% for human experts. These findings demonstrate the importance of grounded evaluation, which cannot be captured by previous MCQ-based benchmarks. We hope EvaOmni brings some new insights to the development of omni-modal models: future models should not only pursue correct answers but also be able to precisely localize the evidence supporting their answers, so that their conclusions can be verified against the audio-visual input. We have publicly released EvaOmni, including its data, annotations, and evaluation framework, to support \textbf{reliable, interpretable, and hierarchical} evaluation of audio-visual reasoning. 

We summarize our contributions as follows:
\begin{itemize}[leftmargin=0.5cm]
\item We \textbf{review existing audio-visual reasoning benchmarks} and analyze their limitations. We find that most MCQ-based evaluation methods cannot fully and objectively assess whether a model actually uses the evidence in the audio and video; on the other hand, MCQ itself suffers from certain shortcut problems. These issues all affect the reliability of evaluation results.
\item We \textbf{introduce EvaOmni}, an evidence-grounded audio-visual reasoning benchmark. EvaOmni contains 392 videos and 1,027 purely human-annotated question-answer pairs, covering 15 fine-grained capabilities and organized into 3 levels of reasoning and evidence complexity. We also propose two new evaluation metrics: Coverage Grounded Score (C.G.) and Bottleneck Grounded Score (B.G.), which incorporate answer correctness to comprehensively evaluate the model's reasoning ability.
\item We \textbf{conduct a large-scale evaluation} and find that, for both open-source and closed-source models, there remains a gap between answering a question correctly and fully solving it. The state-of-the-art model Gemini3.7-Flash achieves 79.4\% accuracy on EvaOmni, but its grounded scores are only 70.8\% C.G. and 35.1\% B.G. This indicates that models are deficient in the complete logical chain of discovering evidence, utilizing evidence, and making correct judgments.
\end{itemize}
\section{Related Work}

\textbf{Omni LLMs.} Omni LLMs process text, images, video, and audio, enabling cross-modal understanding and reasoning~\citep{vita,vita1.5,ola}. ImageBind~\citep{imagebind} learned shared representations across text, audio, depth, and images; PandaGPT~\citep{pandagpt} connected such representations to an LLM; and NExT-GPT~\citep{Next-gpt} extended omni-modal systems to image, audio, and video generation. GPT-4o~\citep{gpt4o} introduced streaming speech generation for natural interaction. The Gemini~\citep{Gemini,gemini1.5,Gemini2.5} and Qwen~\citep{Qwen2.5-Omni,Qwen3-Omni,Qwen3.5-Omni} families are leading proprietary and open-weight Omni LLMs, respectively, advancing cross-modal understanding and reasoning. SeedRealtime~\citep{seedrealtime} further integrates audio, video, and text in a full-duplex architecture for continuous scene understanding, interaction-timing awareness, and responsive behavior. Together, these systems mark a shift toward joint reasoning and natural interaction.

\textbf{Benchmarks for Omni LLMs.} Unlike benchmarks of individual image, audio, or video understanding~\citep{mmau,mmsu,mmstar,mmmu,lvbench,mlvu}, Omni LLM benchmarks assess joint cross-modal reasoning. DailyOmni~\citep{DailyOmni} covers everyday scenarios; WorldSense~\citep{WorldSense} emphasizes audio-visual understanding in real-world videos; and OmniVideoBench~\citep{Omnivideobench} requires reasoning over both modalities. Yet text and unimodal shortcuts can undermine answer-only scores, and existing benchmarks do not jointly score answer correctness and the localization of supporting audio-visual evidence. Another evaluation approach, Open-Ended, relies on judge models and is thus insufficiently objective and convenient. Therefore, the research community currently urgently needs an evaluation system that is objective, convenient, and rigorous.

\section{EvaOmni: A New Evidence-Driven Paradigm for Omni-Modal Evaluation}

\subsection{Rethinking Audio-Visual Reasoning Benchmarks}

Before introducing EvaOmni, let us revisit the existing audio-visual reasoning benchmarks. Table~\ref{tab:omni_benchmark_comparison} compares seven representative benchmarks.
\begin{center}
  \centering
  \begin{minipage}[t]{0.39\textwidth}
    \vspace{0pt}
    \centering
    \includegraphics[width=\linewidth]{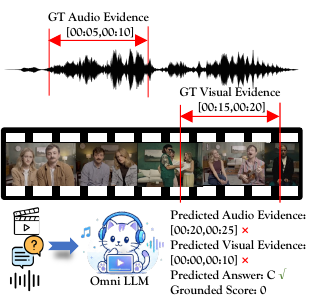}
    \vspace{-8mm}
    \captionof{figure}{Grounded evaluation.}
    \label{fig:evidence_grounded_evaluation}
  \end{minipage}\hfill
  \begin{minipage}[t]{0.61\textwidth}
    \vspace{0pt}
    \captionof{table}{\textbf{Comparison of omni-modal benchmark designs.}
    We compare shortcut checks, evidence annotation, grounded evaluation,
    and task taxonomies. Grounded Ann. requires question-level audio and visual
    evidence intervals; Grounded Eval. scores answer correctness and evidence
    grounding.}
    \label{tab:omni_benchmark_comparison}
    \centering
    \begingroup
    \renewcommand{\arraystretch}{1.16}
    \setlength{\tabcolsep}{3.2pt}
    \resizebox{\linewidth}{!}{\begin{tabular}{@{}lcccc@{}}
        \hline
        \multirow{2}{*}{\textbf{Benchmark}} &
        \multicolumn{2}{c}{\textbf{Reliability Control}} &
        \multicolumn{2}{c}{\textbf{Evaluation Design}} \\
        \cline{2-5}
        & \textbf{Shortcut checks} &
        \textbf{Grounded Ann.} &
        \textbf{Grounded Eval.} &
        \textbf{Taxonomy} \\
        \hline
        OmniBench & T/I/Tr/H & No & No & 3 cat./8 tasks\\
        Daily-Omni & T/H & No & No & 6 tasks \\
        WorldSense & V/H & No & No & 3 levels/26 tasks \\
        OmniVideoBench & T/U/H & No & No & 13 tasks \\
        JointAVBench & D/H & No & No & 5 dims./15 tasks\\
        LVOmniBench & T/A/V/H & No & No & 3 diff./9 types \\
        SocialOmni & -- & No & No & 3 tasks\\
        \hline
        \textbf{EvaOmni (ours)} & \textbf{T/A/V/H} & \textbf{Yes} & \textbf{Yes} & \textbf{3 levels/15 caps.} \\
        \hline
      \end{tabular}}
    \endgroup
  \end{minipage}
  \vspace{-0.6\baselineskip}
\end{center}

In the shortcut-checks column, T/A/V/I/Tr denote text-, audio-, video-, image-, and transcript-based tests; U denotes an unspecified unimodal test, D a dependency and logic check, and H human shortcut review. In the taxonomy column, cat., dims., diff., and caps. denote categories, dimensions, difficulty tiers, and capabilities. OmniBench uses static images, while SocialOmni focuses on social interaction rather than video question answering. Most existing benchmarks use multiple-choice questions (MCQs) and score responses by answer correctness. This raises \textbf{RQ1: Can answer-only MCQ accuracy establish whether a correct answer is grounded in both audio and visual evidence?} It cannot: the same credit is assigned to evidence-supported answers and unsupported correct choices. For four-option questions, uniform random guessing yields an expected accuracy of 25\%, illustrating one way to obtain correct answers without using either modality. Answer accuracy also provides no measure of whether a model can locate the evidence supporting its prediction. We also ask \textbf{RQ2: To what extent can existing questions be answered through textual, logical, or unimodal shortcuts?} We evaluate three representative video question-answering benchmarks---Daily-Omni, WorldSense, and OmniVideoBench---under text-, audio-, and video-only inputs. Figure~\ref{fig:shortcut} illustrates shortcut cases; the full protocol and aggregate results appear in Appendix~\ref{app:shortcut_evaluation}. We found that a considerable proportion of questions in existing benchmarks can still be answered correctly without complete audio-visual input. For example, in WorldSense, 14.4\% of questions can be answered correctly by looking only at the question and options, 14.2\% can be answered correctly by listening only to the audio, and 15.7\% can be answered correctly by watching only the video. Note that this does not represent a strict shortcut ratio (it depends on our test models), but it still indicates that many shortcut problems exist in existing benchmarks. These observations motivate \textbf{RQ3: How can we build a benchmark that retains the objectivity and convenience of MCQs while effectively and reliably evaluating model performance in audio-visual reasoning?} We address this question with the following EvaOmni.

\subsection{Introducing EvaOmni}

To address RQ3, we introduce \textbf{EvaOmni}, an evidence-grounded benchmark for audio-visual reasoning. For each question, expert annotators identify the audio and visual intervals supporting the correct answer. At evaluation time, models must provide both an answer and timestamped evidence intervals for each modality. EvaOmni evaluates answer correctness together with evidence coverage and temporal grounding, thereby limiting the credit awarded to correct answers that lack matching evidence. Text-only, audio-only, and vision-only checks during question construction further reduce shortcut susceptibility, as described below. EvaOmni organizes 15 fine-grained capabilities into three levels of reasoning and evidence complexity:

\textbf{Level~1: Retrieval and Grounding.}
Tests temporal alignment, event binidng, and entity binding, such as identifying the speaker of an utterance or linking a sound to its visible source.

\textbf{Level~2: Basic Understanding.}
Requires models to interpret and integrate complementary audio-visual cues to understand events, scene context, and their relationships. 

\textbf{Level~3: Complex Reasoning.}
Tests long-range reasoning, multi-evidence chains, and multi-hop reasoning. These tasks require models to integrate evidence from different moments, perform multiple inference steps, or both.

The three levels contain 3, 9, and 3 capabilities, respectively, as summarized in Fig.~\ref{fig:distribution}(a). The detailed capability distribution is provided in the Appendix~\ref{app:capability_taxonomy}.

\subsection{Data Collection }

To mitigate potential training data contamination, we collect recent videos from open platforms such as YouTube and Bilibili. The selected 1--12-minute videos span diverse real-world scenarios and contain informative visual events and audio cues. We exclude videos with hard-coded subtitles or prominent textual overlays to limit OCR-based shortcuts. Detailed selection and screening criteria are provided in Appendix~\ref{app:dataset_construction}.

\begin{figure}[t!]
    \centering
    \includegraphics[width=1.0\textwidth]{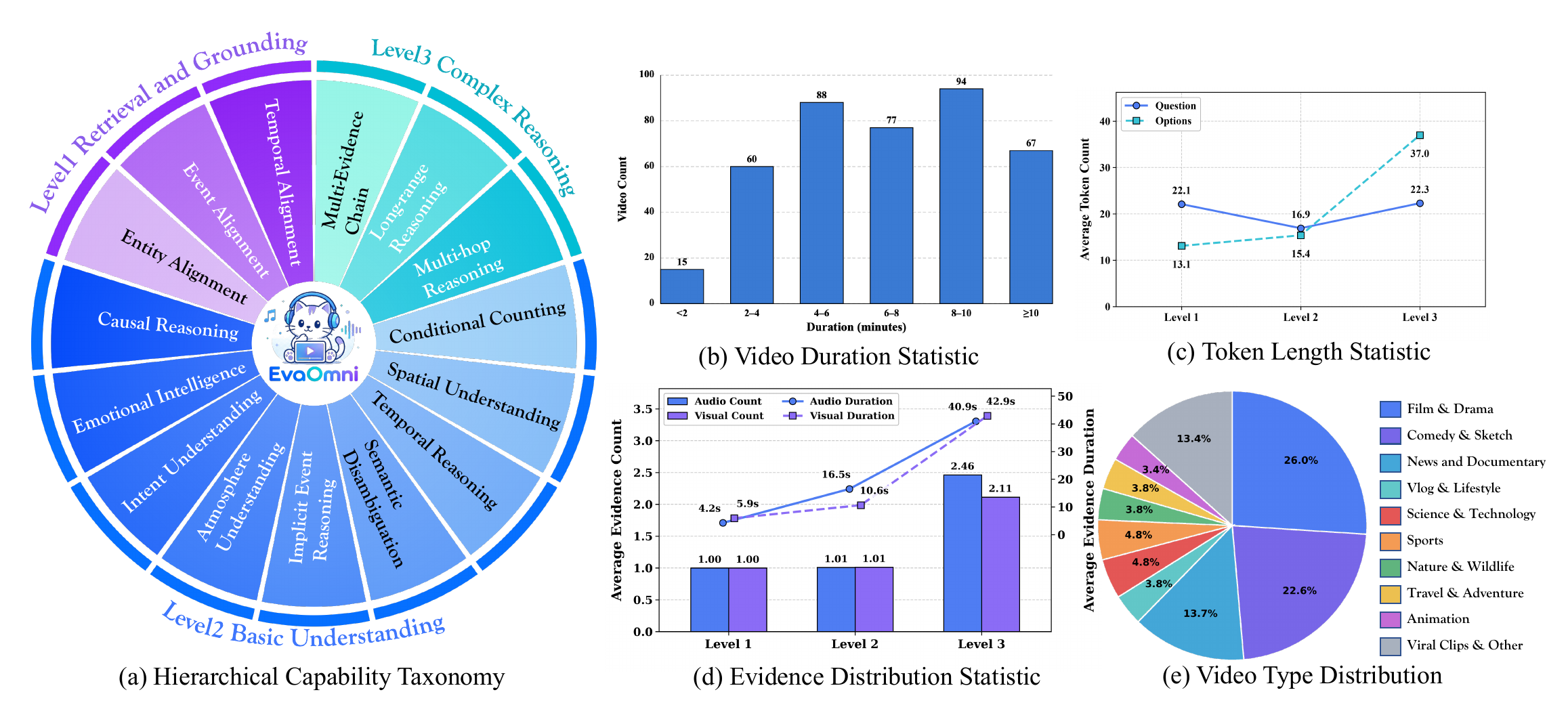}
    \caption{\textbf{Overview and statistics of EvaOmni.}
    (a) Hierarchical capability taxonomy.
    (b) Video duration statistics.
    (c) Mean question and answer-option lengths in tokens by level.
    (d) Mean counts and durations of annotated audio and visual evidence segments.
    (e) Video type distribution.}
    \label{fig:distribution}
\end{figure}

\subsection{Annotation Pipeline}

Expert annotators construct questions requiring joint use of audio and visual information and mark the temporal evidence for each modality. Each sample then undergoes human cross-validation for capability alignment, evidence quality, and potential textual or unimodal shortcuts. Disputed pairs receive independent review by two additional experts and are discarded unless both approve them. Model-assisted screening under text-only, audio-only, and visual-only inputs identifies additional shortcuts; pairs flagged at this stage are revised and re-evaluated. Appendix~\ref{app:dataset_construction} provides the full annotation and quality-control procedure.

\subsection{Evaluation Metrics}
\label{sec:metrics}
EvaOmni adopts a composite evaluation metric that simultaneously assesses a model's answer correctness and evidence localization ability. We primarily use answer accuracy (\textbf{Acc.}) as well as additional grounded metrics, which are introduced below. For each question, the model returns an answer and audio and visual evidence intervals specified by start and end timestamps. We report two grounded scores that combine the answer score $s$ ($s=1$ for a correct option and $s=0$ otherwise) with evidence localization quality:

\textbf{Coverage Grounded Score (C.G.).}
This metric measures how much annotated audio and visual evidence is
covered by the predictions. Let $\mathrm{audio}_{I}$ and $\mathrm{visual}_{I}$ be the
numbers of human-annotated audio and visual evidence segments that have a
positive-duration temporal overlap with at least one predicted segment in the
same modality. Each annotated segment is counted at most once. Let
$\mathrm{audio}_{\mathrm{all}}$ and $\mathrm{visual}_{\mathrm{all}}$ denote the
total numbers of annotated segments in the respective modalities. We define
\begin{equation}
\mathrm{C.G.} = \frac{s}{2}\left(
\frac{\mathrm{audio}_{I}}{\mathrm{audio}_{\mathrm{all}}}
+ \frac{\mathrm{visual}_{I}}{\mathrm{visual}_{\mathrm{all}}}
\right).
\label{eq:coverage_grounded_score}
\end{equation}
C.G. averages the two modality-specific evidence hit rates and weights them
by answer correctness. We interpret C.G. as a lenient evidence hit-rate metric,
not a measure of precise temporal localization: any positive-duration overlap
counts as a hit, and overly broad predictions are not directly penalized for
imprecise interval boundaries.

\textbf{Bottleneck Grounded Score (B.G.).}
Let $\operatorname{IoU}_{\mathrm{audio}}$ and
$\operatorname{IoU}_{\mathrm{visual}}$ denote the temporal intersection over
union between predicted and human-annotated evidence for the audio and visual
modalities, respectively. For each modality, we first take the union of all
predicted intervals and the union of all annotated intervals; their IoU is the
duration of the intersection of these two sets divided by the duration of
their union. We define
\begin{equation}
\mathrm{B.G.} = s \times \min\!\left(
\operatorname{IoU}_{\mathrm{audio}},
\operatorname{IoU}_{\mathrm{visual}}\right).
\label{eq:bottleneck_grounded_score}
\end{equation}
Taking the minimum makes B.G. sensitive to the less accurately localized
modality: strong localization in one modality cannot compensate for poor
localization in the other. Thus, C.G. measures broad evidence coverage,
whereas B.G. requires accurate localization in both modalities. Additional
grounded metrics are provided in Appendix~\ref{app:additional_grounded_results}.

\section{Experiments}
\begingroup
\newcommand{\mainresultafter}{\begin{minipage}{\linewidth}
    \centering
    \begin{minipage}[t]{0.32\textwidth}
        \vspace{0pt}
        \centering
        \includegraphics[width=\linewidth]{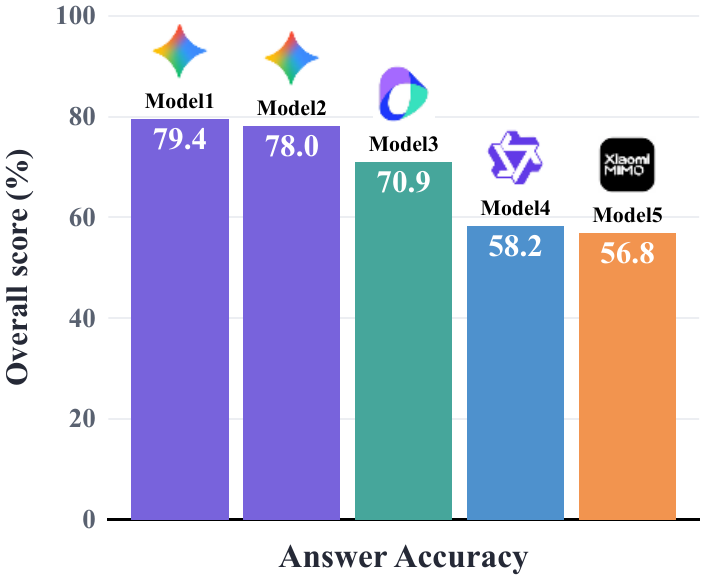}
    \end{minipage}\hfill
    \begin{minipage}[t]{0.32\textwidth}
        \vspace{0pt}
        \centering
        \includegraphics[width=\linewidth]{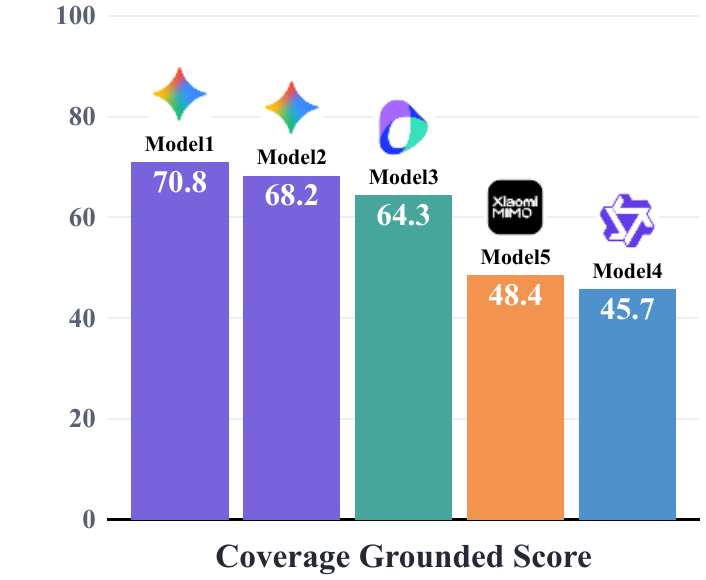}
    \end{minipage}\hfill
    \begin{minipage}[t]{0.32\textwidth}
        \vspace{0pt}
        \centering
        \includegraphics[width=\linewidth]{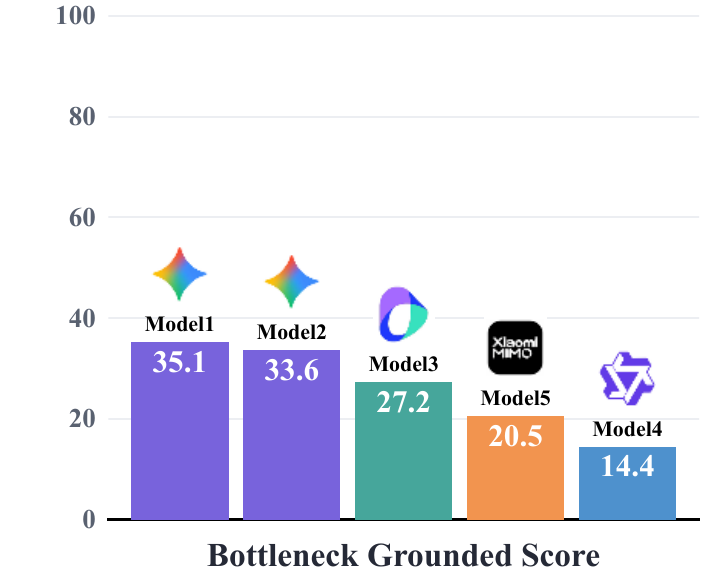}
    \end{minipage}
    \captionof{figure}{\textbf{Overall performance rankings for selected models.}
    We select the top two Gemini models and the top Doubao, Qwen, and
    MiMo models by overall Acc. in Table~\ref{tab:main_results}, then rank
    the same five models independently by Acc., C.G., and B.G.
    Model1: Gemini3.7-Flash; Model2: Gemini3.8-Flash;
    Model3: Doubao-Seed-2.1-Lite; Model4: Qwen3.8-Omni-Flash;
    Model5: MiMo-V2.5-310B-A15B.
    Model IDs remain fixed across panels, which share the same score scale.}
    \label{fig:overall_rankings}
\end{minipage}
}
\begin{table}[!tp]
  \centering
  \caption{\textbf{Overall performance on EvaOmni.} We report Acc., C.G., and B.G. for each capability level and overall. All values are in \%. The best results are shown in \textbf{bold}, and the second-best results are \underline{underlined.}}
  \label{tab:main_results}
  \vspace{0.4em}
  \begingroup
  \newcommand{\modelicon}[1]{\makebox[1.35em][c]{\raisebox{-0.15em}{\includegraphics[width=1.05em,height=1.05em,keepaspectratio]{figures/content/evaomni/#1}}}\hspace{0.2em}}
  \newcommand{\modeliconspace}{\hspace{1.55em}}
  \renewcommand{\arraystretch}{1.10}
  \setlength{\tabcolsep}{4.0pt}
  \resizebox{\textwidth}{!}{\begin{tabular}{@{}l*{4}{c>{\columncolor[RGB]{235,245,255}}c>{\columncolor[RGB]{220,235,250}}c}@{}}
      \hline
      \multirow{2}{*}{\textbf{Model}} &
      \multicolumn{3}{c}{\textbf{Level1 R.\&G.}} &
      \multicolumn{3}{c}{\textbf{Level2 B.U.}} &
      \multicolumn{3}{c}{\textbf{Level3 C.R.}} &
      \multicolumn{3}{c}{\textbf{Overall}} \\
      \cline{2-13}
      &
      \textbf{Acc.} & \multicolumn{1}{c}{\textbf{C.G.}} & \multicolumn{1}{c}{\textbf{B.G.}} &
      \textbf{Acc.} & \multicolumn{1}{c}{\textbf{C.G.}} & \multicolumn{1}{c}{\textbf{B.G.}} &
      \textbf{Acc.} & \multicolumn{1}{c}{\textbf{C.G.}} & \multicolumn{1}{c}{\textbf{B.G.}} &
      \textbf{Acc.} & \multicolumn{1}{c}{\textbf{C.G.}} & \multicolumn{1}{c}{\textbf{B.G.}} \\
      \hline
      \multicolumn{13}{c}{\textcolor[gray]{0.45}{\textbf{\rmfamily\itshape Open-Weight Omni LLMs}}} \\
      \hline
      \modelicon{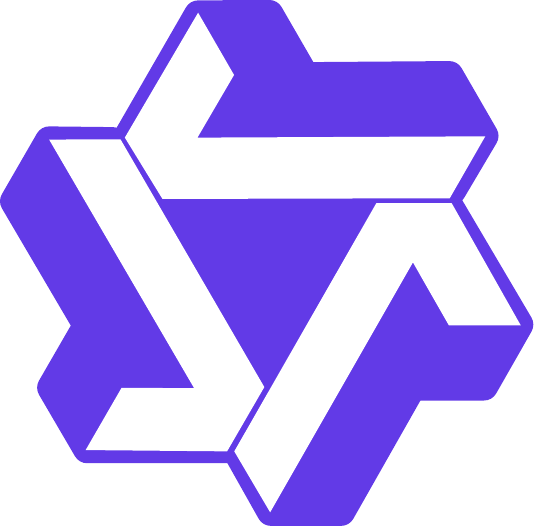}Qwen2.5-Omni-7B & 16.6 & 1.2 & 0.6 & 17.4 & 0.5 & 0.0 & 13.7 & 0.3 & 0.0 & 16.4 & 0.8 & 0.3 \\
      \modelicon{qwen-icon-seeklogo.pdf}Qwen3-Omni-Instruct-30B-A3B & 28.6 & 5.1 & 0.6 & 22.4 & 3.4 & 0.4 & 22.2 & 2.4 & 0.1 & 25.2 & 4.0 & 0.4 \\
      \modelicon{qwen-icon-seeklogo.pdf}Qwen3-Omni-Thinking-30B-A3B & 26.3 & 11.4 & 4.8 & 24.4 & 6.2 & 1.2 & 21.6 & 4.1 & 0.6 & 24.8 & 8.3 & 2.8 \\
      \modelicon{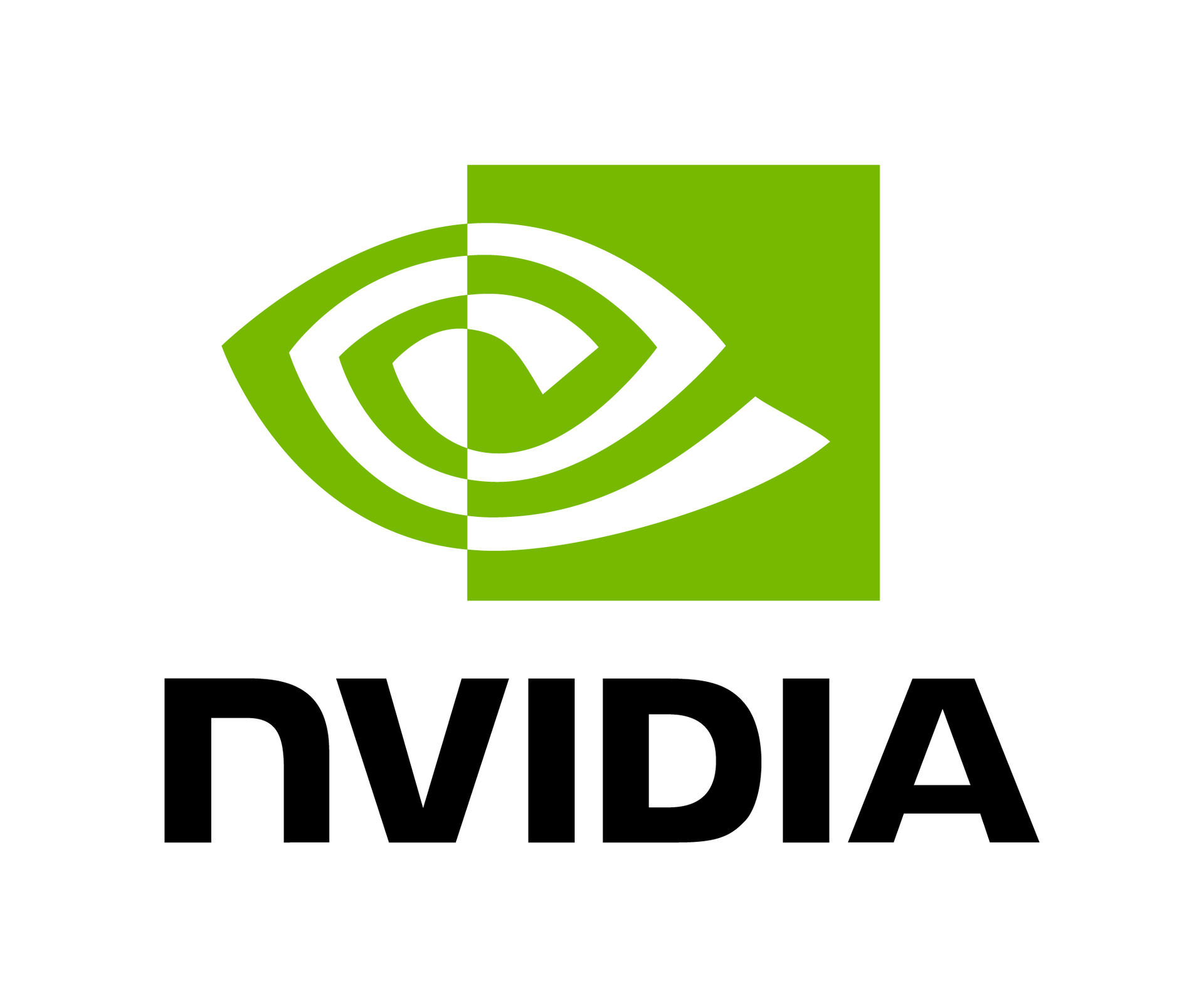}Nemotron3-Nano-Omni-30B-A3B & 30.7 & 15.0 & 5.3 & 22.1 & 13.1 & 3.9 & 10.5 & 4.7 & 0.7 & 24.2 & 12.6 & 4.0 \\
      \modelicon{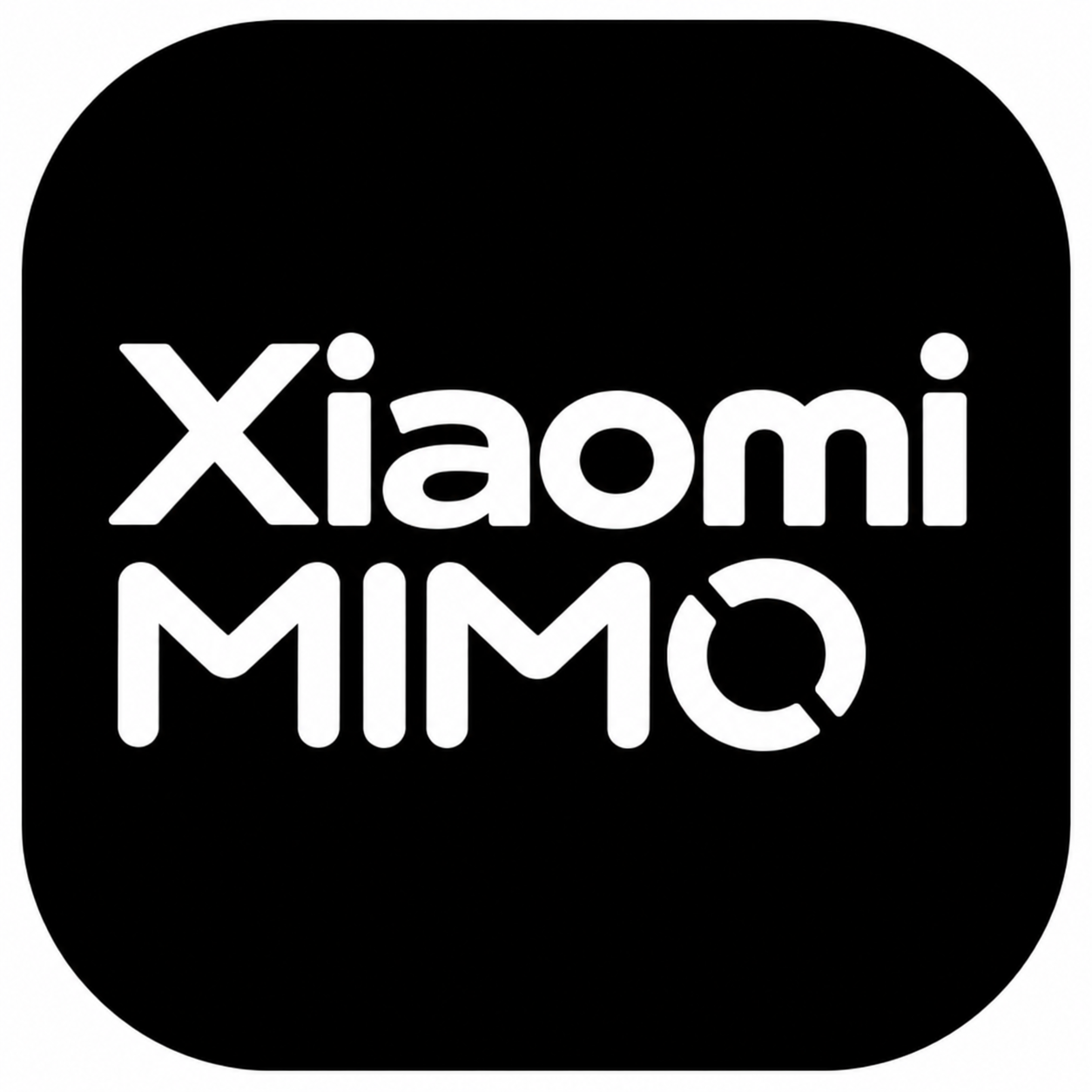}MiMo-V2.5-310B-A15B & \textbf{65.4} & \textbf{61.3} & \textbf{30.8} & \textbf{52.3} & \textbf{42.3} & \textbf{15.1} & 45.1 & \textbf{29.6} & \textbf{6.7} & \textbf{56.8} & \textbf{48.4} & \textbf{20.5} \\
      \modelicon{mimo_icon.png}MiMo-V2.6-Flash-310B-A15B & \underline{61.3} & \underline{55.3} & \underline{29.0} & \underline{48.6} & \underline{41.4} & \underline{14.5} & \underline{45.8} & \underline{28.6} & \underline{5.9} & \underline{53.6} & \underline{45.3} & \underline{19.3} \\
      \modelicon{mimo_icon.png}MiMo-V2.6-Pro-1T-A42B & 60.1 & 51.3 & 28.8 & 44.3 & 35.2 & 12.0 & \textbf{49.7} & 26.6 & 5.0 & 51.8 & 40.7 & 18.1 \\
      \hline
      \multicolumn{13}{c}{\textcolor[gray]{0.45}{\textbf{\rmfamily\itshape Proprietary Omni LLMs}}} \\
      \hline
      \modelicon{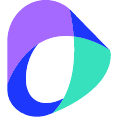}Doubao-Seed-2.0-Lite & 54.1 & 36.4 & 10.8 & 54.1 & 36.1 & 11.7 & 62.7 & 32.2 & 5.9 & 55.4 & 35.6 & 10.4 \\      
      \modelicon{doubao-icon-seeklogo.pdf}Doubao-Seed-2.1-Lite & 78.8 & 74.8 & 38.0 & 67.0 & 61.5 & 22.6 & 59.5 & 42.7 & 9.4 & 70.9 & 64.3 & 27.2 \\
      \modelicon{qwen-icon-seeklogo.pdf}Qwen3.5-Omni-Flash & 47.5 & 41.9 & 15.5 & 30.9 & 24.0 & 7.2 & 29.4 & 15.7 & 2.7 & 37.7 & 30.4 & 10.0 \\
      \modelicon{qwen-icon-seeklogo.pdf}Qwen3.5-Omni-Plus & 58.1 & 52.0 & 22.1 & 42.3 & 26.0 & 8.6 & 42.5 & 19.8 & 3.3 & 49.0 & 36.0 & 13.5 \\
      \modelicon{qwen-icon-seeklogo.pdf}Qwen3.8-Omni-Flash & 65.0 & 53.6 & 19.3 & 52.3 & 41.4 & 11.8 & 56.2 & 35.9 & 7.8 & 58.2 & 45.7 & 14.4 \\
      \modelicon{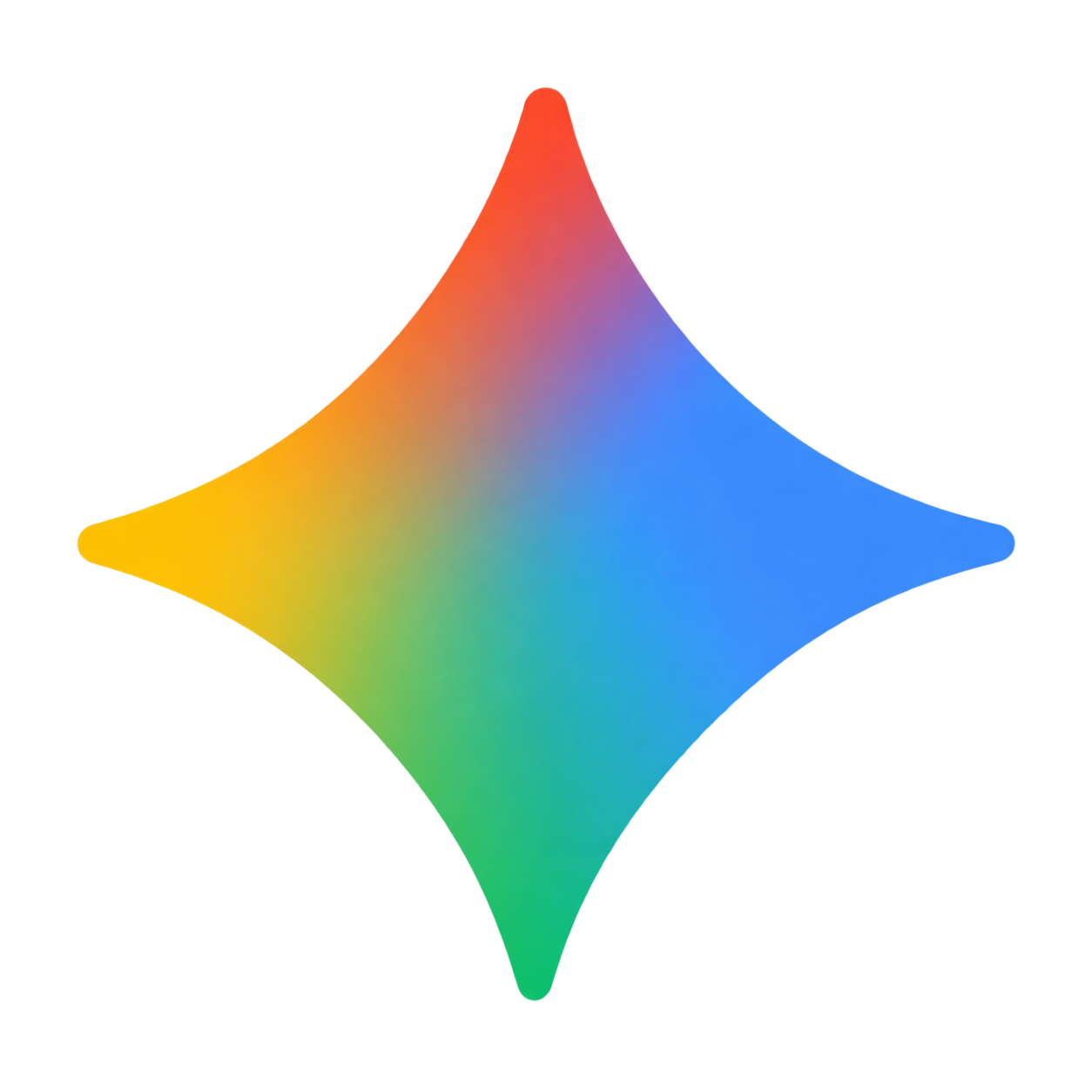}Gemini2.5-Flash & 59.2 & 55.2 & 26.7 & 39.3 & 31.4 & 11.8 & 41.8 & 24.6 & 4.0 & 48.1 & 40.4 & 16.9 \\
      \modelicon{gemini_icon.png}Gemini2.5-Flash-Lite & 53.0 & 47.2 & 22.1 & 30.9 & 22.4 & 8.0 & 27.5 & 14.8 & 2.2 & 39.7 & 31.8 & 13.1 \\
      \modelicon{gemini_icon.png}Gemini2.5-Pro & 76.5 & 73.8 & 42.3 & 65.0 & 55.0 & 22.5 & 72.5 & 45.8 & 9.4 & 71.0 & 61.6 & 28.9 \\
      \modelicon{gemini_icon.png}Gemini3.1-Flash-Lite & 59.4 & 52.8 & 24.2 & 36.6 & 30.7 & 10.5 & 32.0 & 19.0 & 3.0 & 45.6 & 38.3 & 15.2 \\
      \modelicon{gemini_icon.png}Gemini3.1-Flash-Lite-Preview & 59.7 & 53.6 & 25.2 & 39.3 & 31.1 & 10.3 & 38.6 & 23.2 & 3.3 & 47.8 & 39.4 & 15.6 \\
      \modelicon{gemini_icon.png}Gemini3.1-Pro-Preview & 74.7 & 72.8 & 41.5 & 68.6 & 59.8 & 24.8 & 76.5 & \underline{48.7} & 11.2 & 72.3 & 63.7 & 29.8 \\
      \modelicon{gemini_icon.png}Gemini3.5-Flash & 64.1 & 59.4 & 28.8 & 69.5 & 61.4 & \underline{27.4} & \underline{77.8} & 47.8 & \underline{11.3} & 68.5 & 58.6 & 25.6 \\
      \modelicon{gemini_icon.png}Gemini3.5-Flash-Lite & 60.6 & 55.6 & 26.5 & 38.2 & 27.9 & 10.7 & 42.5 & 24.2 & 3.5 & 48.3 & 39.1 & 16.3 \\
      \modelicon{gemini_icon.png}Gemini3.6-Flash & 80.9 & 78.0 & 44.7 & 68.4 & 61.3 & 26.1 & 73.2 & 44.2 & 10.6 & 74.4 & 65.8 & 31.7 \\
      \modelicon{gemini_icon.png}Gemini3.7-Flash & \textbf{85.3} & \textbf{84.0} & \textbf{49.1} & \underline{73.6} & \textbf{65.3} & \textbf{29.2} & \textbf{79.1} & \textbf{49.1} & \textbf{12.3} & \textbf{79.4} & \textbf{70.8} & \textbf{35.1} \\
      \modelicon{gemini_icon.png}Gemini3.8-Flash & \underline{83.4} & \underline{81.9} & \underline{48.4} & \textbf{73.9} & \underline{63.3} & \underline{27.4} & 74.5 & 43.1 & 9.5 & \underline{78.0} & \underline{68.2} & \underline{33.6} \\
      \modelicon{gemini_icon.png}Gemini4-Argon & 67.1 & 55.9 & 29.3 & 52.0 & 33.7 & 13.0 & 72.5 & 33.5 & 7.5 & 61.4 & 43.0 & 19.1 \\
      \hline
      \modelicon{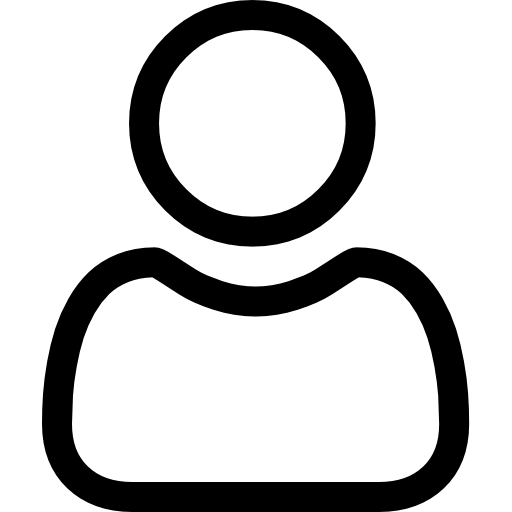}\textcolor{gray}{Human Expert} & \textcolor{gray}{97.8} & \textcolor{gray}{97.8} & \textcolor{gray}{95.8} & \textcolor{gray}{96.5} & \textcolor{gray}{94.8} & \textcolor{gray}{93.2} & \textcolor{gray}{94.4} & \textcolor{gray}{93.8} & \textcolor{gray}{93.0} & \textcolor{gray}{96.8} & \textcolor{gray}{96.0} & \textcolor{gray}{94.3} \\
      \hline
    \end{tabular}}
  \endgroup
  \ifdefined\begin{minipage}{\linewidth}
    \centering
    \begin{minipage}[t]{0.32\textwidth}
        \vspace{0pt}
        \centering
        \includegraphics[width=\linewidth]{figures/content/evaomni/overall_rankings_acc.pdf}
    \end{minipage}\hfill
    \begin{minipage}[t]{0.32\textwidth}
        \vspace{0pt}
        \centering
        \includegraphics[width=\linewidth]{figures/content/evaomni/overall_rankings_cg.pdf}
    \end{minipage}\hfill
    \begin{minipage}[t]{0.32\textwidth}
        \vspace{0pt}
        \centering
        \includegraphics[width=\linewidth]{figures/content/evaomni/overall_rankings_bg.pdf}
    \end{minipage}
    \captionof{figure}{\textbf{Overall performance rankings for selected models.}
    We select the top two Gemini models and the top Doubao, Qwen, and
    MiMo models by overall Acc. in Table~\ref{tab:main_results}, then rank
    the same five models independently by Acc., C.G., and B.G.
    Model1: Gemini3.7-Flash; Model2: Gemini3.8-Flash;
    Model3: Doubao-Seed-2.1-Lite; Model4: Qwen3.8-Omni-Flash;
    Model5: MiMo-V2.5-310B-A15B.
    Model IDs remain fixed across panels, which share the same score scale.}
    \label{fig:overall_rankings}
\end{minipage}

    \par\medskip
    \begin{minipage}{\linewidth}
    \centering
    \begin{minipage}[t]{0.32\textwidth}
        \vspace{0pt}
        \centering
        \includegraphics[width=\linewidth]{figures/content/evaomni/overall_rankings_acc.pdf}
    \end{minipage}\hfill
    \begin{minipage}[t]{0.32\textwidth}
        \vspace{0pt}
        \centering
        \includegraphics[width=\linewidth]{figures/content/evaomni/overall_rankings_cg.pdf}
    \end{minipage}\hfill
    \begin{minipage}[t]{0.32\textwidth}
        \vspace{0pt}
        \centering
        \includegraphics[width=\linewidth]{figures/content/evaomni/overall_rankings_bg.pdf}
    \end{minipage}
    \captionof{figure}{\textbf{Overall performance rankings for selected models.}
    We select the top two Gemini models and the top Doubao, Qwen, and
    MiMo models by overall Acc. in Table~\ref{tab:main_results}, then rank
    the same five models independently by Acc., C.G., and B.G.
    Model1: Gemini3.7-Flash; Model2: Gemini3.8-Flash;
    Model3: Doubao-Seed-2.1-Lite; Model4: Qwen3.8-Omni-Flash;
    Model5: MiMo-V2.5-310B-A15B.
    Model IDs remain fixed across panels, which share the same score scale.}
    \label{fig:overall_rankings}
\end{minipage}

  \fi
\end{table}

\endgroup
\subsection{Evaluated Models}

We evaluate 24 Omni LLMs on EvaOmni, comprising 7 open-weight and
17 proprietary models. The open-weight models include
Qwen2.5-Omni-7B~\citep{Qwen2.5-Omni}, Qwen3-Omni-Instruct and
Qwen3-Omni-Thinking~\citep{Qwen3-Omni},
Nemotron3-Nano-Omni~\citep{nemotron3omni}, MiMo-V2.5-310B-A15B~\citep{mimov2_5},
MiMo-V2.6-Flash-310B-A15B, and MiMo-V2.6-Pro-1T-A42B~\citep{mimo_2_6}. The proprietary
models include Qwen3.5-Omni-Flash and Qwen3.5-Omni-Plus~\citep{Qwen3.5-Omni},
Qwen3.8-Omni-Flash~\citep{qwen38omniflash}, Doubao-Seed-2.0-Lite~\citep{seed_2_1_lite}, Doubao-Seed-2.1-Lite~\citep{seed_2_1_lite}, and 12 Gemini variants
spanning the Pro, Flash, and Flash-Lite series~\citep{Gemini2.5,gemini_3_1pro,gemini_3_7flash,gemini_3_8flash,gemini_3_6falsh,gemini_4_argon}. The complete model list
is provided in Table~\ref{tab:main_results}. We also report human expert
performance as a reference for comparing answer correctness and evidence
grounding; its protocol is described in Appendix~\ref{app:human_baseline}.
\subsection{Experimental Settings}

We evaluate all 24 models on 1,027 question-answer pairs drawn from
392 videos, using the same prompt and preprocessed audio-visual content.
Video frames are uniformly sampled at 1.0 FPS and resized so that their
longer side is 480 pixels while preserving the aspect ratio. Open-weight
models outside the MiMo family are evaluated on a single server equipped
with eight NVIDIA H20 GPUs, using the inference frameworks and settings
recommended in their model cards. The open-weight MiMo models and all
proprietary models are evaluated through their providers' official APIs.
The prompt, input preprocessing procedures, and further evaluation
details are provided in the Appendix.
\subsection{Main Findings}

\begin{figure}[t!]
    \centering
    \begin{minipage}[t]{0.40\linewidth}
        \vspace{0pt}
        \centering
        \includegraphics[width=\linewidth]{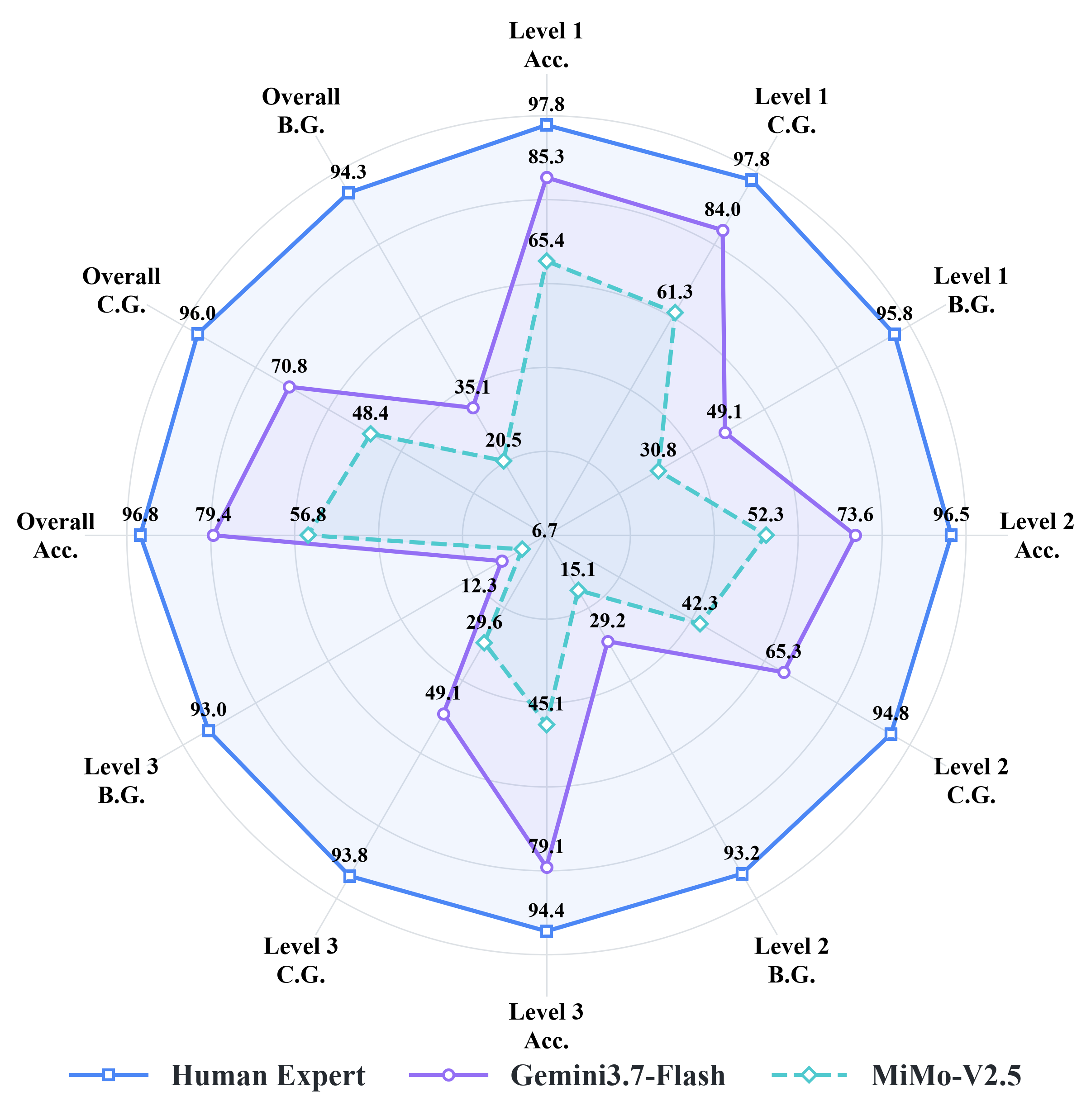}
        \captionof{figure}{Overall performance.}
        \label{fig:overall}
    \end{minipage}\hfill
    \begin{minipage}[t]{0.57\linewidth}
        \vspace{0pt}
        \begin{minipage}{\linewidth}
  \centering
  \small
  \captionof{table}{\textbf{Cost efficiency.} Cost efficiency is USD per unit of
  normalized score ($\downarrow$), computed from the Overall results in
  Table~\ref{tab:main_results}; the lowest value in each column is \textbf{bold}.}
  \label{tab:cost_efficiency}
  \begingroup
  \newcommand{\modelicon}[1]{\makebox[1.35em][c]{\raisebox{-0.15em}{\includegraphics[width=1.05em,height=1.05em,keepaspectratio]{figures/content/evaomni/#1}}}\hspace{0.2em}}
  \renewcommand{\arraystretch}{1.15}
  \setlength{\tabcolsep}{4.5pt}
  \resizebox{\linewidth}{!}{\begin{tabular}{@{}lrrrr@{}}
      \toprule
      \multirow{2}{*}{\textbf{Model}} & \multirow{2}{*}{\makecell{\textbf{Total cost}\\\textbf{(USD)}}} &
      \multicolumn{3}{c}{\textbf{Cost efficiency (USD) $\downarrow$}} \\
      \cmidrule(l){3-5}
      & & \textbf{Acc.} & \textbf{C.G.} & \textbf{B.G.} \\
      \midrule
      \modelicon{doubao-icon-seeklogo.pdf}Doubao-Seed-2.1-Lite & 13.71 & \textbf{0.0188} & \textbf{0.0208} & \textbf{0.0491} \\
      \modelicon{qwen-icon-seeklogo.pdf}Qwen3.8-Omni-Flash & 11.54 & 0.0193 & 0.0246 & 0.0780 \\
      \modelicon{gemini_icon.png}Gemini3.1-Flash-Lite & 18.67 & 0.0399 & 0.0475 & 0.1196 \\
      \modelicon{gemini_icon.png}Gemini3.7-Flash & 44.41 & 0.0545 & 0.0611 & 0.1232 \\
      \modelicon{qwen-icon-seeklogo.pdf}Qwen3.5-Omni-Flash & 41.90 & 0.1082 & 0.1342 & 0.4080 \\
      \modelicon{gemini_icon.png}Gemini3.5-Flash & 88.93 & 0.1264 & 0.1478 & 0.3382 \\
      \modelicon{gemini_icon.png}Gemini2.5-Flash & 67.68 & 0.1370 & 0.1631 & 0.3900 \\
      \modelicon{gemini_icon.png}Gemini3.1-Pro-Preview & 118.57 & 0.1597 & 0.1812 & 0.3874 \\
      \modelicon{qwen-icon-seeklogo.pdf}Qwen3.5-Omni-Plus & 130.55 & 0.2594 & 0.3531 & 0.9416 \\
      \modelicon{gemini_icon.png}Gemini2.5-Pro & 224.48 & 0.3079 & 0.3548 & 0.7563 \\
      \bottomrule
    \end{tabular}}
  \endgroup
  \par\smallskip
\end{minipage}

    \end{minipage}
\end{figure}

Table~\ref{tab:main_results} reports performance across capability levels and overall, while Figure~\ref{fig:overall} compares two representative models with human performance. We summarize the key findings below.

\textbf{Overall Performance Analysis.}
(1) \textit{A substantial gap remains between the strongest evaluated open-weight
and proprietary models.} MiMo-V2.5-310B-A15B achieves overall Acc., C.G., and B.G.
scores of 56.8\%, 48.4\%, and 20.5\%, respectively, whereas
Gemini3.7-Flash reaches 79.4\%, 70.8\%, and 35.1\%. The respective
gaps of 22.6, 22.4, and 14.6 percentage points span answer correctness,
evidence coverage, and precise dual-modal localization. (2) \textit{The gap to human experts is especially pronounced in precise grounding.}
Human experts achieve 96.8\% Acc., 96.0\% C.G., and 94.3\% B.G.,
exceeding Gemini3.7-Flash by 17.4, 25.2, and 59.2 percentage points,
respectively. The model's 79.4\% accuracy and 70.8\% C.G. indicate strong
answering performance and substantial evidence coverage, yet its 35.1\%
B.G. reveals persistent difficulty in precisely localizing evidence in
both modalities. Although grounded scores also incorporate answer correctness,
the much smaller decline from Acc. to B.G. for humans (2.5 points, versus
44.3 for Gemini3.7-Flash) highlights the remaining grounding challenge. (3) \textit{Grounding limitations are more severe among the four open-weight models
with fewer than 100B total parameters.} Their overall C.G. ranges from
0.8\% to 12.6\% and B.G. from 0.3\% to 4.0\%, compared with
30.4\%--70.8\% and 10.0\%--35.1\% for the proprietary models.
For example, Qwen2.5-Omni-7B achieves 16.4\% accuracy but only 0.8\%
C.G. and 0.3\% B.G. Jointly requiring correct
answers and grounded evidence thus exposes limitations that answer-only
evaluation can obscure. These models leave substantial room for improvement
in consistently satisfying both dual-modal localization and answer correctness. In Figure~\ref{fig:overall_rankings}, we compare the rankings of five models. Model rankings by grounding score broadly follow those by Acc., although a few shifts in position highlight differences in evidence localization among models with similar answer accuracy.

\begin{takeawaybox}
\textbf{Finding 1:}
Grounded evaluation shows that even leading models trail humans in evidence localization, while open-weight models under 100B parameters lag behind proprietary models in evidence perception and localization.
\end{takeawaybox}

\textbf{Analysis Across Capability Levels.}
Answer accuracy decreases from
Level~1 to Level~2 for 21 of the 24 models, but increases from Level~2 to
Level~3 for 13 models. For example, Gemini3.7-Flash improves from 73.6\%
to 79.1\% accuracy between Levels~2 and~3, while its C.G. drops from
65.3\% to 49.1\% and B.G. from 29.2\% to 12.3\%. This divergence is consistent with the design of Level~3, which emphasizes
long-range reasoning, combinations of multiple evidence segments, and
multi-hop inference. These tasks place greater demands on assembling and
localizing supporting evidence, without necessarily producing a monotonic
decline in multiple-choice accuracy. One plausible explanation is that
partial evidence can help eliminate distractors and narrow the candidate
set, allowing a model to select the correct answer without fully localizing
the supporting evidence. The aggregate results are compatible with this
explanation, but do not directly establish the models' answer-selection
strategies. In contrast, Level~3 yields the lowest B.G. for all 24 models, including
one tie, and the lowest C.G. for all 24 models. This broad trend across
open-weight and proprietary models highlights the value of grounded
evaluation: C.G. rewards evidence coverage alongside correct answers,
whereas B.G. rewards accurate temporal localization in both modalities
alongside correct answers. Reporting Acc. together with these two metrics
therefore provides a more complete assessment of answering, evidence
coverage, and precise grounding across capability levels.
\begin{takeawaybox}
\textbf{Finding 2:}
As the task level increases, evidence-grounding quality declines, while answer accuracy does not follow the same trend.
\end{takeawaybox}

\textbf{Cost-Efficiency Analysis.}
Table~\ref{tab:cost_efficiency} reports cost-efficiency calculations for
10 selected proprietary models. We calculate total costs from provider
API prices under a common setting of 1,027 requests, each with a
10-minute audio-visual input. Cost efficiency is calculated separately
using the Overall Acc., C.G., and B.G. in Table~\ref{tab:main_results}:
for total cost $C$, request count $N$, and percentage score $s$, the
corresponding cost efficiency is $C/(N s/100)$. This gives the USD cost
per unit of accumulated normalized score; lower values indicate better
efficiency in answer correctness, evidence coverage, and precise
dual-modal localization, respectively.
Doubao-Seed-2.1-Lite achieves the best cost efficiency on all three
metrics, at \$0.0188, \$0.0208, and \$0.0491, respectively, with a total
cost of \$13.71. Qwen3.8-Omni-Flash has the lowest total cost (\$11.54)
and ranks second on all three metrics, at \$0.0193, \$0.0246, and
\$0.0780. Gemini3.7-Flash achieves the highest Overall scores among these
models on all three metrics, but ranks fourth in each corresponding
cost-efficiency column, at \$0.0545, \$0.0611, and \$0.1232.
Nevertheless, it outperforms Gemini2.5-Pro across all three metrics while
reducing total cost from \$224.48 to \$44.41 and lowering the respective
costs per unit of normalized score by 82.3\%, 82.8\%, and 83.7\%.
These results show that stronger performance does not necessarily
require higher spending or imply better cost efficiency. Model selection
should jointly consider the budget and the required levels of answer
correctness, evidence coverage, and localization precision.

\begin{takeawaybox}
\textbf{Finding 3:}
Higher scores do not necessarily imply better cost efficiency across
Acc., C.G., and B.G. Model selection should balance answering and
grounding performance with budget constraints.
\end{takeawaybox}

\subsection{Further Analysis}

\textbf{Input Modality Ablation.}
In Table~\ref{tab:modality_ablation} and the following analysis, we abbreviate
MiMo-V2.5-310B-A15B, Qwen3VL-8B-Instruct, and Qwen3VL-Instruct-30B-A3B
as MiMo-V2.5, Qwen3VL-8B, and Qwen3VL-A3B, respectively.
Table~\ref{tab:modality_ablation} compares three models under different
input conditions, yielding three observations.
(1) \textit{Native audio outperforms ASR transcripts in the evaluated MiMo-V2.5 settings.}
With audio alone, MiMo-V2.5 achieves 30.0\% Acc. and 10.0\% C.G.,
compared with 23.8\% and 7.1\% using ASR alone, respectively.
The advantage persists when video is available: Video+Audio achieves
56.8\% Acc., 48.4\% C.G., and 20.5\% B.G., exceeding Video+ASR
by 8.1, 10.3, and 8.9 percentage points. Thus, ASR transcripts do not
fully substitute for audio in these settings. Non-speech sounds, prosody,
and transcription or timestamp errors may contribute to this gap. (2) \textit{Removing either modality substantially degrades answering and grounding.}
Relative to Video+Audio, MiMo-V2.5's accuracy drops by 26.8 percentage
points with Audio alone and 30.0 points with Video alone. C.G. similarly
falls from 48.4\% to 10.0\% and 9.3\%, respectively, while B.G. drops
from 20.5\% to 0.0\% in both cases. These declines highlight the value\begin{table}[t]
  \centering
  \begin{minipage}[t]{0.42\textwidth}
    \vspace{0pt}
    \begingroup
  \centering
  \captionof{table}{\textbf{Input modality ablation.} Bold marks the best input setting for each model and metric.}
  \label{tab:modality_ablation}
  \fontsize{7.5}{9}\selectfont
  \begingroup
  \renewcommand{\arraystretch}{1.08}
  \setlength{\tabcolsep}{2pt}
  \begin{tabular}{@{}lcccc@{}}
    \hline
    \textbf{Model} & \textbf{Input} & \textbf{Acc.} & \textbf{C.G.} & \textbf{B.G.} \\
    \hline
    \multirow{5}{*}{MiMo-V2.5}
      & Audio       & 30.0 & 10.0 &  0.0 \\
      & ASR         & 23.8 &  7.1 &  0.0 \\
      & Video       & 26.8 &  9.3 &  0.0 \\
      & Video+ASR   & 48.7 & 38.1 & 11.6 \\
      & Video+Audio & \textbf{56.8} & \textbf{48.4} & \textbf{20.5} \\
    \hline
    \multirow{2}{*}{Qwen3VL-8B}
      & Video       & 13.8 &  0.3 &  0.0 \\
      & Video+ASR   & \textbf{23.2} & \textbf{9.7} & \textbf{1.9} \\
    \hline
    \multirow{2}{*}{Qwen3VL-A3B}
      & Video       &  8.5 &  0.7 &  0.0 \\
      & Video+ASR   & \textbf{19.3} & \textbf{7.0} & \textbf{0.8} \\
    \hline
  \end{tabular}
  \endgroup
\par
\endgroup

  \end{minipage}\hfill
  \begin{minipage}[t]{0.56\textwidth}
    \vspace{0pt}
    \begingroup
\centering
\begingroup\bfseries\boldmath
\captionof{table}{\textbf{Answer--grounding diagnostics.}
A/V: unweighted audio/visual IoU; $C$: correct answer;
$G$: both IoUs $\geq 0.5$. Here $\neg$ means ``not'' and a comma means ``and''.
Bold marks the highest A, V, and $C,G$ values and the lowest values in the three error columns.}
\label{tab:grounding_diagnostics}
\endgroup
\fontsize{7.5}{9}\selectfont
\setlength{\tabcolsep}{2pt}
\renewcommand{\arraystretch}{1.08}
\resizebox{\linewidth}{!}{\begin{tabular}{@{}lrrrrrr@{}}
\hline
\textbf{Model} & \textbf{A} & \textbf{V} & $\boldsymbol{C,G}$ & $\boldsymbol{C,\neg G}$ & $\boldsymbol{\neg C,G}$ & $\boldsymbol{\neg C,\neg G}$ \\
\hline
MiMo-V2.5 & 41.1 & 39.3 & 21.5 & 35.3 & 8.3 & 34.7 \\
Qwen3.5-Omni-Plus & 30.5 & 29.8 & 13.1 & 36.6 & 5.5 & 44.7 \\
Qwen3.8-Omni-Flash & 27.4 & 34.3 & 11.8 & 46.5 & \textbf{4.4} & 37.3 \\
Gemini2.5-Flash & 40.2 & 36.7 & 18.2 & \textbf{31.3} & 8.8 & 41.5 \\
Gemini2.5-Pro & 47.8 & 46.2 & 32.4 & 38.9 & 7.2 & 21.5 \\
Gemini3.5-Flash & 45.6 & 41.9 & 27.0 & 41.0 & 6.8 & 25.2 \\
Gemini3.7-Flash & \textbf{52.6} & \textbf{51.1} & \textbf{41.8} & 38.2 & 5.8 & \textbf{14.3} \\
\hline
\end{tabular}}
\par
\endgroup

  \end{minipage}
  \vspace{-3mm}
\end{table}

of complementary audio-visual evidence. (3) \textit{ASR improves vision-language models, but substantial grounding limitations remain.}
Adding ASR to video improves Qwen3VL-8B's accuracy from 13.8\% to 23.2\%
and Qwen3VL-A3B's from 8.5\% to 19.3\%, gains of 9.4 and 10.8
percentage points. Their C.G. scores also increase from 0.3\% to 9.7\%
and from 0.7\% to 7.0\%, respectively. Nevertheless, their B.G. scores
remain at only 1.9\% and 0.8\%, compared with MiMo-V2.5's 11.6\%
under the same Video+ASR condition. Transcripts therefore provide useful
additional evidence, but do not by themselves ensure effective evidence
integration and precise localization.

\begin{takeawaybox}
\textbf{Finding 4:}
Both modalities improve answer accuracy. Native audio outperforms ASR, while ASR improves the evaluated VLMs but leaves their precise grounding performance low.
\end{takeawaybox}

\textbf{Error Analysis.}
Table~\ref{tab:grounding_diagnostics} decomposes answering and grounding outcomes for seven models; details are in Appendix~\ref{app:grounding_diagnostics}.
Audio IoU exceeds visual IoU by 0.7--3.7 points for six of the seven models; Qwen3.8-Omni-Flash is the only exception (27.4\% versus 34.3\%). The two failure modes are asymmetric. Requiring both modality IoUs to reach 0.5, 48.1\%--79.9\% of correct answers fail the criterion ($C,\neg G$): Gemini3.7-Flash answers 79.4\% correctly but satisfies both criteria on only 41.8\%, and Qwen3.8-Omni-Flash drops
from 58.2\% to 11.8\%. Conversely, 12.2\%--32.6\% of grounding-passing predictions still choose the wrong option ($\neg C,G$), and $\neg C,G$ never exceeds 8.8\% of all questions, so accurate dual-modal localization is rarely reached where answers are wrong. Therefore, neither answer correctness nor temporal overlap alone is sufficient to establish reliable evidence-grounded answering.

\begin{takeawaybox}
\textbf{Finding 5:}
Answer correctness does not guarantee accurate dual-modal grounding, and grounding success does not guarantee a correct answer.
\end{takeawaybox}

\section{Conclusion}

We introduce EvaOmni, a new evidence-grounded benchmark for audio-visual reasoning. Rather than evaluating only whether a model selects the correct answer, EvaOmni also requires the model to identify the supporting evidence necessary to justify its prediction. Through a series of rigorous annotation and quality-control procedures, EvaOmni comprises 1,027 questions organized into three hierarchical levels, providing diverse scenarios and rich data for comprehensive evaluation. We further introduce multiple Grounded Score metrics to evaluate the accuracy of evidence localization. Among them, the Coverage Grounded Score rewards evidence coverage, while the Bottleneck Grounded Score prevents strong localization in one modality from masking failure in the other. Additional Grounded Score variants are provided in the Appendix. Large-scale evaluation shows that combining grounded evaluation with accuracy reveals insights missed by accuracy alone. EvaOmni provides a reproducible, interpretable basis for measuring progress and identifies precise temporal grounding as the capability current Omni LLMs most clearly lack. We have fully released the data, annotations, and evaluation framework, and we hope they will serve as a valuable resource for the research community to advance the development of Omni LLMs.

\bibliographystyle{plainnat}
\bibliography{references}

\clearpage
\renewcommand{\textfraction}{0.01}      \renewcommand{\floatpagefraction}{0.6}  \renewcommand{\topfraction}{0.99}       \renewcommand{\bottomfraction}{0.99}    \newpage
\appendix
\startcontents[appendix]
\section*{Appendix Contents}
\printcontents[appendix]{}{1}{\setcounter{tocdepth}{2}}
\bigskip

\section{Dataset Construction and Annotation Details}
\label{app:dataset_construction}

\subsection{Video Collection and Filtering}

We select candidate videos according to four criteria, aiming to reduce
potential training data contamination and textual shortcuts while providing
sufficient cues for constructing audio-visual reasoning questions.

\textbf{Temporal freshness and potential contamination screening.}
We consider only videos uploaded after January 2025. As an additional
screening step, we randomly sample three frames from each video and provide
them to Gemini3.1-Pro-Preview, asking it to describe the video content.
Videos for which the model can reconstruct the full content from these
frames are flagged as potential contamination risks and excluded from the
candidate pool. This procedure is a heuristic: neither the upload date nor
the model's ability to reconstruct the content establishes whether a video
was present in its training data.

\textbf{Avoiding subtitle-based textual shortcuts.}
We exclude videos with hard-coded subtitles to prevent models from
substituting on-screen text recognition for audio understanding. This
criterion supports the intended requirement that questions depend jointly
on audio and visual information.

\textbf{Manageable video duration.}
We select videos lasting 1--12 minutes to balance temporal scope, model
input budgets, and annotation feasibility. At a sampling rate of 1.0 FPS,
this range corresponds to approximately 60--720 frames, helping control
the visual input size while retaining temporally structured event
sequences. Extending the collection to longer videos is a direction for
future work on audio-visual reasoning over longer time spans.

\textbf{Sufficient diversity of entities and events.}
We prioritize videos with varied participants, actions, or event
developments. Videos with highly repetitive content, such as a single
person continuously performing one action, may provide too few cues for
constructing plausible distractors. We manually exclude videos that do
not support adequate distractor construction, reducing opportunities to
answer questions through unimodal cues or simple elimination.

\subsection{Annotation Principles and Schema}

Five senior PhD students with research backgrounds in computer science
and specialized knowledge of linguistics participate in annotation.
Each sample records the video URL, question, answer options, capability
level, and fine-grained capability category. Annotators separately mark
the visual and audio evidence supporting the answer, with each evidence
segment containing start and end timestamps in \texttt{mm:ss} format and
a content description. For Level~3 multi-hop reasoning questions, the
annotation additionally includes a step-by-step reasoning chain.

\textbf{Question and distractor design.}
Each question must match its assigned capability category and level and
be designed to require joint use of audio and visual information. Distractors draw on visual and audio cues present in the video so
that a single modality is insufficient to reliably distinguish the
correct option from the alternatives. They also represent logically
plausible alternative explanations, limiting opportunities to eliminate
incorrect options solely through wording or logical cues in the question
and answer choices.

\subsection{Human Cross-Validation}

After initial annotation, each expert's question-answer pairs are randomly
divided into four parts, which are randomly assigned, one to each of the other
four experts, for cross-validation. Each expert therefore reviews approximately
as many pairs as they initially annotated. The reviewer checks three
dimensions:
\begin{enumerate}
    \item \textbf{Capability alignment:} whether the question matches its
    assigned fine-grained capability and capability level.
    \item \textbf{Shortcut susceptibility:} whether the question can be
    reliably answered using only the question and options, audio alone,
    or vision alone through logical or unimodal cues.
    \item \textbf{Evidence quality:} whether the audio and visual evidence
    intervals and descriptions are accurate and support the annotated
    answer.
\end{enumerate}
Approximately 78\% of the initially submitted pairs pass this first review.
The remaining disputed pairs are independently reviewed by two additional
experts, neither of whom is the original annotator or the first reviewer. A
pair is discarded if either additional expert identifies a problem; it passes
only if both approve it. Overall, approximately 82\% of the initially
submitted pairs pass human cross-validation, while the remaining 18\% are
discarded at this stage.

\subsection{Model-Assisted Shortcut Screening}

Following human validation, we use Gemini3.1-Pro-Preview and Qwen3-Omni-Instruct-30B-A3B
for additional shortcut screening under text-only, audio-only, and
vision-only conditions. We inspect both the answers and their supporting
rationales, as well as the evidence from the available modality in
unimodal settings. If a model can reliably answer a question correctly
and provide a plausible supporting rationale despite the absence of one
or both modalities, the sample is returned for revision. Revised samples
must pass model-assisted screening again before inclusion.

The resulting EvaOmni benchmark contains 1,027 question-answer pairs
curated through this annotation and quality-control process.

\subsection{EvaOmni Structure}
\label{app:capability_taxonomy}

EvaOmni contains 392 videos and 1,027 question-answer pairs across three
capability levels. The main text defines the three levels; here we define
the fine-grained capabilities within each one.
\textbf{Level~1 (436 questions).}
\begin{enumerate}
    \item \textbf{Temporal Alignment (165 questions):} Identify the main content of one modality using information from the other modality.
    \item \textbf{Event Binding (143 questions):} Summarize the main event in one modality using information from the other modality.
    \item \textbf{Entity Binding (128 questions):} Link the main content of one modality to the corresponding content in the other modality.
\end{enumerate}

\textbf{Level~2 (438 questions).}
\begin{enumerate}
    \item \textbf{Causal Reasoning (127 questions):} Infer the causes of events.
    \item \textbf{Emotional Intelligence (87 questions):} Interpret emotional states from audio-visual cues.
    \item \textbf{Intent Understanding (56 questions):} Infer a person's intent from audio-visual cues.
    \item \textbf{Atmosphere Understanding (28 questions):} Interpret the atmosphere of a scene from audio-visual cues.
    \item \textbf{Implicit Event Reasoning (18 questions):} Infer unstated events, such as an omitted or likely future event.
    \item \textbf{Semantic Disambiguation (30 questions):} Resolve the specific meaning of ambiguous speech or behavior using context from both modalities.
    \item \textbf{Temporal Understanding (31 questions):} Determine the order, overlap, or duration relationships between audio and visual events.
    \item \textbf{Spatial Understanding (30 questions):} Reason about spatial locations and relationships.
    \item \textbf{Conditional Counting (31 questions):} Count entities or events that satisfy specified conditions.
\end{enumerate}

\textbf{Level~3 (153 questions).}
\begin{enumerate}
    \item \textbf{Multi-hop Reasoning (83 questions):} Answer questions that require multiple linked inference steps.
    \item \textbf{Long-range Reasoning (35 questions):} Integrate dispersed evidence whose most widely separated start times are at least four minutes apart.
    \item \textbf{Multi-Evidence Chain (35 questions):} Reason over at least five temporally separated pieces of evidence.
\end{enumerate}

\section{Experimental Details}
\subsection{Input Preprocessing and Inference Setup}
\label{app:input_preprocessing}

\textbf{Input preprocessing.}
For evaluation, all videos are resized to 480 pixels on the longer side
while preserving their original aspect ratios. Frames are uniformly
sampled at 1.0 FPS and provided to the Omni LLMs together with the
corresponding audio. For the 1--12-minute videos in EvaOmni, this
configuration keeps the audio-visual inputs within the context windows
of the evaluated models.

\textbf{Inference setup.}
All open-weight models except those in the MiMo family are evaluated on
a single server equipped with eight NVIDIA H20 GPUs, using the inference
frameworks and parameter settings recommended in their respective model
cards. The three MiMo models are evaluated
through MiMo's official API platform with the same input preprocessing
settings described above. Since the weights of these three models have been publicly released, we classify them as open-weight models.

\subsection{Human Expert Baseline}
\label{app:human_baseline}

The human expert baseline is obtained from test participants who are
independent of the annotators who constructed the questions and marked the
reference evidence. They answer the same questions using the same test videos
as the models and report supporting audio and visual intervals; their
responses are scored using the same Acc., C.G., and B.G. metrics. Participants
may pause and replay the videos when answering and locating evidence. This
allowance helps compensate for human memory, which is much more limited than a
model's access to the supplied video context, but it also makes the viewing
protocol different from model inference. Even with replay, the need to retain
and revisit details across a video may disadvantage humans, especially for
long or temporally dispersed evidence. We therefore treat the human scores as
a reference baseline rather than a strictly matched comparison of memory or
viewing conditions.

\section{Prompt Templates}
\label{app:prompt_templates}
\begin{figure}[!htbp]
    \centering
    \includegraphics[width=0.9\textwidth]{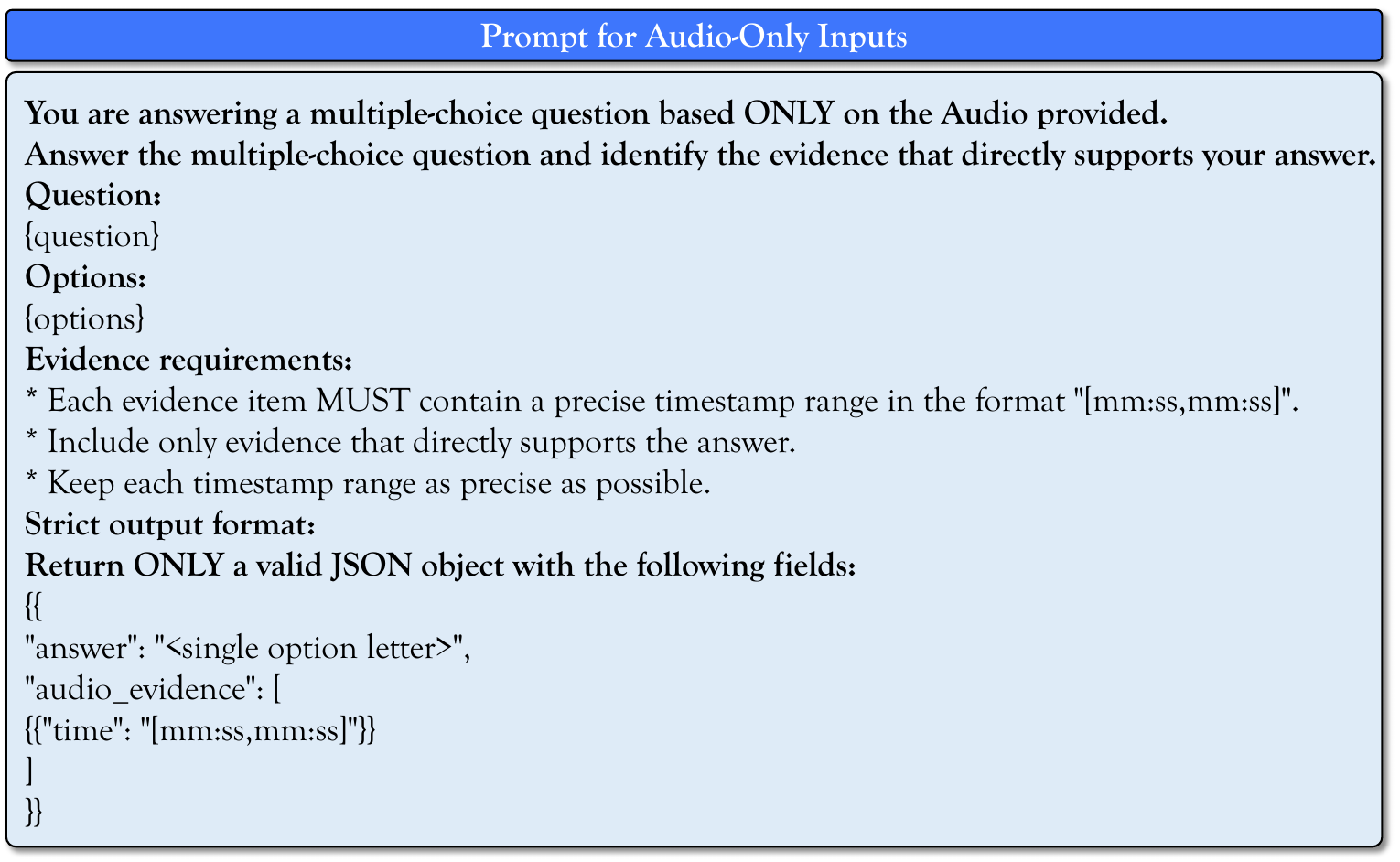}
    \caption{Prompt used for audio-only inputs.}
    \label{fig:prompt_audio_only}
    \par\vspace{0.5em}
    \includegraphics[width=0.9\textwidth]{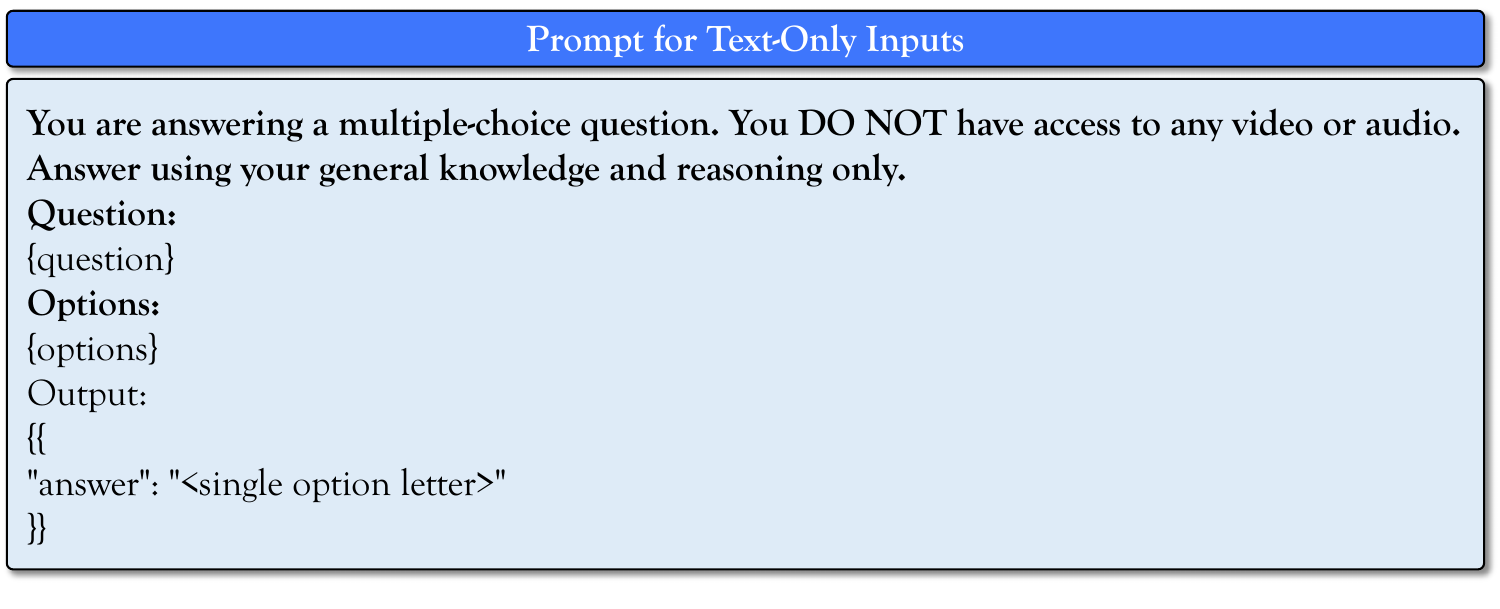}
    \caption{Prompt used for text-only inputs.}
    \label{fig:prompt_text_only}
\end{figure}

\begin{figure}[!htbp]
    \centering
    \includegraphics[width=0.9\textwidth]{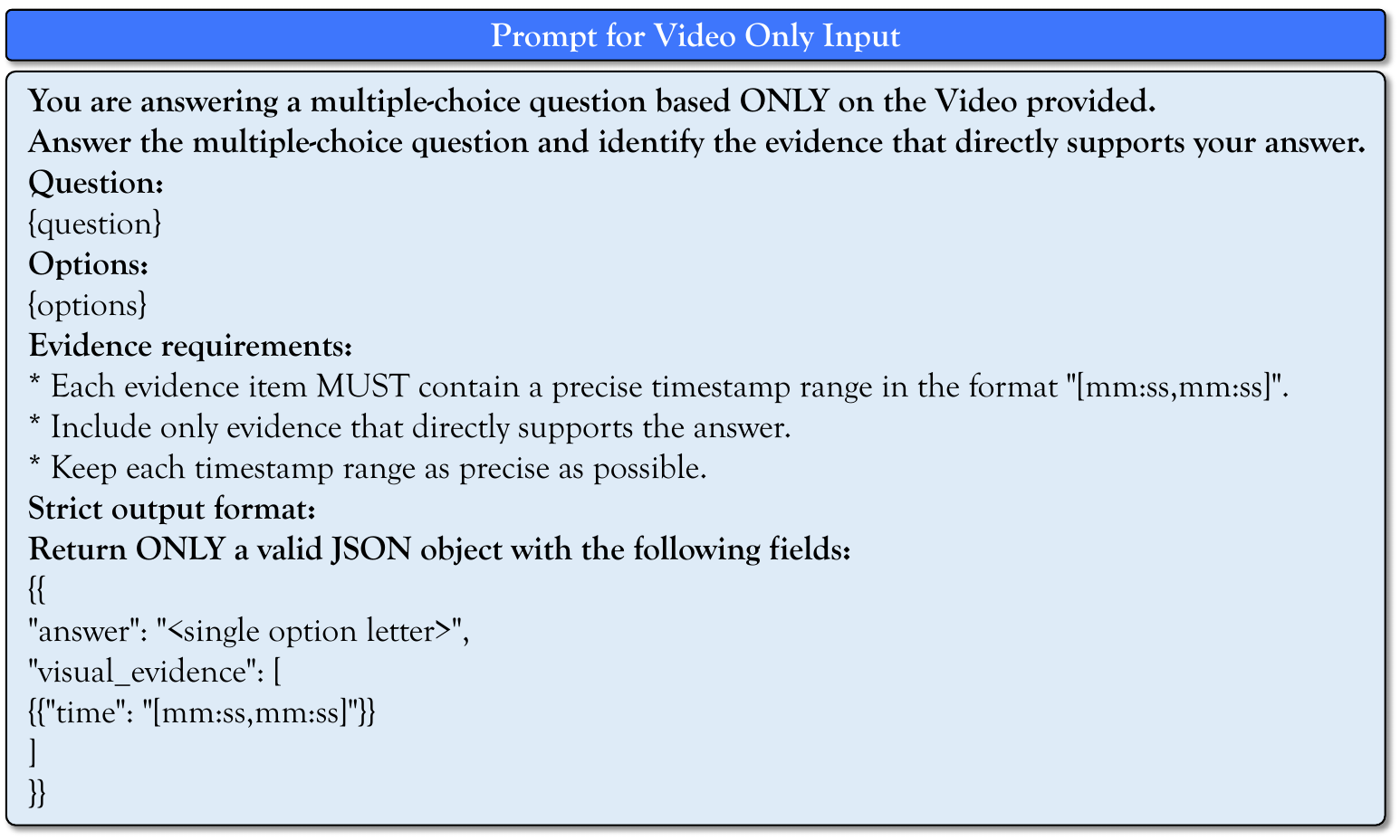}
    \caption{Prompt used for video-only inputs.}
    \label{fig:prompt_video_only}
    \par\vspace{0.5em}
    \includegraphics[width=0.9\textwidth]{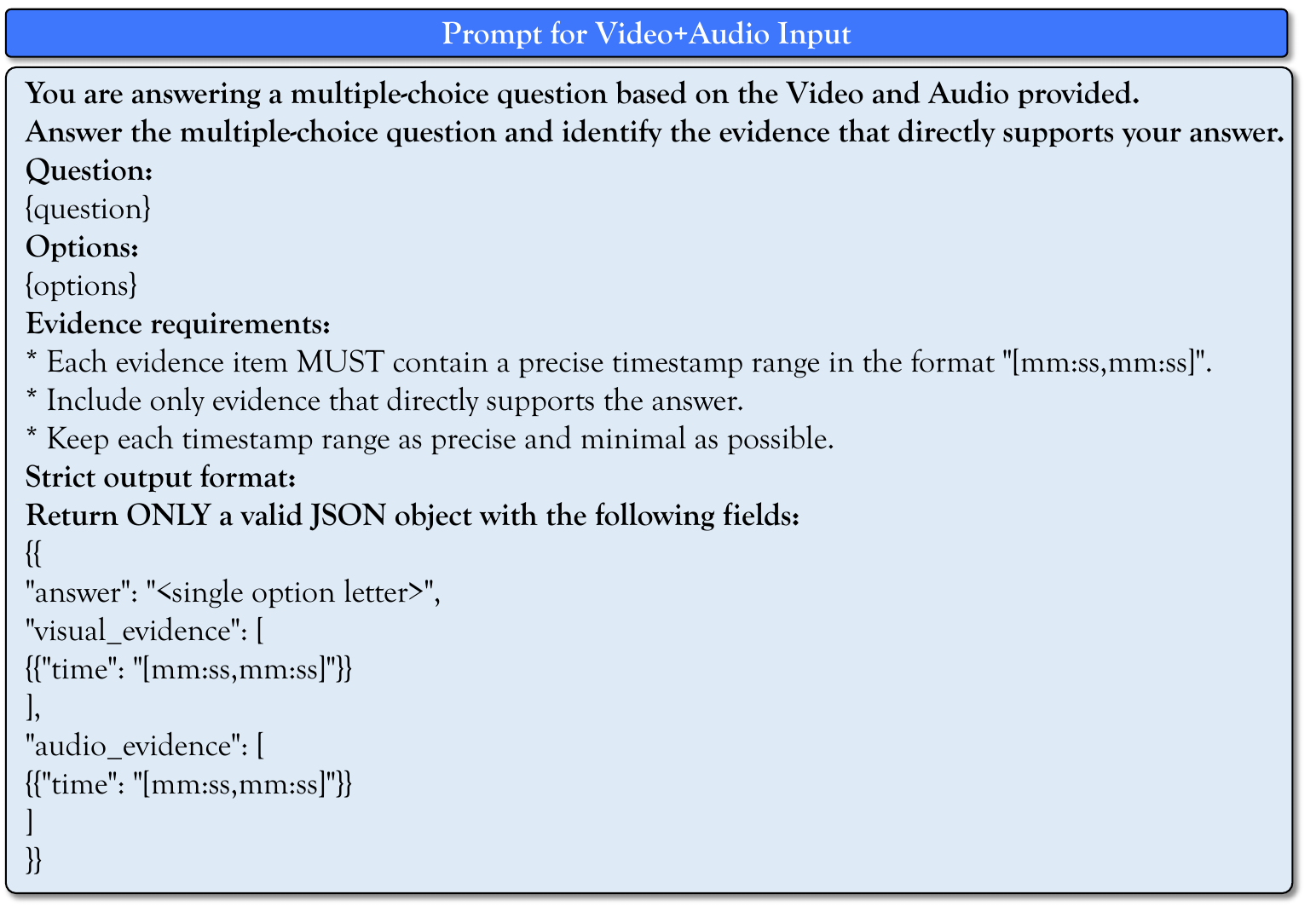}
    \caption{Prompt used for Video+Audio inputs.}
    \label{fig:prompt_video_audio}
\end{figure}

\begin{figure}[!htbp]
    \centering
    \includegraphics[width=0.9\textwidth]{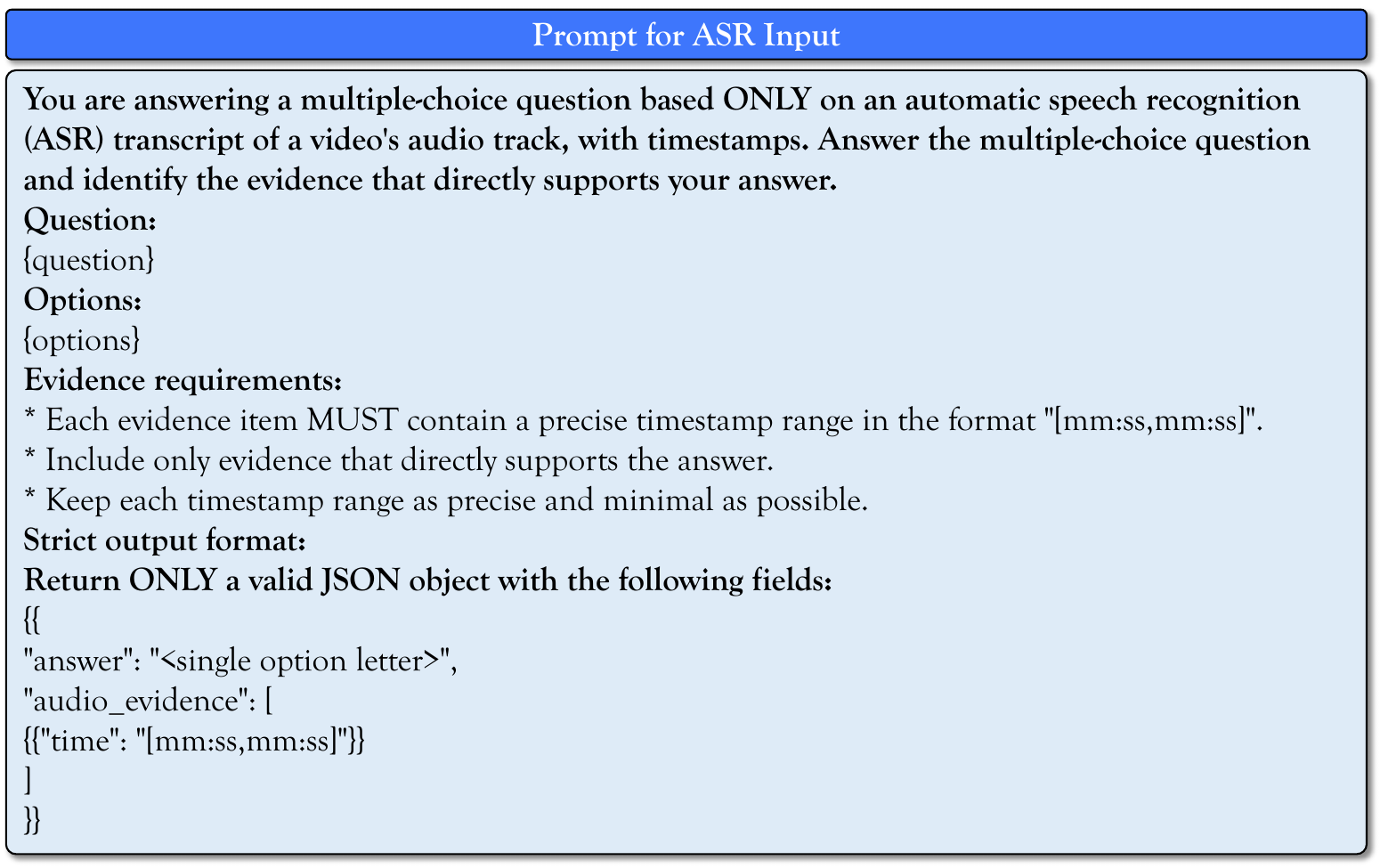}
    \caption{Prompt used for ASR-only inputs.}
    \label{fig:prompt_asr_only}
    \par\vspace{0.5em}
    \includegraphics[width=0.9\textwidth]{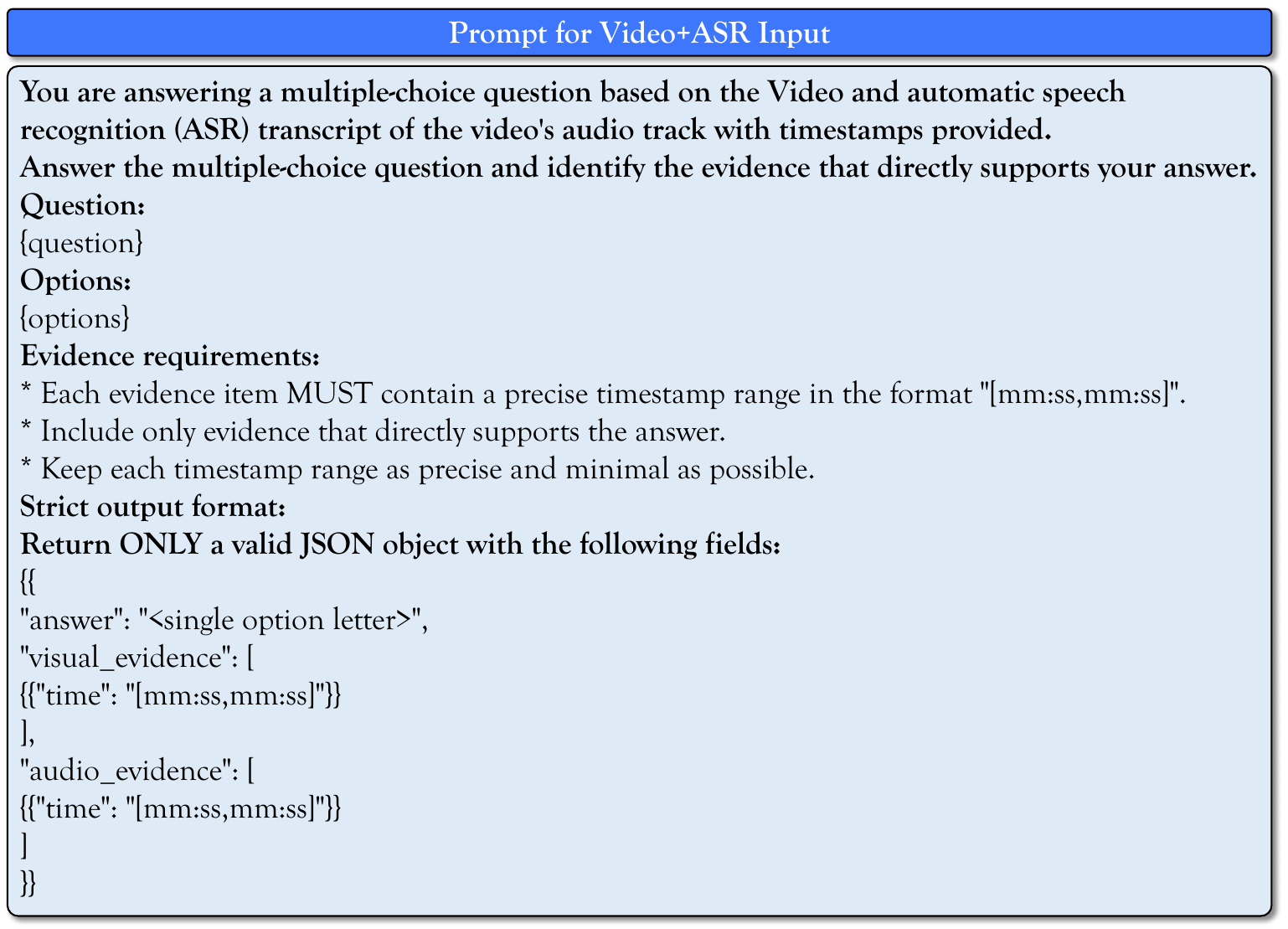}
    \caption{Prompt used for Video+ASR inputs.}
    \label{fig:prompt_video_asr}
\end{figure}

Figures~\ref{fig:prompt_audio_only}, \ref{fig:prompt_text_only},
\ref{fig:prompt_video_only}, \ref{fig:prompt_video_audio},
\ref{fig:prompt_asr_only}, and~\ref{fig:prompt_video_asr} show the prompt
templates used for the six input settings.

\section{Additional Quantitative Results and Analysis}

\subsection{Shortcut Evaluation and Case Studies}
\label{app:shortcut_evaluation}

Figure~\ref{fig:shortcut} summarizes the shortcut evaluation results for
three representative benchmarks for Omni LLMs, alongside EvaOmni.
Here, we provide further details of the evaluation protocol and report
the complete results in Table~\ref{tab:shortcut_comparison}.

\textbf{Evaluation protocol.}
For the visual-only condition, denoted as \textit{Video Only} in the
table, we use Qwen3-Omni-30B-A3B-Instruct\footnote{\url{https://huggingface.co/Qwen/Qwen3-Omni-30B-A3B-Instruct}} and
Qwen3-VL-30B-A3B-Instruct\footnote{\url{https://huggingface.co/Qwen/Qwen3-VL-30B-A3B-Instruct}}. For the audio-only condition, we use
Qwen3-Omni-30B-A3B-Instruct and Audio-Flamingo-3\footnote{\url{https://huggingface.co/nvidia/audio-flamingo-3}}. For the text-only
condition, we use Gemini3.1-Pro-Preview\footnote{\url{https://deepmind.google/models/gemini/pro/}} and Qwen3.8-Flash\footnote{\url{https://huggingface.co/collections/Qwen/qwen38-flash-next}}, providing
only the question and answer options. Under each input condition,
we evaluate each model twice, yielding four sets of predictions, and
take the intersection of the sets of correctly answered questions
across all four runs. The \textit{Intersection} column reports the
percentage of all evaluated questions answered correctly in every run.
This design aims to reduce the influence of chance successes in a
single run and biases specific to an individual model, although it
does not eliminate these factors.

\textbf{Interpreting the results.}
These results serve as diagnostic indicators of potential shortcut
susceptibility. Consistently correct answers from multiple models
under unimodal or text-only inputs do not necessarily imply that a
question is flawed, nor do they alone establish that the models exploit
a shortcut. Instead, they indicate the extent to which correct answers
can be repeatedly obtained without full audio-visual input under the
tested models and settings, highlighting a potential vulnerability of
answer-only evaluation. The reported rates should therefore not be
interpreted directly as the proportion of flawed questions or the
probability that a benchmark can be exploited.

\begin{table}[t]
  \centering
  \caption{\textbf{Benchmark comparison under text-only, audio-only, and
  video-only inputs.}
All values are percentages. Each model is evaluated twice, and an answer is considered correct only if both independent runs produce the correct answer.; Intersection denotes the proportion of evaluated questions
  answered correctly in all four runs under the corresponding input condition.}
  \label{tab:shortcut_comparison}
  \vspace{0.4em}
  \begingroup
  \small
  \renewcommand{\arraystretch}{1.22}
  \setlength{\tabcolsep}{3pt}
  \begin{minipage}[t]{0.48\linewidth}
    \centering
    \textbf{(a) Daily-Omni}\par\smallskip
    \begin{tabular*}{\linewidth}{@{\extracolsep{\fill}}lccc@{}}
      \hline
      \textbf{Input} & \textbf{Model 1} & \textbf{Model 2} & \textbf{Intersection} \\
      \hline
      Text Only  & 45.1 & 38.6 & 28.4 \\
      Audio Only & 30.2 & 57.9 & 25.6 \\
      Video Only & 41.2 & 45.2 & 29.4 \\
      \hline
    \end{tabular*}
  \end{minipage}\hfill
  \begin{minipage}[t]{0.48\linewidth}
    \centering
    \textbf{(b) WorldSense}\par\smallskip
    \begin{tabular*}{\linewidth}{@{\extracolsep{\fill}}lccc@{}}
      \hline
      \textbf{Input} & \textbf{Model 1} & \textbf{Model 2} & \textbf{Intersection} \\
      \hline
      Text Only  & 24.7 & 24.5 & 14.4 \\
      Audio Only & 20.7 & 46.0 & 14.2 \\
      Video Only & 25.5 & 39.8 & 15.7 \\
      \hline
    \end{tabular*}
  \end{minipage}

  \par\vspace{1em}
  \noindent
  \begin{minipage}[t]{0.48\linewidth}
    \centering
    \textbf{(c) OmniVideoBench}\par\smallskip
    \begin{tabular*}{\linewidth}{@{\extracolsep{\fill}}lccc@{}}
      \hline
      \textbf{Input} & \textbf{Model 1} & \textbf{Model 2} & \textbf{Intersection} \\
      \hline
      Text Only  & 29.2 & 25.8 & 16.0 \\
      Audio Only & 13.6 & 30.0 &  8.2 \\
      Video Only & 22.8 & 26.8 & 12.1 \\
      \hline
    \end{tabular*}
  \end{minipage}\hfill
  \begin{minipage}[t]{0.48\linewidth}
    \centering
    \textbf{(d) EvaOmni}\par\smallskip
    \begin{tabular*}{\linewidth}{@{\extracolsep{\fill}}lccc@{}}
      \hline
      \textbf{Input} & \textbf{Model 1} & \textbf{Model 2} & \textbf{Intersection} \\
      \hline
      Text Only  & \textbf{9.7} & \textbf{3.8}  & \textbf{1.8} \\
      Audio Only & \textbf{3.2} & \textbf{18.2} & \textbf{1.0} \\
      Video Only & \textbf{7.3} & \textbf{12.5} & \textbf{3.6} \\
      \hline
    \end{tabular*}
  \end{minipage}

  \par\medskip
  \begin{minipage}{\linewidth}
    \footnotesize
    \textbf{Model 1 / Model 2 by input condition:}\par
    \textit{Text Only:} Gemini3.1-Pro-Preview / Qwen3.8-Flash.\par
    \textit{Audio Only:} Audio-FlamingGo-3 / Qwen3-Omni-Instruct-30B-A3B.\par
    \textit{Video Only:} Qwen3VL-Instruct-30B-A3B / Qwen3-Omni-Instruct-30B-A3B.
  \end{minipage}
  \endgroup
\end{table}

\begin{figure}[p]
    \centering
    \includegraphics[width=\textwidth]{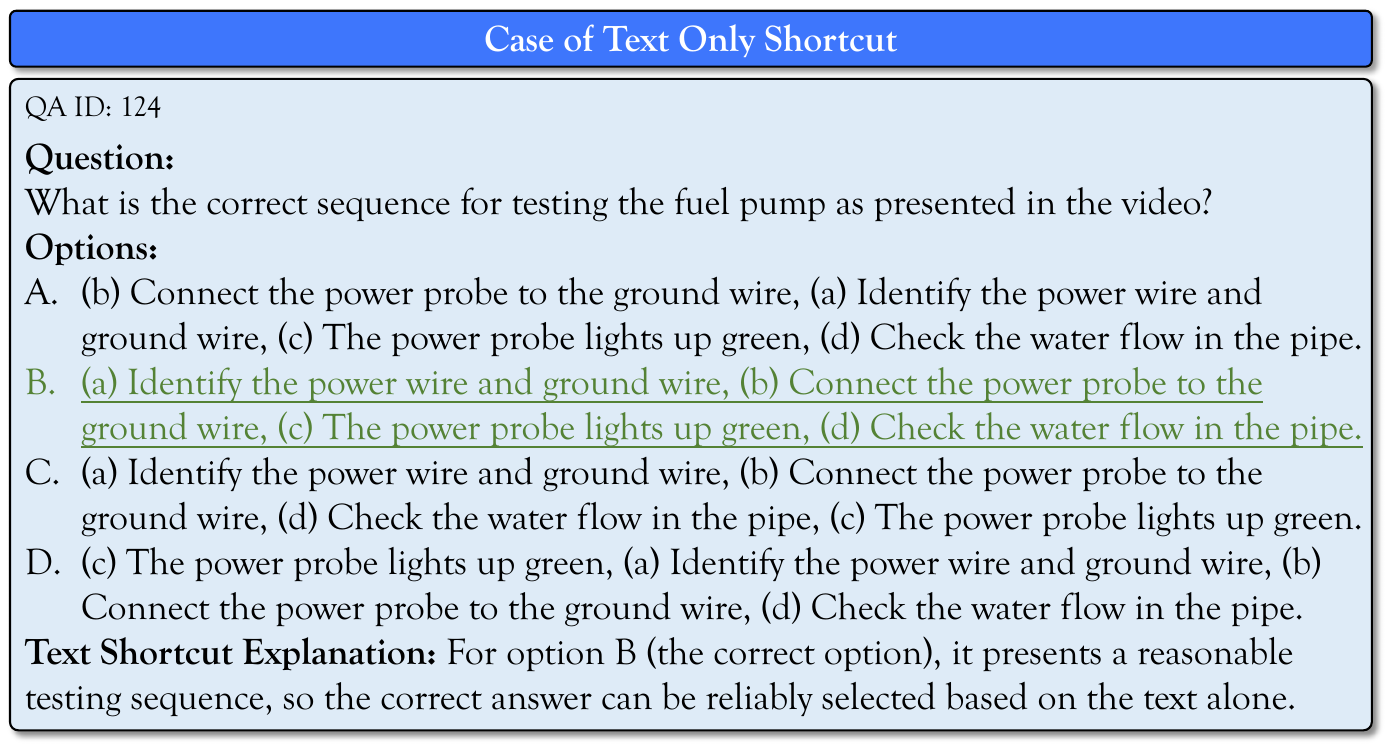}
    \caption{Text-only shortcut case 1 from WorldSense.}
    \label{fig:worldsense_text1}
    \par\vspace{0.5em}
    \includegraphics[width=\textwidth]{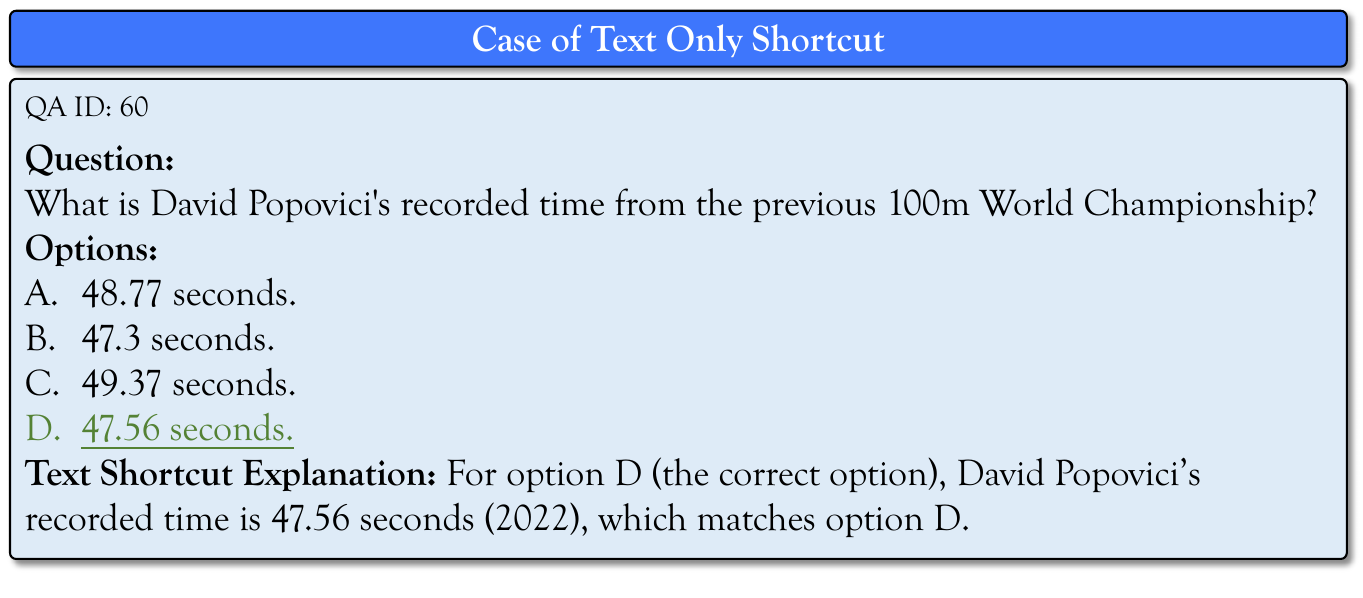}
    \caption{Text-only shortcut case 2 from WorldSense.}
    \label{fig:worldsense_text2}
\end{figure}

\begin{figure}[p]
    \centering
    \includegraphics[width=\textwidth]{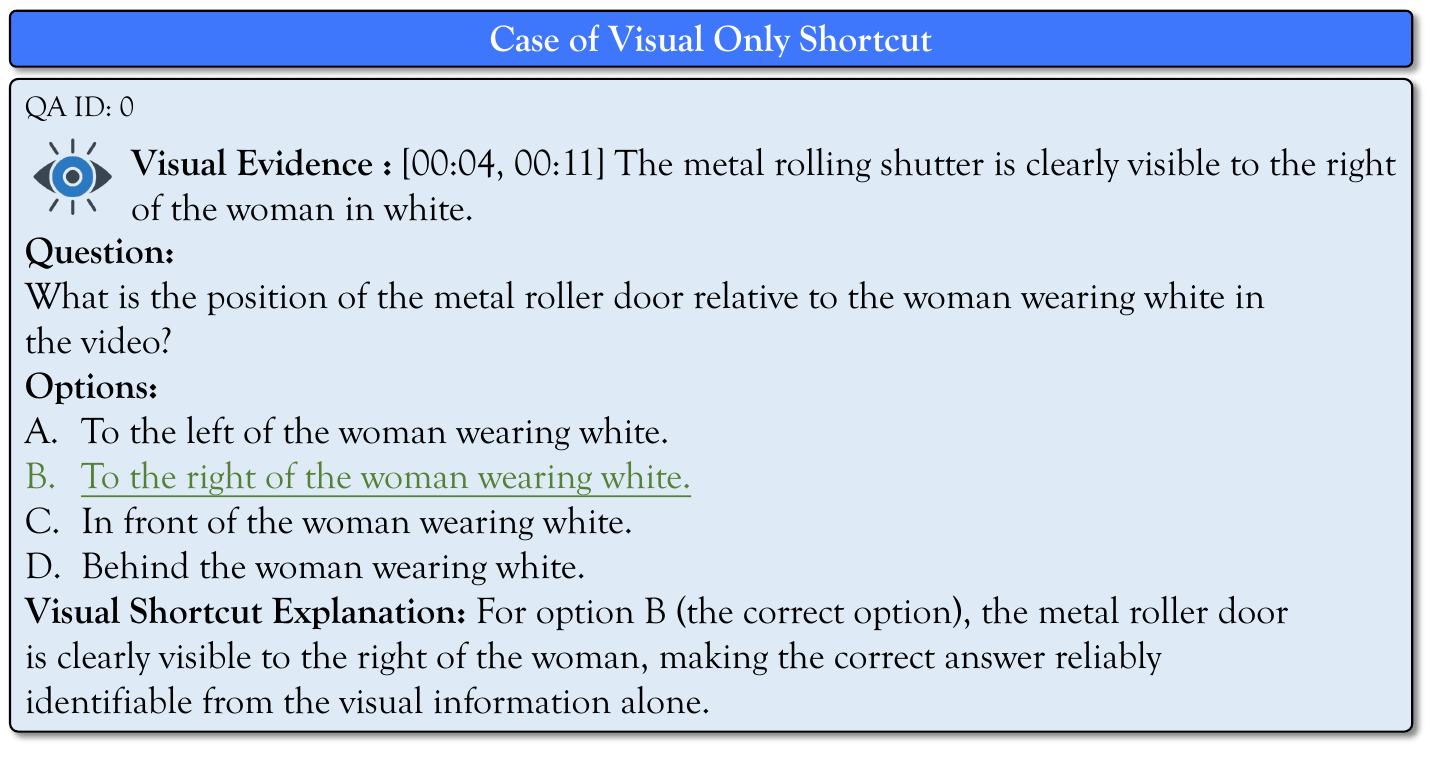}
    \caption{Visual-only shortcut case 1 from WorldSense.}
    \label{fig:worldsense_visual1}
    \par\vspace{0.5em}
    \includegraphics[width=\textwidth]{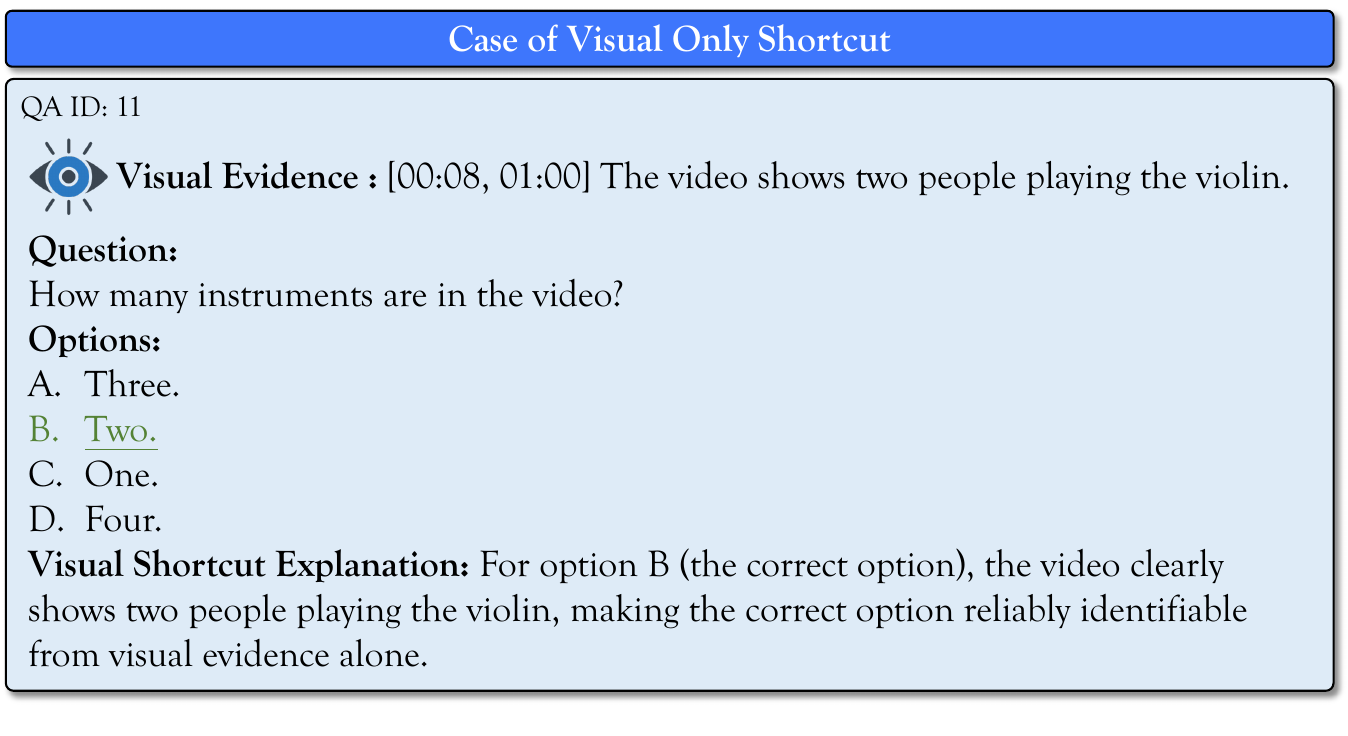}
    \caption{Visual-only shortcut case 2 from WorldSense.}
    \label{fig:worldsense_visual2}
\end{figure}

\begin{figure}[p]
    \centering
    \includegraphics[width=\textwidth]{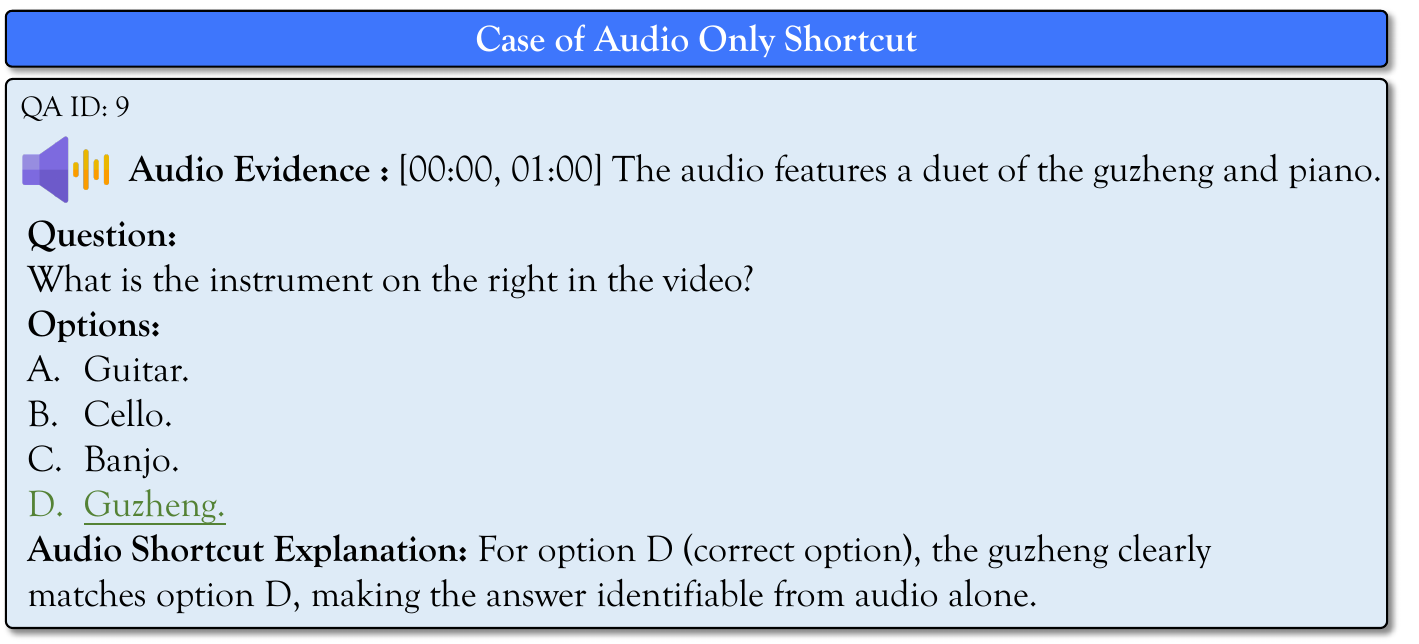}
    \caption{Audio-only shortcut case 1 from WorldSense.}
    \label{fig:worldsense_audio1}
    \par\vspace{0.5em}
    \includegraphics[width=\textwidth]{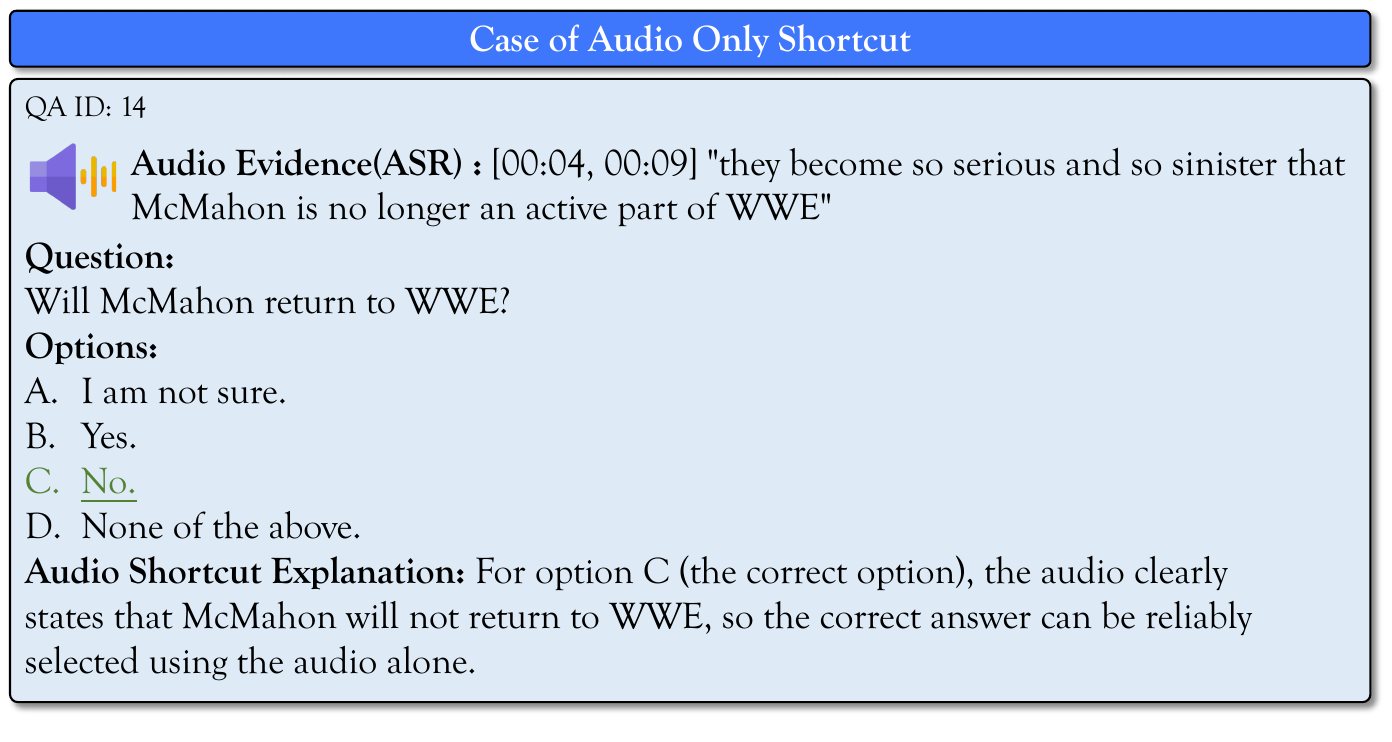}
    \caption{Audio-only shortcut case 2 from WorldSense.}
    \label{fig:worldsense_audio2}
\end{figure}

\begin{figure}[p]
    \centering
    \includegraphics[width=\textwidth]{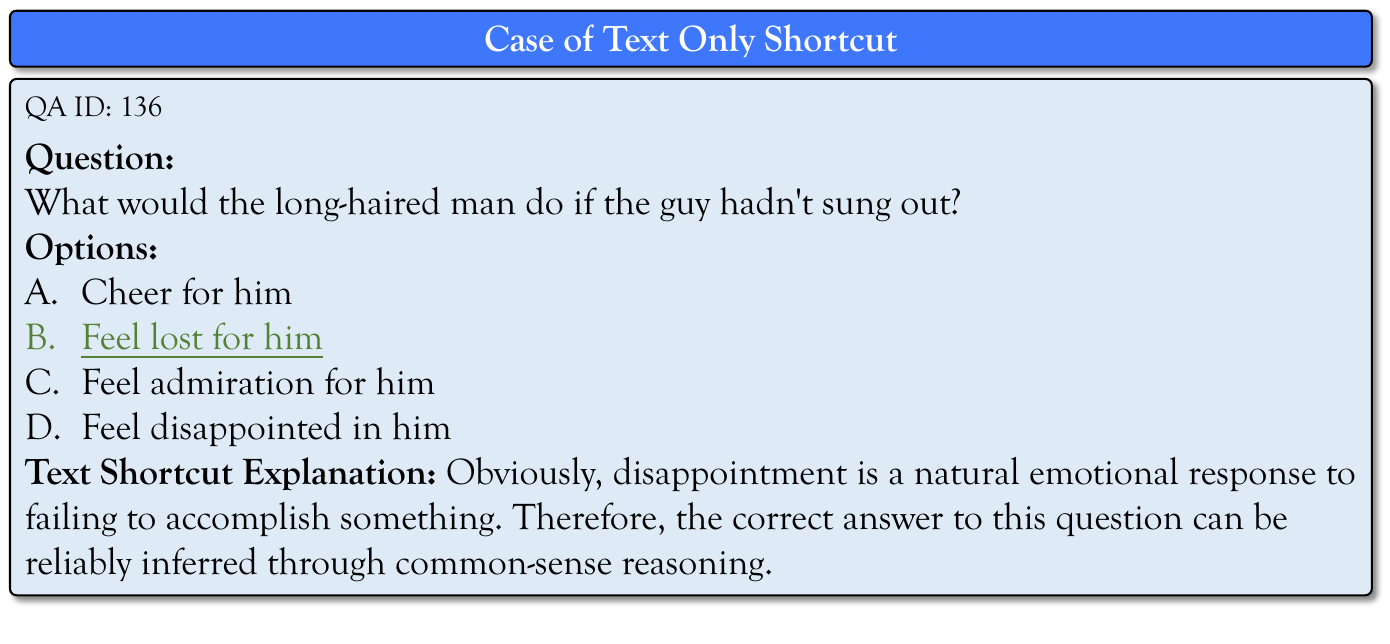}
    \caption{Text-only shortcut case 1 from OmniVideoBench.}
    \label{fig:omnivideobench_text1}
    \par\vspace{0.5em}
    \includegraphics[width=\textwidth]{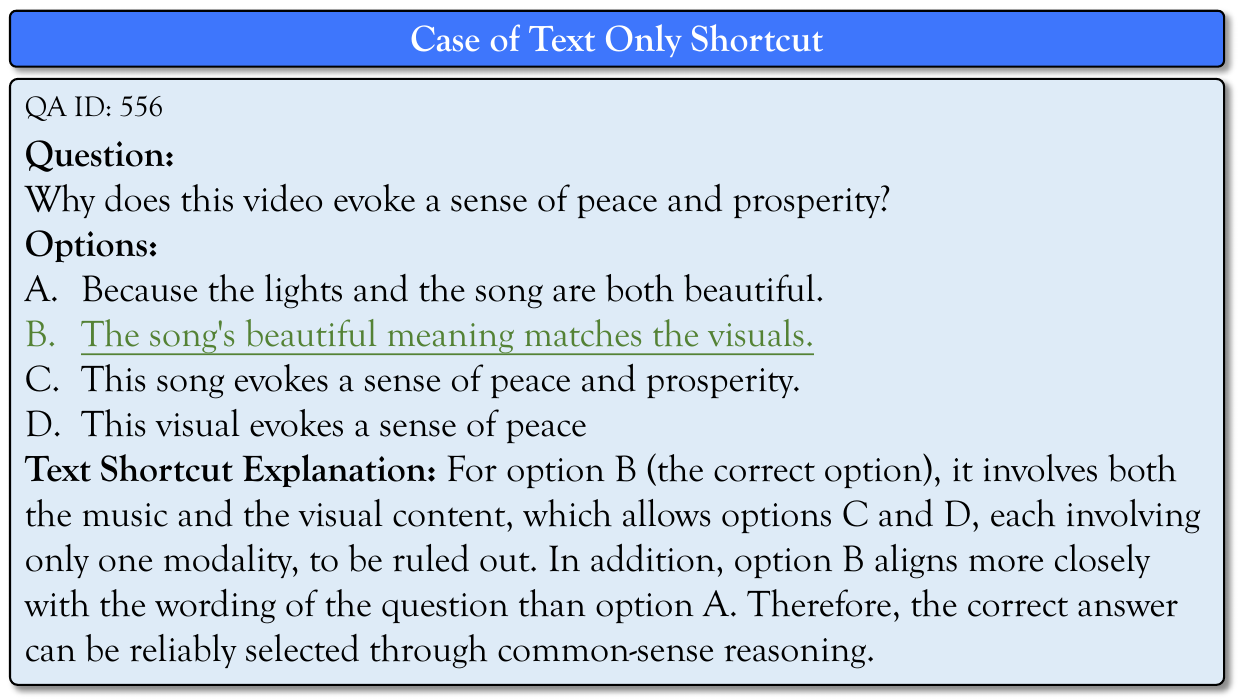}
    \caption{Text-only shortcut case 2 from OmniVideoBench.}
    \label{fig:omnivideobench_text2}
\end{figure}

\begin{figure}[p]
    \centering
    \includegraphics[width=\textwidth]{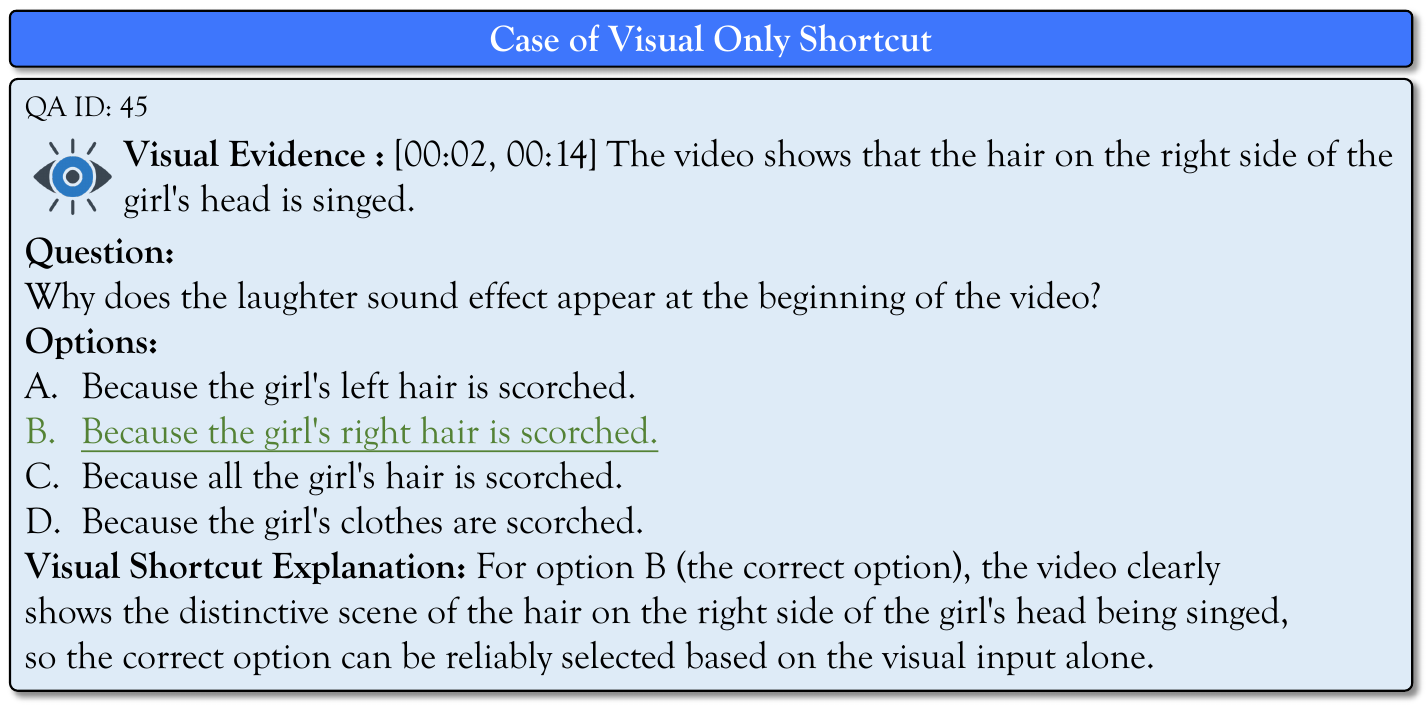}
    \caption{Visual-only shortcut case 1 from OmniVideoBench.}
    \label{fig:omnivideobench_visual1}
    \par\vspace{0.5em}
    \includegraphics[width=\textwidth]{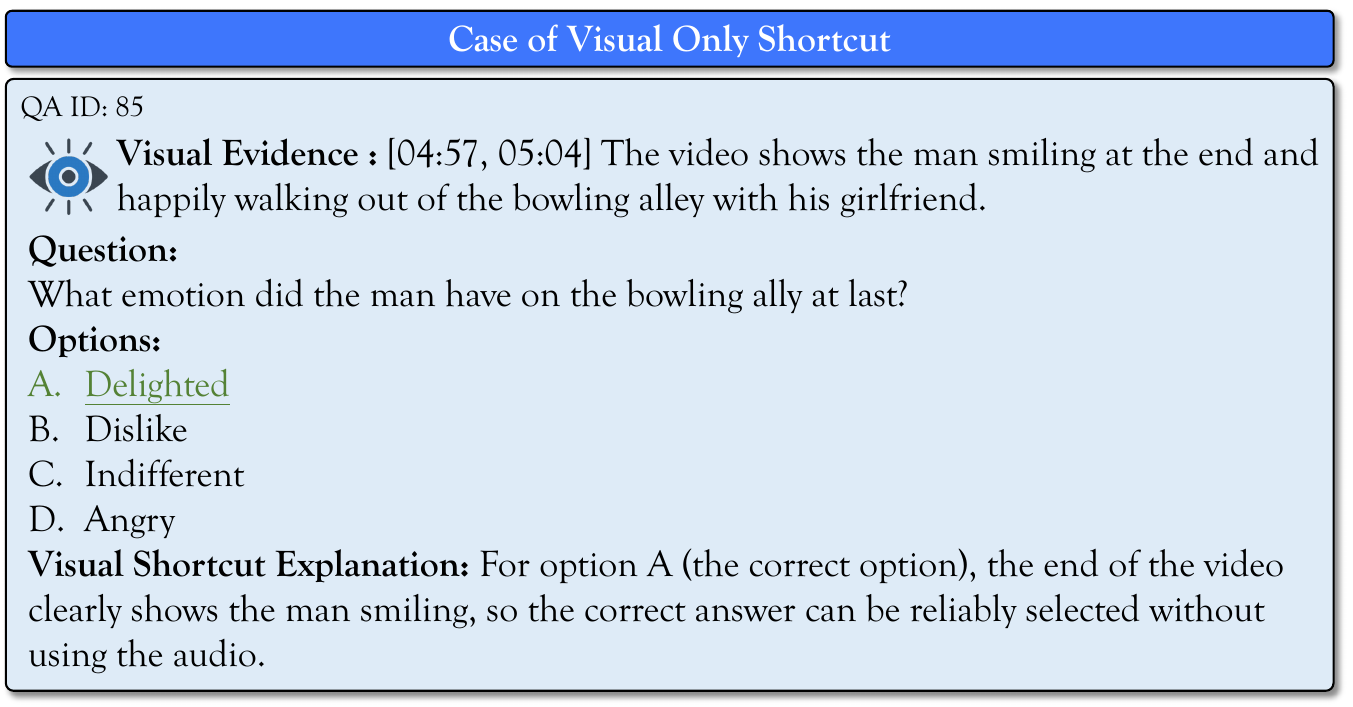}
    \caption{Visual-only shortcut case 2 from OmniVideoBench.}
    \label{fig:omnivideobench_visual2}
\end{figure}

\begin{figure}[p]
    \centering
    \includegraphics[width=\textwidth]{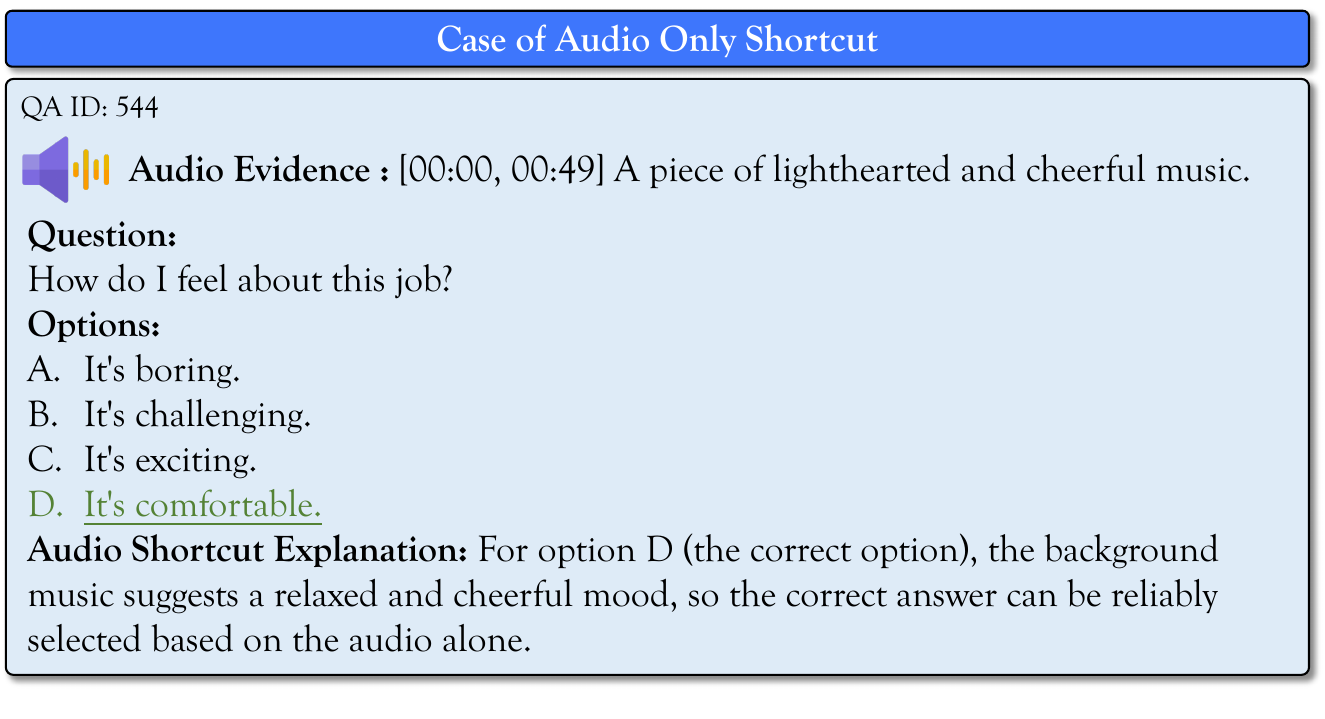}
    \caption{Audio-only shortcut case 1 from OmniVideoBench.}
    \label{fig:omnivideobench_audio1}
    \par\vspace{0.5em}
    \includegraphics[width=\textwidth]{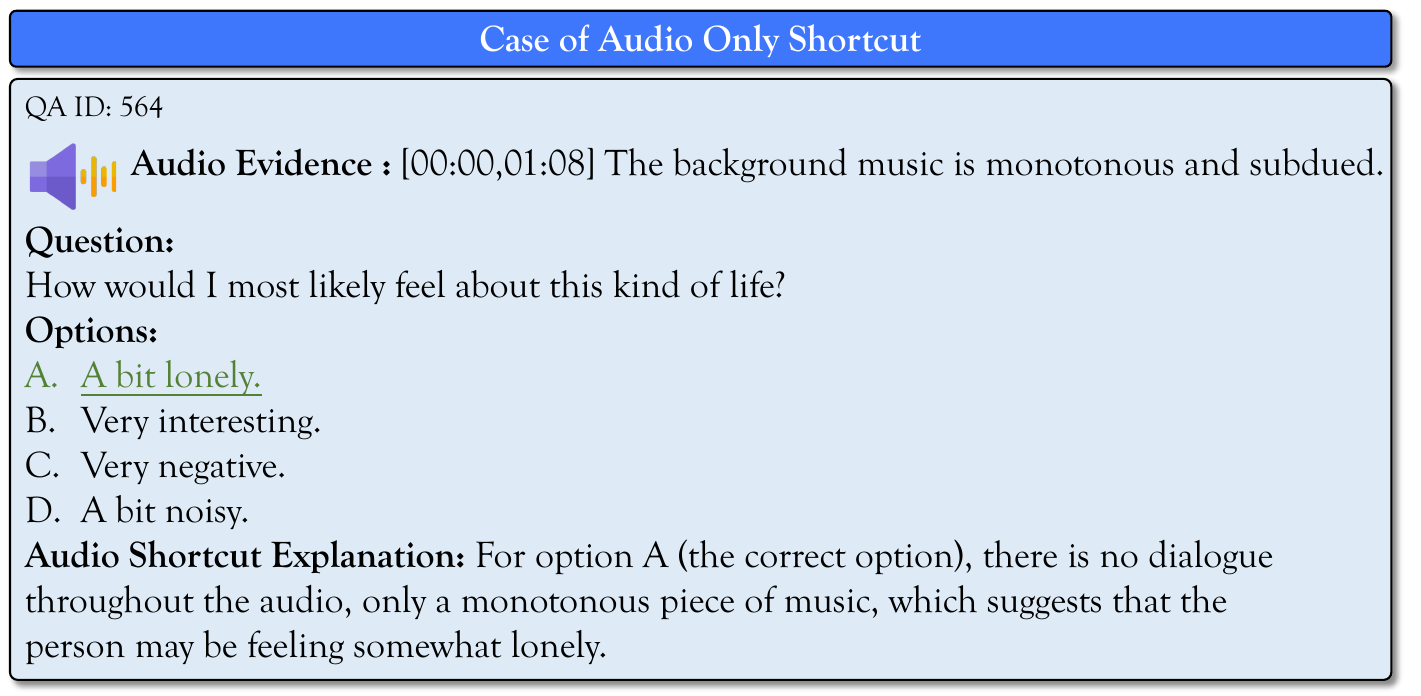}
    \caption{Audio-only shortcut case 2 from OmniVideoBench.}
    \label{fig:omnivideobench_audio2}
\end{figure}

\begin{figure}[p]
    \centering
    \includegraphics[width=\textwidth]{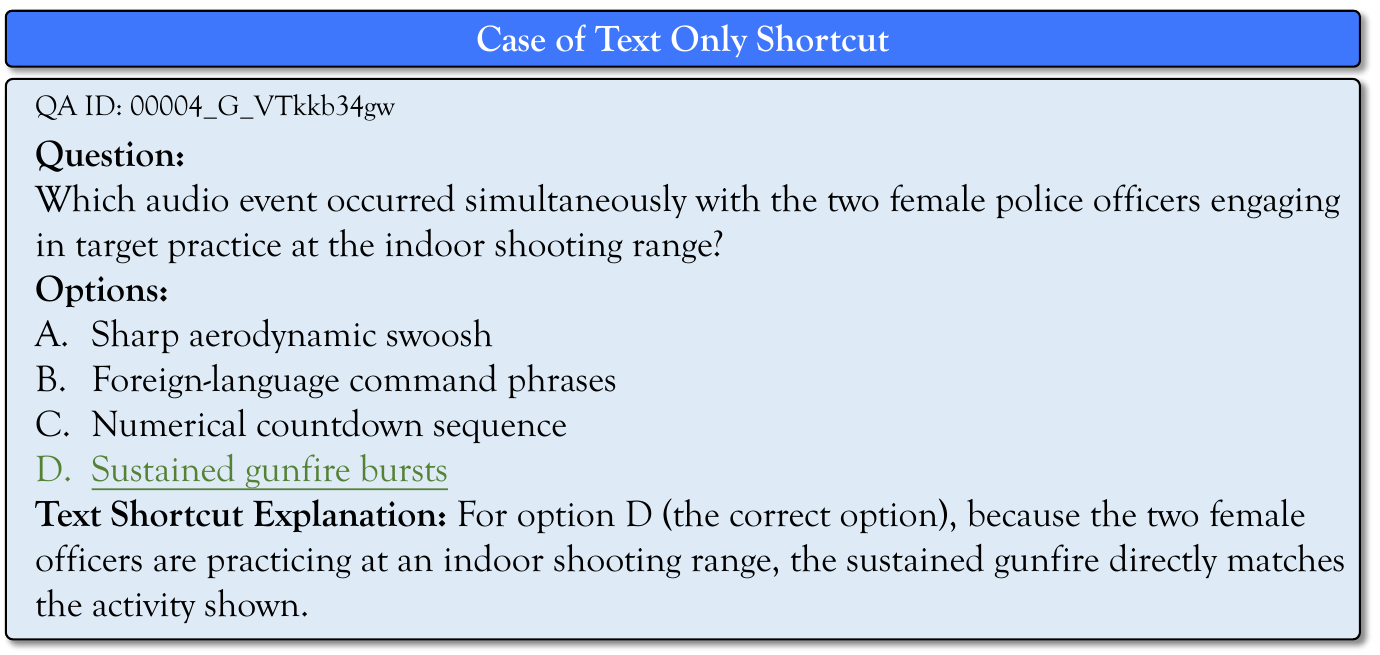}
    \caption{Text-only shortcut case 1 from Daily-Omni.}
    \label{fig:dailyomni_text1}
    \par\vspace{0.5em}
    \includegraphics[width=\textwidth]{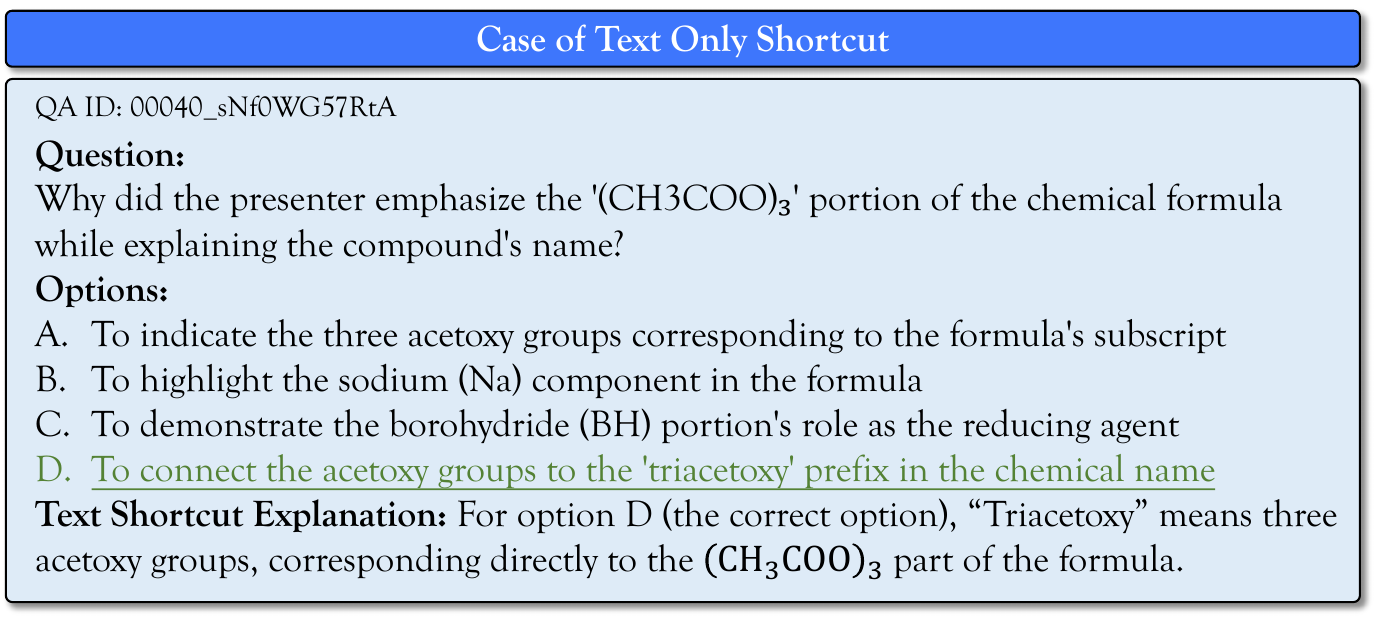}
    \caption{Text-only shortcut case 2 from Daily-Omni.}
    \label{fig:dailyomni_text2}
\end{figure}

\begin{figure}[p]
    \centering
    \includegraphics[width=\textwidth]{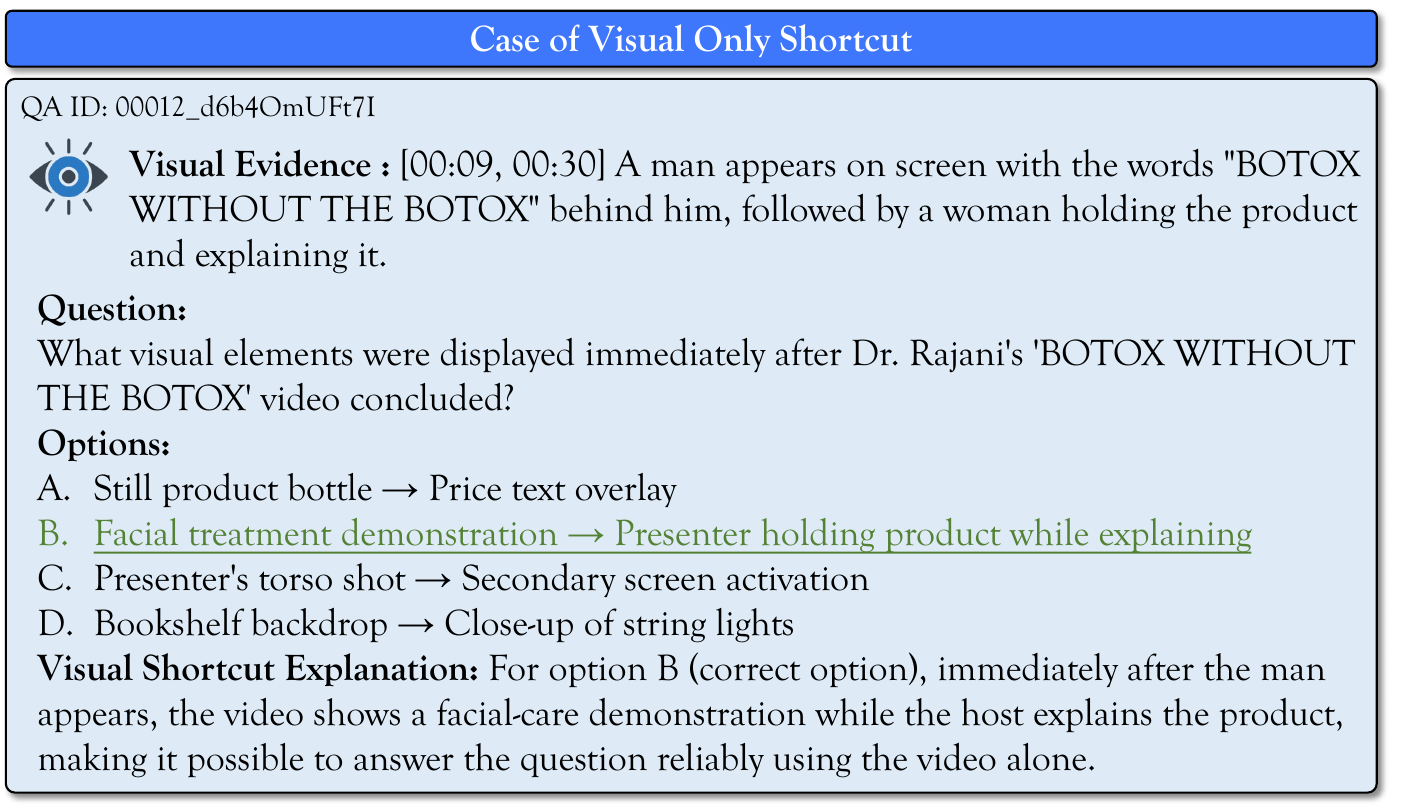}
    \caption{Visual-only shortcut case 1 from Daily-Omni.}
    \label{fig:dailyomni_visual1}
    \par\vspace{0.5em}
    \includegraphics[width=\textwidth]{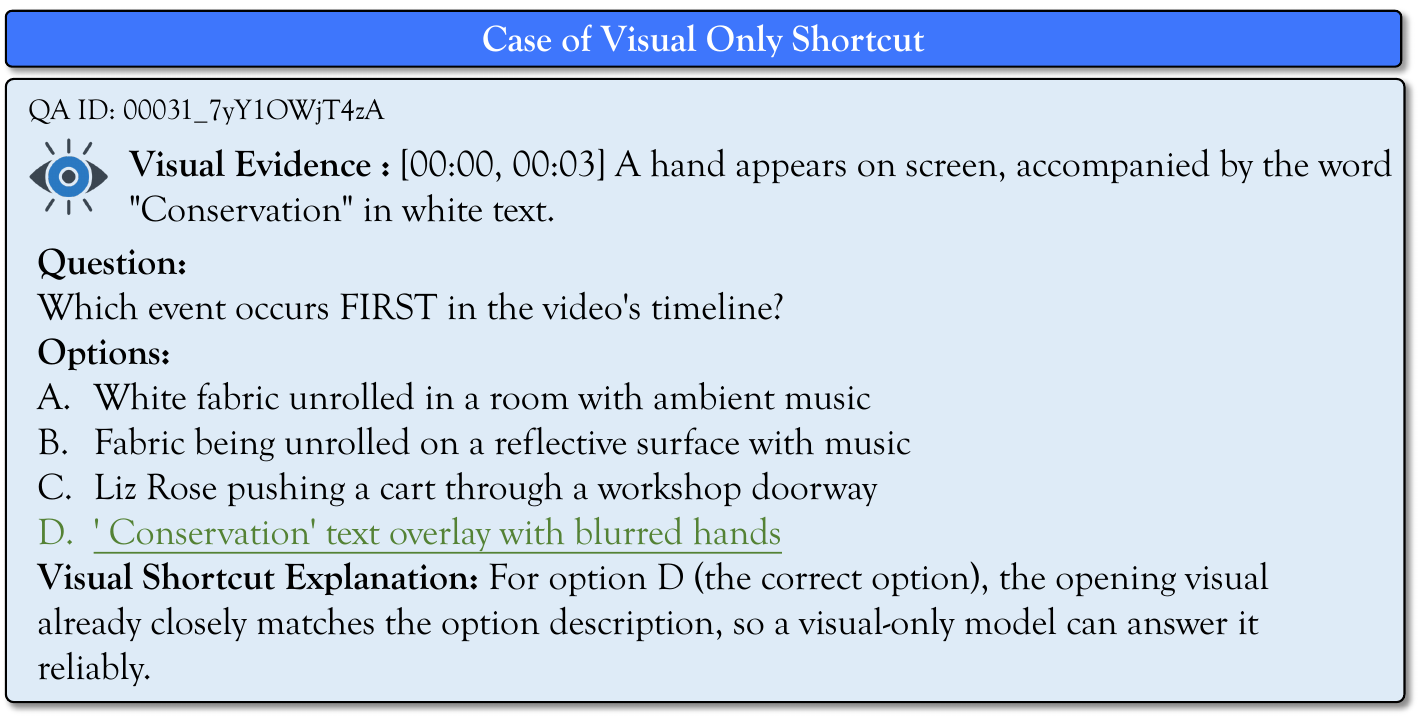}
    \caption{Visual-only shortcut case 2 from Daily-Omni.}
    \label{fig:dailyomni_visual2}
\end{figure}

\begin{figure}[p]
    \centering
    \includegraphics[width=\textwidth]{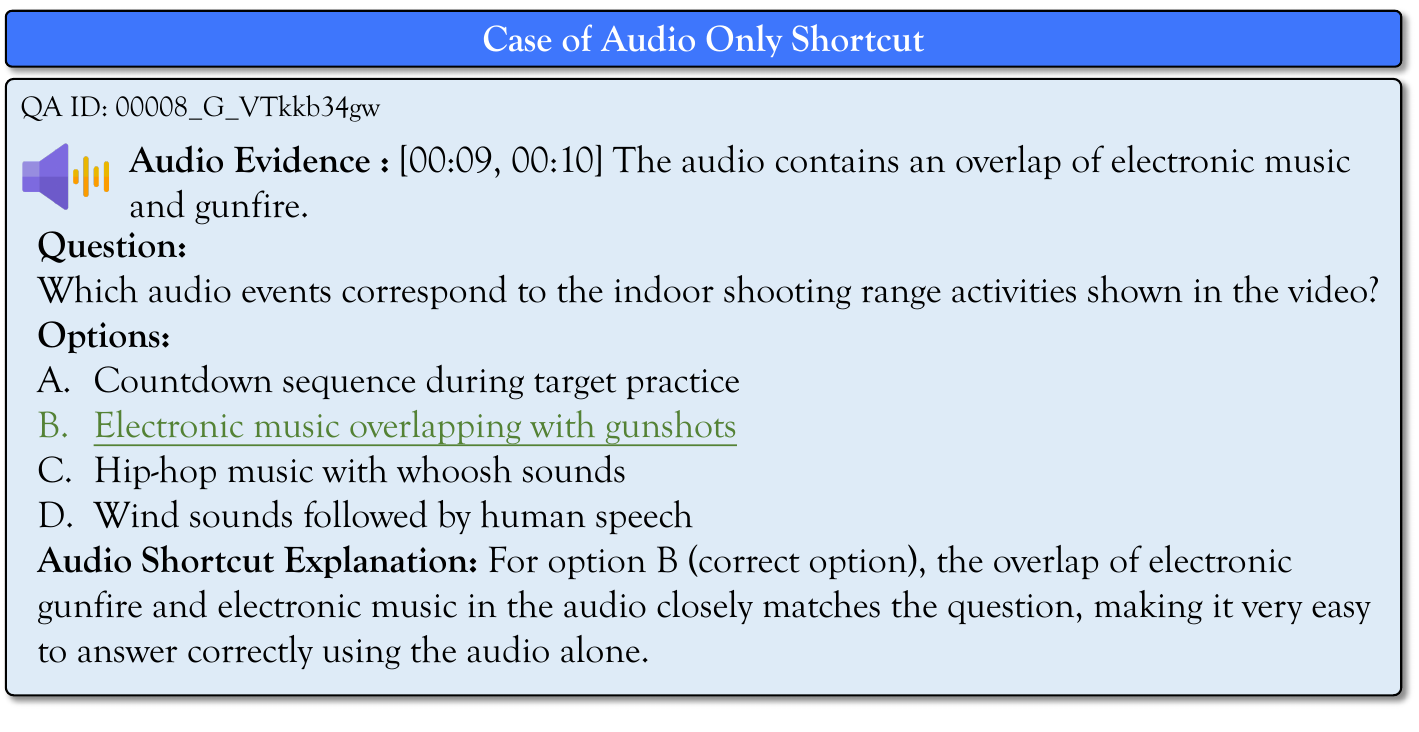}
    \caption{Audio-only shortcut case 1 from Daily-Omni.}
    \label{fig:dailyomni_audio1}
    \par\vspace{0.5em}
    \includegraphics[width=\textwidth]{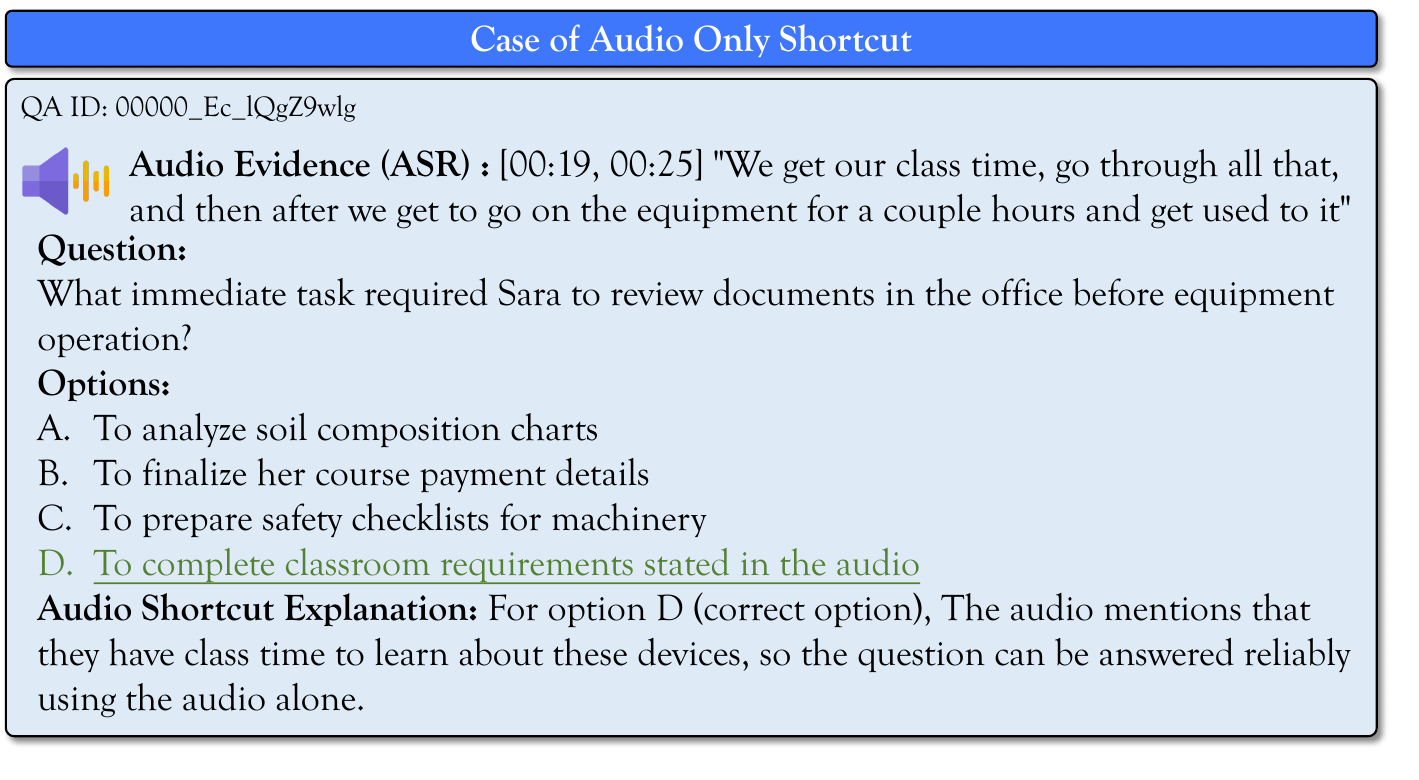}
    \caption{Audio-only shortcut case 2 from Daily-Omni.}
    \label{fig:dailyomni_audio2}
\end{figure}

\textbf{Illustrative Shortcut Cases.}
Figures~\ref{fig:worldsense_text1}--\ref{fig:worldsense_audio2},
\ref{fig:omnivideobench_text1}--\ref{fig:omnivideobench_audio2}, and
\ref{fig:dailyomni_text1}--\ref{fig:dailyomni_audio2} show two
text-only, two visual-only, and two audio-only shortcut cases from each
of WorldSense, OmniVideoBench, and Daily-Omni, respectively.

\clearpage

\clearpage

\clearpage
\subsection{Additional Grounded Evaluation Results}
\label{app:additional_grounded_results}

In addition to the Bottleneck Grounded Score (B.G.) and Coverage Grounded
Score (C.G.) reported in the main text, we report a continuous score and
threshold-based scores derived from the average temporal IoU across
audio and visual evidence. These metrics provide complementary views of
performance under different localization requirements.

For sample $i$, let $s_i$ denote the answer score defined in the main
text, and let $a_i$ and $v_i$ denote the temporal IoUs for audio and
visual evidence, respectively. Their average is
\begin{equation}
    m_i = \frac{a_i + v_i}{2}.
    \label{eq:mean_modal_iou}
\end{equation}
For an evaluation set of $N$ samples, we define
\begin{align}
    \mathrm{Mean}
    &= \frac{1}{N}\sum_{i=1}^{N} s_i m_i,
    \label{eq:mean_grounded_score}\\
    \mathrm{Thr}@\tau
    &= \frac{1}{N}\sum_{i=1}^{N}
    s_i\,\mathbf{1}\{m_i \geq \tau\},
    \qquad \tau \in \{0.3, 0.5, 0.7\},
    \label{eq:threshold_grounded_score}
\end{align}
where $\mathbf{1}\{\cdot\}$ is the indicator function. All results are
reported as percentages.

Mean continuously measures the joint performance of answering and
average evidence localization across modalities; it is not a localization
metric independent of answer correctness. For MCQs with binary answer
scores, $\mathrm{Thr}@\tau$ is the fraction of all samples that are
answered correctly and have an average IoU of at least $\tau$, rather
than a pass rate computed only over correct answers. Higher thresholds
impose stricter requirements on average localization. Unlike B.G., which
is sensitive to the weaker modality, average IoU allows stronger
localization in one modality to compensate for weaker localization in
the other. Passing a threshold on average IoU therefore does not require
both modality-specific IoUs to exceed that threshold.

\begin{table}[tbp]
  \centering
  \caption{\textbf{Additional grounded evaluation results across capability levels.}
  Mean averages the answer score weighted by the mean audio and visual IoU.
  Thr@$\tau$ averages the answer score after gating it by whether that
  mean IoU reaches $\tau$, for $\tau \in \{0.3,0.5,0.7\}$.
  All values are percentages; higher is better.
  As in Table~\ref{tab:main_results}, the first four open-weight models
  retain results from the earlier evaluation.
  Within each model group, \textbf{bold} and \underline{underlined} values
  indicate the best and second-best distinct scores in each column,
  respectively; ties receive the same formatting.}
  \label{tab:threshold_results}
  \vspace{0.4em}
  \begingroup
  \newcommand{\modelicon}[1]{\makebox[1.35em][c]{\raisebox{-0.15em}{\includegraphics[width=1.05em,height=1.05em,keepaspectratio]{figures/content/evaomni/#1}}}\hspace{0.2em}}
  \newcommand{\modeliconspace}{\hspace{1.55em}}
  \renewcommand{\arraystretch}{1.15}
  \setlength{\tabcolsep}{3pt}
  \resizebox{\linewidth}{!}{\begin{tabular}{@{}l*{16}{c}@{}}
      \hline
      \multirow{2}{*}{\textbf{Model}} &
      \multicolumn{4}{c}{\textbf{Level 1}} &
      \multicolumn{4}{c}{\textbf{Level 2}} &
      \multicolumn{4}{c}{\textbf{Level 3}} &
      \multicolumn{4}{c}{\textbf{Overall}} \\
      \cline{2-5}\cline{6-9}\cline{10-13}\cline{14-17}
      & \textbf{Mean} & \textbf{Thr@0.3} & \textbf{Thr@0.5} & \textbf{Thr@0.7}
        & \textbf{Mean} & \textbf{Thr@0.3} & \textbf{Thr@0.5} & \textbf{Thr@0.7}
        & \textbf{Mean} & \textbf{Thr@0.3} & \textbf{Thr@0.5} & \textbf{Thr@0.7}
        & \textbf{Mean} & \textbf{Thr@0.3} & \textbf{Thr@0.5} & \textbf{Thr@0.7} \\
      \hline
      \multicolumn{17}{c}{\textcolor[gray]{0.45}{\textbf{\rmfamily\itshape Open-Weight Omni LLMs}}} \\
      \hline
      \modelicon{qwen-icon-seeklogo.pdf}Qwen2.5-Omni-7B
        & 0.7 & 0.9 & 0.9 & 0.5 & 0.1 & 0.0 & 0.0 & 0.0
        & 0.0 & 0.0 & \underline{0.0} & 0.0 & 0.4 & 0.4& 0.0 & 0.0 \\
      \modelicon{qwen-icon-seeklogo.pdf}Qwen3-Omni-Instruct-30B-A3B
        & 2.8 & 5.3 & 2.5 & 0.0 & 1.2 & 1.1 & 0.6 & 0.0
        & 0.5 & 0.0 & \underline{0.0} & 0.0 & 1.8 & 2.9 & 1.4 & 0.0 \\
      \modelicon{qwen-icon-seeklogo.pdf}Qwen3-Omni-Thinking-30B-A3B
        & 6.5 & 10.2 & 7.4 & 2.8 & 2.5 & 3.4 & 1.1 & 0.6
        & 1.0 & 0.7 & \underline{0.0} & 0.0 & 4.1 & 6.0 & 3.8 & 1.5 \\
      \modelicon{nvidia_icon.png}Nemotron3-Nano-Omni-30B-A3B
        & 7.7 & 11.3 & 7.6 & 3.2 & 5.9 & 9.5 & 5.3 & 1.7
        & 1.7 & 2.0 & \underline{0.0} & 0.0 & 6.1 & 9.1 & 5.5 & 2.1 \\
      \modelicon{mimo_icon.png}MiMo-V2.5-310B-A15B & \textbf{37.2} & \textbf{56.7} & \underline{42.4} & \textbf{22.4}
        & \underline{20.9} & \underline{31.1} & \textbf{20.5} & \textbf{9.3}
        & \underline{10.5} & \underline{14.4} & \textbf{2.0} & \textbf{1.3}
        & \textbf{26.3} & \textbf{39.4} & \underline{27.0} & \textbf{13.6} \\
      \modelicon{mimo_icon.png}MiMo-V2.6-Flash-310B-A15B & \underline{35.3} & \underline{52.3} & \textbf{43.8} & \underline{20.3}
        & \textbf{21.0} & \textbf{32.5} & \textbf{20.5} & \underline{8.4}
        & \textbf{11.0} & \textbf{17.0} & \textbf{2.0} & 0.0
        & \underline{25.5} & \underline{38.6} & \textbf{27.6} & \underline{12.2} \\
      \modelicon{mimo_icon.png}MiMo-V2.6-Pro-1T-A42B & 33.0 & 48.6 & 40.6 & \underline{20.3}
        & 17.5 & 27.5 & \underline{16.4} & 7.0
        & 9.5 & 9.8 & \textbf{2.0} & \underline{0.7}
        & 22.8 & 33.8 & 24.4 & 11.7 \\
      \hline
      \multicolumn{17}{c}{\textcolor[gray]{0.45}{\textbf{\rmfamily\itshape Proprietary Omni LLMs}}} \\
      \hline
      \modelicon{doubao-icon-seeklogo.pdf}Doubao-Seed-2.0-Lite & 17.8 & 30.2 & 15.4 & 4.4
        & 18.7 & 28.6 & 16.8 & 8.9
        & 12.6 & 17.0 & 2.6 & 0.7
        & 17.4 & 27.6 & 14.1 & 5.7 \\
      \modelicon{doubao-icon-seeklogo.pdf}Doubao-Seed-2.1-Lite & 47.5 & 71.9 & 56.5 & 27.0
        & 30.8 & 46.6 & 30.0 & 14.1
        & 16.9 & 23.5 & 6.5 & 1.3
        & 35.8 & 53.8 & 37.7 & 17.6 \\
      \modelicon{qwen-icon-seeklogo.pdf}Qwen3.5-Omni-Flash & 20.7 & 34.3 & 18.4 & 6.9
        & 10.2 & 15.2 & 8.9 & 3.2
        & 4.9 & 3.9 & 1.3 & 0.7
        & 13.9 & 21.6 & 11.8 & 4.4 \\
      \modelicon{qwen-icon-seeklogo.pdf}Qwen3.5-Omni-Plus & 28.5 & 45.9 & 30.0 & 11.8
        & 12.1 & 18.2 & 10.7 & 3.9
        & 6.8 & 7.8 & 2.0 & 0.7
        & 18.2 & 28.3 & 17.5 & 6.7 \\
      \modelicon{qwen-icon-seeklogo.pdf}Qwen3.8-Omni-Flash & 26.7 & 44.5 & 27.4 & 7.1
        & 18.7 & 30.7 & 14.3 & 5.0
        & 13.8 & 17.0 & 5.9 & 1.3
        & 21.4 & 34.5 & 18.6 & 5.4 \\
      \modelicon{gemini_icon.png}Gemini2.5-Flash & 33.4 & 51.6 & 38.9 & 17.3
        & 16.5 & 24.3 & 15.9 & 8.9
        & 7.5 & 7.2 & 1.3 & 0.0
        & 22.3 & 33.3 & 23.5 & 11.1 \\
      \modelicon{gemini_icon.png}Gemini2.5-Flash-Lite & 27.4 & 41.7 & 30.9 & 13.4
        & 10.8 & 15.5 & 8.6 & 5.7
        & 4.3 & 2.6 & 0.7 & 0.0
        & 16.9 & 24.6 & 16.8 & 8.1 \\
      \modelicon{gemini_icon.png}Gemini2.5-Pro & 49.3 & 70.3 & 60.6 & 34.3
        & 30.6 & 45.9 & 32.3 & 14.5
        & 17.8 & 22.9 & 7.2 & 1.3
        & 36.6 & 52.8 & 40.5 & 20.9 \\
      \modelicon{gemini_icon.png}Gemini3.1-Flash-Lite & 31.1 & 47.9 & 36.2 & 15.7
        & 15.0 & 23.6 & 13.4 & 6.1
        & 5.3 & 3.9 & 0.7 & 0.7
        & 20.4 & 31.0 & 21.1 & 9.3 \\
      \modelicon{gemini_icon.png}Gemini3.1-Flash-Lite-Preview & 32.1 & 50.5 & 37.6 & 16.8
        & 15.2 & 23.4 & 12.7 & 6.1
        & 6.5 & 6.5 & 0.0 & 0.0
        & 21.1 & 32.3 & 21.3 & 9.7 \\
      \modelicon{gemini_icon.png}Gemini3.1-Pro-Preview & 48.6 & 68.9 & 60.8 & 31.6
        & 33.0 & 48.0 & 32.7 & 16.6
        & 20.1 & 28.8 & 5.2 & \underline{2.6}
        & 37.7 & 53.9 & 40.5 & 20.8 \\
      \modelicon{gemini_icon.png}Gemini3.5-Flash & 36.8 & 56.0 & 44.2 & 21.7
        & 35.5 & 52.0 & 36.8 & 20.0
        & \underline{20.6} & 27.5 & \underline{9.8} & 1.3
        & 33.8 & 50.0 & 35.9 & 17.9 \\
      \modelicon{gemini_icon.png}Gemini3.5-Flash-Lite & 33.5 & 51.6 & 38.7 & 18.4
        & 15.1 & 23.4 & 13.9 & 8.0
        & 8.3 & 9.8 & 0.7 & 0.7
        & 21.9 & 33.3 & 22.4 & 11.3 \\
      \modelicon{gemini_icon.png}Gemini3.6-Flash & 52.5 & 75.1 & 66.1 & 36.9
        & 35.3 & \underline{53.0} & 36.1 & 19.3
        & 20.5 & \underline{31.4} & 8.5 & 2.0
        & 40.4 & 59.1 & 44.7 & 24.1 \\
      \modelicon{gemini_icon.png}Gemini3.7-Flash & \textbf{57.0} & \textbf{81.8} & \textbf{72.4} & \underline{40.6}
        & \textbf{37.6} & \textbf{55.5} & \textbf{40.7} & \textbf{21.4}
        & \textbf{23.5} & \textbf{35.9} & \textbf{12.4} & \textbf{3.3}
        & \textbf{43.7} & \textbf{63.7} & \textbf{49.9} & \underline{26.8} \\
      \modelicon{gemini_icon.png}Gemini3.8-Flash & \underline{56.4} & \underline{79.3} & \underline{69.4} & \textbf{43.5}
        & \underline{36.6} & \underline{53.0} & \underline{38.9} & \underline{21.1}
        & 18.3 & 26.1 & 5.9 & 2.0
        & \underline{42.3} & \underline{60.1} & \underline{46.8} & \textbf{27.8} \\
      \modelicon{gemini_icon.png}Gemini4-Argon & 35.6 & 51.6 & 43.3 & 21.4
        & 17.5 & 28.6 & 17.0 & 6.4
        & 14.5 & 22.2 & 5.2 & 2.0
        & 24.7 & 37.4 & 26.4 & 12.1 \\
      \hline

    \end{tabular}}
  \endgroup
\end{table}

\textbf{Overall performance.}
Table~\ref{tab:threshold_results} shows that Gemini3.7-Flash achieves
the highest overall Mean (43.7\%), followed by Gemini3.8-Flash
(42.3\%) and Gemini3.6-Flash (40.4\%). Among the updated open-weight
models, MiMo-V2.5 obtains the highest overall Mean at 26.3\%, trailing
Gemini3.7-Flash by 17.4 percentage points. Thus, a substantial gap
between the leading open-weight and proprietary models persists when
the weaker-modality IoU used in B.G. is replaced by average IoU.

\textbf{Performance across capability levels.}
Mean decreases from Level~1 to Level~2 and again to Level~3 for all
20 models. For example, Gemini3.7-Flash scores 57.0\%,
37.6\%, and 23.5\% across the three levels, while MiMo-V2.5 scores
37.2\%, 20.9\%, and 10.5\%. This pattern is consistent with the greater
demands on evidence integration and temporal localization at higher
levels. Level~3 remains challenging even when localization quality
can be traded off across modalities. Because these scores also
incorporate answer correctness, however, the decline cannot be
attributed entirely to localization errors.

\textbf{Sensitivity to localization thresholds.}
Stricter average-IoU thresholds substantially reduce scores.
Gemini3.7-Flash obtains overall scores of 63.7\%, 49.9\%, and
26.8\% at thresholds of 0.3, 0.5, and 0.7, respectively; the
corresponding scores for MiMo-V2.5 are 39.4\%, 27.0\%, and 13.6\%.
Thus, many correctly answered samples satisfying a permissive
localization criterion fail to meet a stricter one. Rankings also
depend on the threshold: Gemini3.7-Flash leads in overall Mean,
Thr@0.3, and Thr@0.5, whereas Gemini3.8-Flash has a slightly higher
Thr@0.7 score (27.8\% versus 26.8\%). Continuous scores and results
at multiple thresholds therefore provide complementary information
about average performance and success under stricter localization
requirements.

\subsection{Details of Failure Mode Analysis }
\label{app:grounding_diagnostics}

\textbf{Scope and validation.}
We analyze the available predictions from Gemini3.7-Flash, Gemini3.5-Flash,
Qwen3.5-Omni-Plus, Gemini2.5-Pro, Gemini2.5-Flash, Qwen3.8-Omni-Flash,
and MiMo-V2.5-310B-A15B. Each model is evaluated on the same subset of the
available annotations, matched by question ID. We also verify
video IDs, questions, options, levels, and reference answers for every record.

\textbf{Scoring protocol.}
Ground-truth evidence is read from the annotation file for every model.
For each modality, we compute temporal IoU between the unions of the
predicted and annotated intervals, without multiplying by answer correctness.
Failed inference, response parsing, or truncated output receives zero answer
and localization scores and remains in the denominator. Comma-based decimal
timestamps are normalized numerically. These choices preserve a common
evaluation set without granting localization credit to malformed predictions.

Let $C$ denote answer correctness and
$G_{\tau}=\mathbb{1}[\min(\mathrm{IoU}_{\mathrm{audio}},
\mathrm{IoU}_{\mathrm{visual}})\geq\tau]$ denote successful dual-modal
localization. Unlike the mean-IoU threshold metric above, this criterion
requires each modality to meet the threshold.
Table~\ref{tab:grounding_diagnostics}
reports the
four outcomes at $\tau=0.5$;
each outcome is expressed as a percentage of all questions.

\textbf{Modality-specific localization.}
Six of the seven models in Table~\ref{tab:grounding_diagnostics} obtain higher
mean audio than visual IoU overall. Among these six, the difference ranges
from 0.7 percentage points for Qwen3.5-Omni-Plus to 3.7 points for
Gemini3.5-Flash. Qwen3.8-Omni-Flash shows the opposite pattern, with 27.4\%
audio IoU and 34.3\% visual IoU.

\textbf{Answer--grounding disagreement.}
Across the seven models, 47.8\%--79.8\% of correct answers fail the
dual-modal localization criterion at $\tau=0.5$.
For Gemini3.7-Flash, 47.8\% of correct answers fail grounding, whereas
12.2\% of grounding-passing predictions answer incorrectly. The converse error also occurs
in the other models: for MiMo-V2.5, 27.9\% of grounding-passing predictions
answer incorrectly. These outcomes distinguish temporal localization from
answer selection; temporal overlap alone establishes neither semantic
understanding nor causal use of the reported evidence.

\textbf{Threshold sensitivity.}
For Gemini3.7-Flash, the fraction of correct answers for which either audio
or visual IoU falls below the grounding threshold is 32.9\%, 47.8\%, and
78.2\% at thresholds of 0.3, 0.5, and 0.7, respectively. These proportions depend on the
threshold and are not direct evidence that correct answers were guessed.

\begin{figure}[t!]
    \centering
    \includegraphics[width=\textwidth]{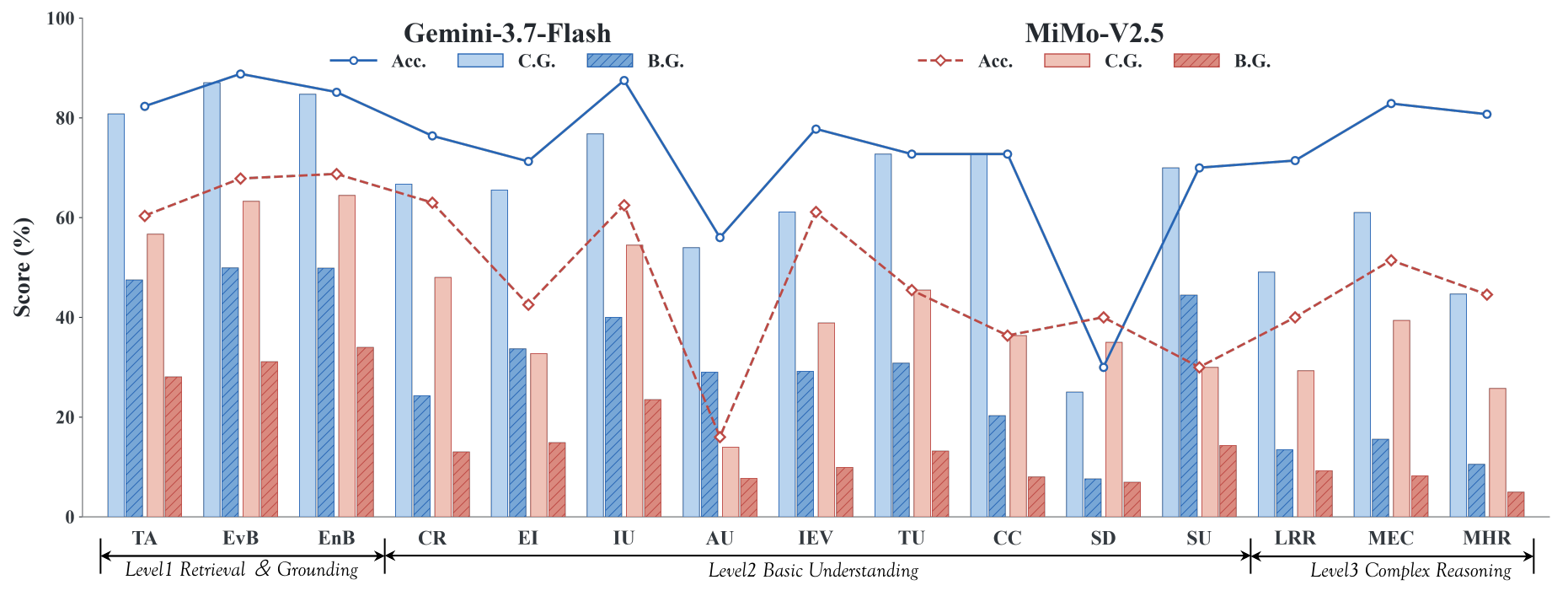}
    \caption{\textbf{Performance across fine-grained capabilities.}
    Lines show answer accuracy (Acc.); solid and hatched bars show Coverage Grounded Score (C.G.) and Bottleneck Grounded Score (B.G.), respectively. All values are percentages.
    Task abbreviations follow the capability taxonomy in Appendix~\ref{app:capability_taxonomy}: Level~1: TA = Temporal Alignment, EvB = Event Binding, EnB = Entity Binding; Level~2: CR = Causal Reasoning, EI = Emotional Intelligence, IU = Intent Understanding, AU = Atmosphere Understanding, IEV = Implicit Event Reasoning, TU = Temporal Understanding, CC = Conditional Counting, SD = Semantic Disambiguation, SU = Spatial Understanding; Level~3: LRR = Long-range Reasoning, MEC = Multi-Evidence Chain, MHR = Multi-hop Reasoning.}
    \label{fig:result_mimo_gemini}
\end{figure}
\subsection{Analysis Across Fine-Grained Capabilities}

Figure~\ref{fig:result_mimo_gemini} compares the two representative models across 15 fine-grained capabilities. On the three Level~1 alignment tasks, Gemini3.7-Flash achieves 82.3\%--88.8\% accuracy and 47.5\%--49.9\% B.G., compared with 60.4\%--68.8\% accuracy and 28.1\%--34.0\% B.G. for MiMo-V2.5. Both models show a larger separation between answering and precise grounding on complex tasks. On multi-hop reasoning, Gemini3.7-Flash reaches 80.7\% accuracy but only 10.6\% B.G., while MiMo-V2.5 reaches 44.6\% accuracy and 4.9\% B.G.; on multi-evidence chains, the corresponding scores are 82.9\%/15.6\% and 51.4\%/8.2\%. These comparisons suggest that selecting an answer is substantially easier for both models than locating the dispersed evidence needed to support it.

\subsection{More Results}
\label{app:more_results}

\begin{table}[!htbp]
    \centering
    \begin{minipage}{0.45\textwidth}
        \vspace{0pt}
        \begingroup
  \centering
  \captionof{table}{\textbf{Results with and without hints.} We report overall Acc. with and without hints. All values are percentages.}
  \label{tab:hint_results}
  \fontsize{7.5}{9}\selectfont
  \setlength{\tabcolsep}{3pt}
  \renewcommand{\arraystretch}{1.08}
  \resizebox{\linewidth}{!}{%
    \begin{tabular}{@{}lcc@{}}
      \toprule
      \textbf{Model} & \textbf{w.o. hint} & \textbf{w.\ hint} \\
      \midrule
      MiMoV2.5 & 56.8 & 81.6 \\
      Qwen3.8Omni-Flash & 58.2 & 85.1 \\
      \bottomrule
    \end{tabular}%
  }
  \par
\endgroup

    \end{minipage}
    \par\vspace{1em}
    \begin{minipage}{0.95\textwidth}
        \vspace{0pt}
        \begingroup
  \centering
  \captionof{table}{\textbf{Open-ended evaluation with different judges.}
  We report overall Acc., C.G., and B.G. using DeepSeekV4.1-flash-1017 and
  Qwen3.8Flash as judges. All values are percentages.}
  \label{tab:judge_results}
  \fontsize{7.5}{9}\selectfont
  \setlength{\tabcolsep}{3pt}
  \renewcommand{\arraystretch}{1.08}
  \resizebox{\linewidth}{!}{%
    \begin{tabular}{@{}lcccccc@{}}
      \toprule
      \multirow{2}{*}{\textbf{Model}}
        & \multicolumn{3}{c}{\textbf{Judge@DeepSeekV4.1-flash-1017}}
        & \multicolumn{3}{c}{\textbf{Judge@Qwen3.8Flash}} \\
      \cmidrule(lr){2-4}\cmidrule(l){5-7}
        & \textbf{Acc.} & \textbf{C.G.} & \textbf{B.G.}
        & \textbf{Acc.} & \textbf{C.G.} & \textbf{B.G.} \\
      \midrule
      MiMoV2.5 & 45.0 & 37.8 & 16.6 & 37.7 & 31.2 & 13.6 \\
      Qwen3.8Omni-Flash & 48.9 & 38.3 & 13.1 & 43.8 & 34.3 & 11.6 \\
      \bottomrule
    \end{tabular}%
  }
  \par
\endgroup

    \end{minipage}
\end{table}

Table~\ref{tab:hint_results} measures the effect of providing the hint during evaluation. We input the annotated evidence time segments and evidence descriptions, together with the audio/video and the question options, into the model as hint information, and observe whether the model can select the correct option. The results show: Adding the hint improves overall answer accuracy for both models: MiMoV2.5 increases from 56.8\% to 81.6\%, a gain of 24.8 percentage points, while Qwen3.8Omni-Flash increases from 58.2\% to 85.1\%, a gain of 26.9 points. The consistent improvement across the two models indicates that the hint supplies useful guidance for answer selection under this evaluation setting. This also demonstrates the reasonableness of our evidence. On the other hand, however, adding the hint does not enable the models to answer all questions correctly, which may be due to the models' own capabilities.

Table~\ref{tab:judge_results} compares evaluations in the open-ended format and also compares the effects of different judge models. The results show that DeepSeekV4.1-flash-1017 produces higher Acc., C.G., and B.G. values than Qwen3.8Flash for both evaluated models. The model ranking by answer accuracy is unchanged: Qwen3.8Omni-Flash outperforms MiMoV2.5 under both judges, with gaps of 3.9 and 6.1 percentage points, respectively. The C.G. ranking is also stable, whereas MiMoV2.5 has the higher B.G. under both judges. These results show that judge selection affects the absolute scores, especially for answer accuracy, while several relative comparisons remain consistent. This also indicates that evaluations in the open-ended format are affected by different judge models; therefore, EvaOmni combines MCQ and evidence supervision to balance the objectivity and interpretability of evaluation.
\FloatBarrier
\section{Limitations and Future Work}
\label{app:limitations_future_work}

\subsection{Limitations}

\textbf{Dataset coverage.}
EvaOmni contains 1,027 question-answer pairs drawn from 392 videos lasting
1--12 minutes with sufficient audio-visual cues. Excluding hard-coded
subtitles and prioritizing varied entities and events support the
construction of questions requiring both modalities, but also constrain
the data distribution. The results therefore do not establish
generalization to substantially longer videos, subtitle-rich content,
scenes with sparse events, or all linguistic and cultural contexts.
The three capability levels characterize different task requirements;
they do not guarantee a strictly increasing difficulty ordering in
answer accuracy across different question sets.

\subsection{Future Work}

\textbf{Broader data and task coverage.}
Future extensions could include longer videos, a wider range of
linguistic and cultural contexts, and scenes with sparser audio-visual
cues while retaining checks for the necessity of both modalities.
Subtitle-rich content would also enable explicit separation of text
reading from audio understanding. Broader evaluation of open-ended
question answering could test whether the current findings hold
without cues from answer options.

\end{document}